\documentclass[runningheads]{llncs}
\newif\ifarxiv
\arxivfalse

\newif\ifblurface
\blurfacetrue
\usepackage{graphicx}

\usepackage{tikz}
\usepackage{comment}
\usepackage{amsmath,amssymb} %
\usepackage{wrapfig} %
\usepackage{color}

\usepackage{mathtools}
\usepackage[accsupp]{axessibility}  %

\usepackage[utf8]{inputenc} %
\usepackage[T1]{fontenc}    %
\usepackage{xr-hyper}
\usepackage{hyperref}       %
\usepackage{url}            %
\usepackage{booktabs}       %
\usepackage{xcolor,colortbl}         %
\usepackage{caption}

\usepackage{multirow}
\usepackage{enumitem}

\usepackage[normalem]{ulem}

\begin{document}
\title{DANBO: Disentangled Articulated Neural Body Representations via Graph Neural Networks}
\ifarxiv %
\author{
Shih-Yang Su$^1$ \qquad Timur Bagautdinov$^2$ \qquad Helge Rhodin$^1$\\
\\
$^1$University of British Columbia \qquad $^2$Reality Labs Research
}%
\else %
\pagestyle{headings}
\mainmatter
\def\ECCVSubNumber{4883}  %

\titlerunning{DANBO: Disentangled Articulated Neural Body Representations}
\author{
Shih-Yang Su$^1$ \qquad Timur Bagautdinov$^2$ \qquad Helge Rhodin$^1$%
}%
\authorrunning{S.-Y. Su et al.}
\institute{University of British Columbia \and
Reality Labs Research}
\fi

\maketitle
\newcommand{\TODO}[1]{\textcolor{red}{TODO: #1}}

\newcommand{\tbf}[1]{\textbf{#1}}

\newcommand{\figref}[1]{Figure~\ref{#1}}
\newcommand{\secref}[1]{Section~\ref{#1}}
\newcommand{\feqref}[1]{Equation~\eqref{#1}}
\newcommand{\tabref}[1]{Table~\ref{#1}}

\newcommand{\app}{appendix} %

\newcommand{\parag}[1]{\noindent\emph{#1}}

\newcommand{\new}[1]{{\color{red}{#1}}}
\newcommand{\old}[1]{\textcolor{red}{\sout{#1}}}

\newcommand{\R}{\mathbb{R}}

\newcommand{\va}{\mathbf{a}}
\newcommand{\vb}{\mathbf{b}}
\newcommand{\vc}{\mathbf{c}}
\newcommand{\vd}{\mathbf{d}}
\newcommand{\ve}{\mathbf{e}}
\newcommand{\vf}{\mathbf{f}}
\newcommand{\vg}{\mathbf{g}}
\newcommand{\vh}{\mathbf{h}}
\newcommand{\vi}{\mathbf{i}}
\newcommand{\vj}{\mathbf{j}}
\newcommand{\vk}{\mathbf{k}}
\newcommand{\vl}{\mathbf{l}}
\newcommand{\vm}{\mathbf{m}}
\newcommand{\vn}{\mathbf{n}}
\newcommand{\vo}{\mathbf{o}}
\newcommand{\vp}{\mathbf{p}}
\newcommand{\vq}{\mathbf{q}}
\newcommand{\vr}{\mathbf{r}}
\newcommand{\vt}{\mathbf{t}}
\newcommand{\vu}{\mathbf{u}}
\newcommand{\vv}{\mathbf{v}}
\newcommand{\vw}{\mathbf{w}}
\newcommand{\vx}{\mathbf{x}}
\newcommand{\vy}{\mathbf{y}}
\newcommand{\vz}{\mathbf{z}}

\newcommand{\mA}{\mathbf{A}}
\newcommand{\mB}{\mathbf{B}}
\newcommand{\mC}{\mathbf{C}}
\newcommand{\mD}{\mathbf{D}}
\newcommand{\mE}{\mathbf{E}}
\newcommand{\mF}{\mathbf{F}}
\newcommand{\mG}{\mathbf{G}}
\newcommand{\mH}{\mathbf{H}}
\newcommand{\mI}{\mathbf{I}}
\newcommand{\mJ}{\mathbf{J}}
\newcommand{\mK}{\mathbf{K}}
\newcommand{\mL}{\mathbf{L}}
\newcommand{\mM}{\mathbf{M}}
\newcommand{\mN}{\mathbf{N}}
\newcommand{\mO}{\mathbf{O}}
\newcommand{\mP}{\mathbf{P}}
\newcommand{\mQ}{\mathbf{Q}}
\newcommand{\mR}{\mathbf{R}}
\newcommand{\mS}{\mathbf{S}}
\newcommand{\mT}{\mathbf{T}}
\newcommand{\mU}{\mathbf{U}}
\newcommand{\mV}{\mathbf{V}}
\newcommand{\mW}{\mathbf{W}}
\newcommand{\mX}{\mathbf{X}}
\newcommand{\mY}{\mathbf{Y}}
\newcommand{\mZ}{\mathbf{Z}}

\newcommand{\cA}{\mathcal A}
\newcommand{\cB}{\mathcal B}
\newcommand{\cC}{\mathcal C}
\newcommand{\cD}{\mathcal D}
\newcommand{\cE}{\mathcal E}
\newcommand{\cF}{\mathcal F}
\newcommand{\cG}{\mathcal G}
\newcommand{\cH}{\mathcal H}
\newcommand{\cI}{\mathcal I}
\newcommand{\cJ}{\mathcal J}
\newcommand{\cK}{\mathcal K}
\newcommand{\cL}{\mathcal L}
\newcommand{\cM}{\mathcal M}
\newcommand{\cN}{\mathcal N}
\newcommand{\cO}{\mathcal O}
\newcommand{\cP}{\mathcal P}
\newcommand{\cQ}{\mathcal Q}
\newcommand{\cR}{\mathcal R}
\newcommand{\cS}{\mathcal S}
\newcommand{\cT}{\mathcal T}
\newcommand{\cU}{\mathcal U}
\newcommand{\cV}{\mathcal V}
\newcommand{\cW}{\mathcal W}
\newcommand{\cX}{\mathcal X}
\newcommand{\cY}{\mathcal Y}
\newcommand{\cZ}{\mathcal Z}

\newcommand{\bR}{\mathbb{R}}
\newcommand{\mx}{\mathbf{x}}
\newcommand{\mj}{\mathbf{j}}
\newcommand{\mb}{\mathbf{b}}

\definecolor{Gray}{gray}{0.85}
\newcolumntype{a}{>{\columncolor{Gray}}c}
\newcolumntype{b}{>{\columncolor{white}}c}

\newcommand{\ourapproachfull}{Disentangled Articulated Neural BOdy}
\newcommand{\ourapproach}{DANBO}
\newcommand{\image}{\mI}
\newcommand{\bodynet}{G}
\newcommand{\pose}{\theta}
\newcommand{\jrot}{\omega} %
\newcommand{\volrefshoulder}{D_\text{shoulder}}
\newcommand{\volrefcollar}{D_\text{collar}}
\newcommand{\volrefknee}{D_\text{knee}}
\newcommand{\volrefmax}{\vert\vert\mj_\text{max}\vert\vert}
\newcommand{\volrefi}{\vert\vert\mj_{i,j}\vert\vert}

\newcommand{\bonevolume}{V} %
\newcommand{\bonefvolume}{v} %
\newcommand{\bonefvolumex}{\bonefvolume^x} %
\newcommand{\bonefvolumey}{\bonefvolume^y} %
\newcommand{\bonefvolumez}{\bonefvolume^z} %
\newcommand{\bonefvolumexi}{\bonefvolumex_i} %
\newcommand{\bonefvolumeyi}{\bonefvolumey_i} %
\newcommand{\bonefvolumezi}{\bonefvolumez_i} %
\newcommand{\bonefscale}{s}
\newcommand{\bonefscalex}{\bonefscale^x}
\newcommand{\bonefscaley}{\bonefscale^y}
\newcommand{\bonefscalez}{\bonefscale^z}
\newcommand{\projected}[1]{\localquery_i(#1)}
\newcommand{\bonefvolumei}{\left[\bonefvolumexi,\bonefvolumeyi,\bonefvolumezi\right]} %

\newcommand{\aggweight}{p}
\newcommand{\aggvoxelfeat}{\hat{\voxelfeat}}
\newcommand{\agglogit}{a}
\newcommand{\oob}{o}

\newcommand{\render}{\hat{\image}}

\newcommand{\bodyfeat}{f}
\newcommand{\bone}{b}
\newcommand{\Bone}{B}
\newcommand{\query}{\mx}
\newcommand{\localquery}{\hat{\mx}}

\newcommand{\proj}{\text{proj}}
\newcommand{\viewdir}{\mathbf{d}}
\newcommand{\voxelfeat}{h}
\newcommand{\aggnet}{A}
\newcommand{\geodist}{G}
\newcommand{\neuralfield}{F}
\newcommand{\radiance}{c}
\newcommand{\density}{\sigma}
\newcommand{\node}{\mathbf{n}}
\newcommand{\nodeli}[2]{\mathbf{n}^{(#1)}_{#2}}
\newcommand{\gnnmsgli}[2]{\mathbf{m}^{#1}_{#2}}
\newcommand{\nodel}[1]{\mathbf{n}^{(#1)}}
\newcommand{\Nodel}[1]{\mN^{(#1)}}
\newcommand{\gnnA}{\mA}
\newcommand{\gnnAl}[1]{\gnnA^{(#1)}}
\newcommand{\gnnWli}[2]{\mW^{(#1)}_{#2}}
\newcommand{\gnnbli}[2]{\mb^{(#1)}_{#2}}
\newcommand{\gnnbl}[1]{\mb^{(#1)}}
\newcommand{\veclist}[2]{
\left[#1_1,#1_2,\cdots,#1_{#2}\right]
}
\newcommand{\veclistdot}[3]{
\left[#1_1 #2_1,#1_2 #2_2,\cdots,#1_{#2} #2_{#2}\right]
}
\newcommand{\mRvr}{\mathbf{R}_{\text{virt}\rightarrow\text{real}}}
\newcommand{\mRrv}{\mRvr^{-1}}%

\newlength\teasersa
\setlength\teasersa{\linewidth} %
\newlength\teasersb
\setlength\teasersb{-0.0mm}

\begin{center}
\parbox[t]{\teasersa}{
\centering%
\ifblurface
\includegraphics[width=\teasersa,trim=1 15 10 35,clip]{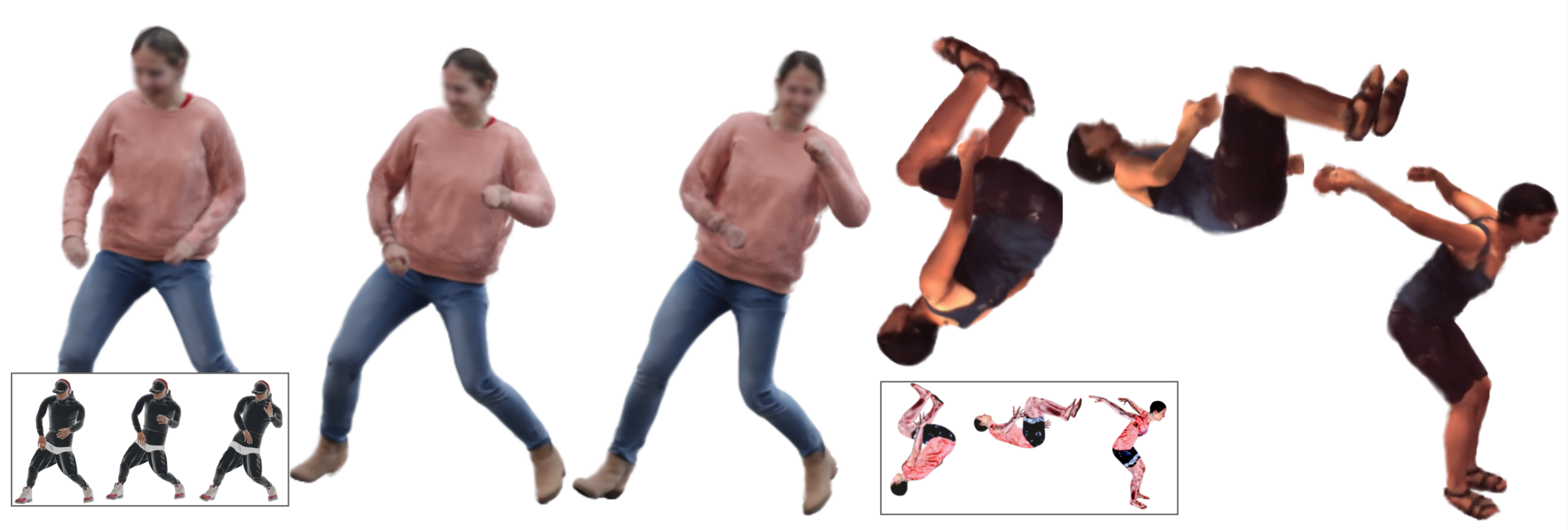}\\%
\else
\includegraphics[width=\teasersa]{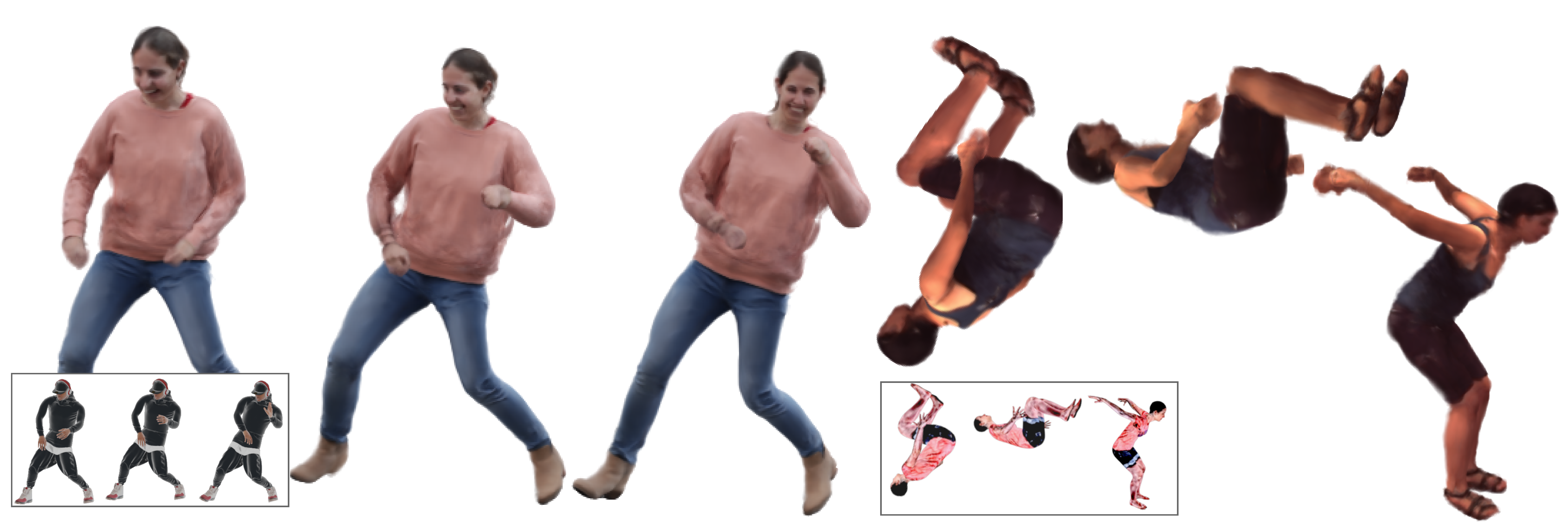}\\%
\fi
 }%
  \captionof{figure}{
    \ourapproach~enables learning volumetric body models from scratch, only 
    requiring a single video as input, yet enable driving by unseen poses (inset) that are out of the training distribution, showing better robustness than existing surface-free approaches.%
    \ifblurface
    \textbf{ Real faces are blurred for anonymity.}
    \fi
  }
  \label{fig:teaser}
\end{center}

\begin{abstract}
Deep learning greatly improved the realism of animatable human models by learning geometry and appearance from collections of 3D scans, template meshes, and multi-view imagery. %
High-resolution models enable photo-realistic avatars but at the cost of 
requiring studio settings not available to end users.
Our goal is to create avatars directly from raw images without relying on 
expensive studio setups and surface tracking.
While a few such approaches exist, those have limited generalization 
capabilities and are prone to learning spurious (chance) correlations between irrelevant body parts, %
 resulting in implausible deformations and 
missing body parts on unseen poses.
We introduce a three-stage method that induces two inductive biases to better disentangled pose-dependent deformation. 
First, we model correlations of body parts explicitly with a
graph neural network. Second, to further reduce the effect of chance correlations, we introduce localized per-bone features
that use a factorized volumetric representation and a new aggregation function.
We demonstrate that our model produces realistic body shapes under challenging 
unseen poses and shows high-quality image synthesis. 
Our proposed representation strikes a better trade-off between 
model capacity, expressiveness, and robustness than competing methods.
Project website: {\color{magenta}\url{https://lemonatsu.github.io/danbo}}.
\end{abstract}

\section{Introduction}
Animating real-life objects in the digital world is a long-pursued goal in computer vision and graphics, and recent advances already enable 3D free-viewpoint video, animation, and human performance retargeting~\cite{gao2021dynerf,peng2020neuralbody,xian2021space}. 
Nevertheless, animating high-definition virtual avatars with user-specific appearance
and dynamic motion still remains a challenge: human body and clothing deformation are inherently complex, unique, and modeling their intricate effects require 
dedicated approaches. 
Recent solutions achieve astonishing results %
\cite{chen2021snarf,liu2021neuralactor,saito2021scanimate,tiwari2020sizer,tiwari21neuralgif} when grounding on 3D data capture in designated studio settings, e.g., with multi-camera capture systems and controlled illumination---inaccessible to the general public for building personalized models.

Less restrictive are methods relying on parametric body models~\cite{loper2015smpl} 
that learn plausible body shape, pose, and deformation from a collection of 3D scans. 
These methods can thereby adapt to a wide range of body shapes~\cite{balan2007detailed,choutas2020smplx,osman2020star}, 
in particular when using neural approaches to model details as a
dynamic corrective~\cite{bhatnagar2020ipnet,burov2021dsfn,corona2021smplicit}. Even though subject-specific details such as clothing can be learned, it remains difficult to generalize to shapes vastly different from the original scans. 
Moreover, the most widely used body models have restrictive commercial licenses~\cite{loper2015smpl} 
and 3D scan datasets to train these afresh are expensive. 

Our goal is to learn a high-quality model with subject-specific details directly from images. %
Recent approaches in this class~\cite{noguchi2021narf,su2021anerf} use a neural radiance field (NeRF) that is attached to a human skeleton initialized with an off-the-shelf 3D human pose estimator. Similar to the original NeRF, the shape and appearance
are modeled implicitly with a neural network that takes as input a query
location and outputs density and radiance, and is only supervised with 
images through differentiable volume rendering.
However, unlike the original that models static scenes, 
articulated NeRFs model time-varying body shape deformations by
conditioning on per-frame body pose and representing each frame
with the same underlying body model albeit in a different state. 
This results in an animatable full-body model that is trained directly 
from videos and can then can be driven with novel skeleton poses.

Not using an explicit surface poses a major difficulty as surface-based solutions exploit surface points to anchor neural features locally as vertex attributes~\cite{peng2020neuralbody}, and leverage skinning weights to associate points on or close to the surface to nearby body parts~\cite{liu2021neuralactor,peng2021animatable}.
In absence of such constraints, 
A-NeRF~\cite{su2021anerf} uses an overparametrization by encoding 3D position 
relative to all body parts. Thereby dependencies between a point and body parts are learned implicitly. 
By contrast, NARF~\cite{noguchi2021narf} explicitly predicts probabilities for the association of 3D points to body parts, similar to NASA~\cite{deng2019nasa}. 
However, this probability predictor
is conditioned on the entire skeleton pose and is itself prone to poor generalization.
Therefore, both approaches rely on large training datasets and 
generalization to unseen poses is limited---in particular because unrelated body parts remain weakly entangled.  

In this paper, we introduce \emph{\ourapproachfull{}} (\ourapproach{}), a surface-free approach that explicitly disentangles independent body parts for learning a generalizable personalized human model from unlabelled videos. 
It extends the established articulated NeRF-based body modeling with two additional stages, a body part-specific volumetric encoding that exploits human skeleton structure as a prior using Graph Neural Networks (GNN)~\cite{kipf2016semi}, and a new aggregation module. Both designs are tailored for learning from few examples and optimized to be parameter efficient.
Our main contributions are the following:
\begin{itemize}
    \item A surface-free human body model with better texture detail and improved generalization when animated.
    \item %
    GNN-based encoding that disentangles features from different body parts and relies on factorized per-bone volumes for efficiency. 
    \item A part-based feature aggregation strategy that improves on and is informed by a detailed evaluation of existing aggregation functions.%
\end{itemize}

We demonstrate that our proposed~\ourapproach~ results in a generalizable neural body model, with quality comparable to surface-based approaches.
\section{Related Work}
We start our review with general-purpose neural fields and then turn to human body modeling with a focus on animatable neural representations. %

\paragraph{Neural fields.}
Neural fields~\cite{mildenhall2020nerf,park2019deepsdf,sitzmann2020implicit_siren} 
have attracted recent attention due to their ability to capture 
astonishing levels of detail. 
Instead of relying on explicit geometry representations such as triangle 
meshes or voxel grids, these methods represent the scene implicitly - as 
a continuous function - that maps every point in the 3D space to quantities
of interest such as radiance, density, or signed distance. 
This approach was popularized with Neural Radiance Fields (NeRF)~\cite{mildenhall2020nerf} demonstrating impressive results on reconstructing static 3D scene presentation directly from calibrated multi-view images. 
Various subsequent works focused on improving performance on static scenes in 
terms of generalization~\cite{yu2020pixelnerf}, level of
detail~\cite{barron2021mipnerf,park2021hypernerf}, 
camera self-calibration~\cite{lin2021barf,yen2020inerf}, and 
resource efficiency~\cite{lindell2021autoint,yu2021plenoctrees}. 
Most relevant are deformable models that capture non-static scenes with 
deformation fields~\cite{gafni2021dynamic,gao2021dynerf,pumarola2020dnerf,tretschk2021nonrigid,xian2021space}. %
However, general deformation fields are unsuitable for animation and no one demonstrated that they can generalize to strongly articulated motion in monocular video.

\paragraph{Template-based body models.}
The highest level of detail can be attained with specialized performance capture
systems, e.g., with dozens of cameras or a laser scanner~\cite{guo2019relightables}. 
The resulting template mesh can then be textured and deformed for capturing 
high-quality human performances, even from a single video~\cite{zhou2018monocap}.
Neural approaches further enable learning pose-dependent appearance and 
geometries~\cite{bagautdinov2021driving}, predict vertex 
displacements~\cite{habermann2021rddc} or local primitive 
volumes~\cite{Lombardi21mixture} for creating fine-grained local geometry 
and appearance including cloth wrinkles and freckles. 
The most recent ones use neural fields to learn implicit body models with the mesh
providing strong supervision signals~\cite{alldieck2021imghum,chen2021snarf,saito2021scanimate,wang2021metaavatar,xu2021hnerf}.
However, template creation is limited to expensive controlled studio environments, 
often entails manual cleaning, and high-quality ground truth annotations.

\paragraph{Parametric Human Body Models}
 learn common shape and pose properties from a large corpus of 3D scans~\cite{balan2007detailed,choutas2020smplx,loper2015smpl,osman2020star}. For classical approaches the result are factorized parameters for controlling pose, shape~\cite{choutas2020smplx,loper2015smpl,osman2020star,xu2020ghum} and even clothing~\cite{tiwari2020sizer} that can fit to a new subject. Most prevalent is the SMPL~\cite{choutas2020smplx,loper2015smpl} mesh model with a linear shape space and pose-dependent deformation. 
However, most existing models have restrictive commercial licenses and modeling person-specific details from images requires additional reconstruction steps. 

\paragraph{Personalized Body Models.}
Learning personalized body models given only videos of a single actor is particularly 
challenging.
Most existing approaches start from estimating a parametric surface model such 
as SMPL and extend it to learn specifics. 
For instance, one can anchor neural features spatially by
associating each SMPL vertex with a learnable latent feature, and then either diffuse vertex features to the 3D space~\cite{kwon2021neural,peng2020neuralbody} or project the 3D query point to the SMPL surface for feature retrieval. 
Incorrect shape estimates and missing details can then be corrected by a 
subsequent neural rendering step. 
As texturing improves classical approaches, neural texture mapping provides additional rendering 
quality~\cite{liu2021neuralactor}.
Another line of work makes use of SMPL blend skinning weights as initialization for learning deformation fields~\cite{peng2021animatable}. The deformation field maps 3D points from 3D world space to canonical space, which enables learning a canonical neural body field that predicts the radiance and density for rendering as for the classical NeRF on static scenes. While the skinning weights in SMPL provide an initialization, \cite{peng2021animatable} showed that fine-tuning the deformation fields via self-supervision helps rendering unseen poses. 
However, relying on body models imposes the previously discussed limitations.

There are few methods that target learning neural body fields from 
images without relying on an explicit surface model. 
Closest to our method are NARF~\cite{noguchi2021narf} and A-NeRF~\cite{su2021anerf}, that learn articulated body models directly from image sequences, leveraging 3D body 
pose estimates produced by off-the-shelf 
approaches~\cite{kolotouros2019learning_spin,kocabas2020vibe}. 
These methods encode 3D query points with respect to each bone on the posed skeleton, 
and either explicitly predict blending weights~\cite{noguchi2021narf} to select the 
parts of influence or  
rely on a neural network to learn the assignment implicitly by feeding it a large
stack of all relative positions~\cite{su2021anerf}. 
However, lacking a prior for part assignments leads to spurious dependencies between irrelevant body parts when the training poses are scarce and have low diversity~\cite{bagautdinov2021driving,saito2021scanimate}.
Our approach follows the same surface-free setting but improves upon these by introducing body part disentangled representations 
and a new aggregation function that achieves better rendering quality and 
improved generalization on novel body poses. A concurrent work COAP~\cite{mihajlovic2022compositional_coap} shares a similar part-disentangle concept, but differs significantly. COAP models part geometries separately from 3D scans, whereas DANBO leverages the skeleton structure as a prior to fuse information from neighboring body parts, and learns both appearance and body geometry from images without 3D supervisions.

\begin{figure*}[t]
\centering
\ifblurface
\includegraphics[width=\linewidth,trim=0 0 0 0,clip]{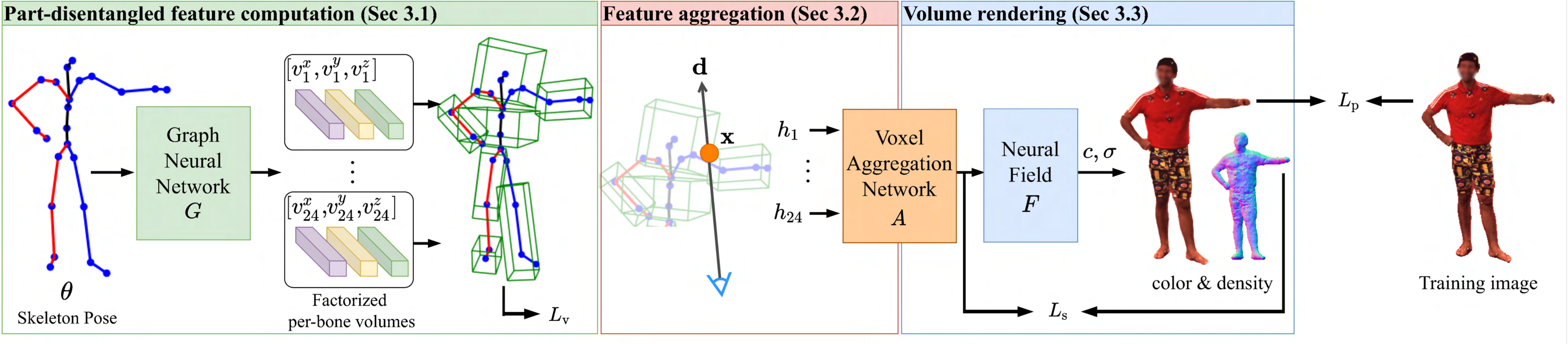}%
\else
\includegraphics[width=\linewidth,trim=0 0 0 0,clip]{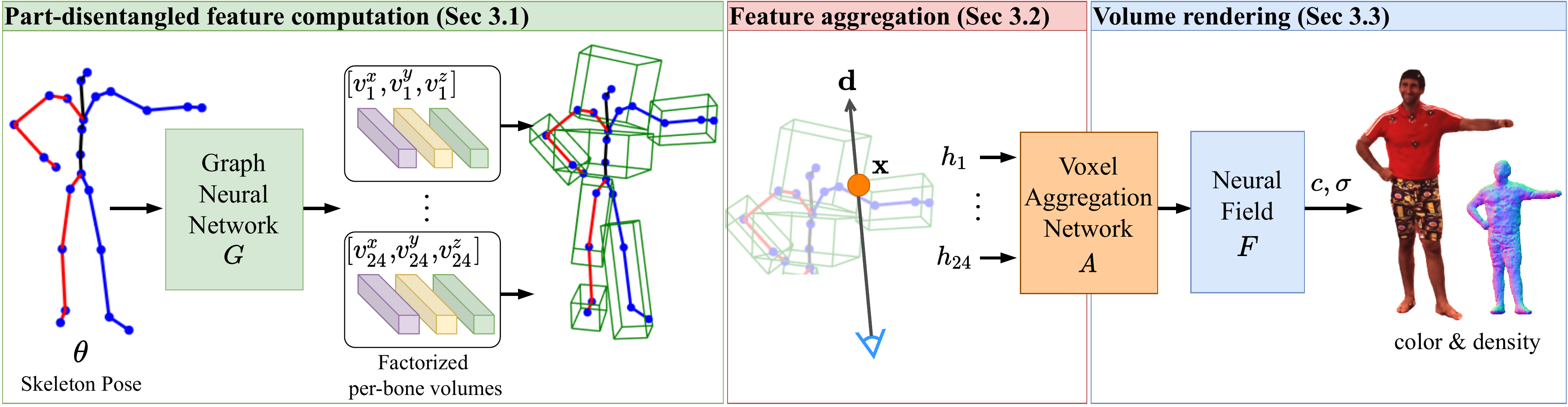}%
\fi
\caption{Overview. The final image is generated via volume rendering by sampling points $\query$ along the ray $\viewdir$ as in the original NeRF. Different is the conditioning on pose.
First, pose features are encoded locally to every bone of a skeleton with a graph neural network using factorized volumes to increase efficiency (green boxes). Second, these disentangled features are queried and aggregated with learned weights (red module).
Finally, the body shape and appearance are predicted via density and radiance fields $\sigma$ and $c$ (blue module). 
}
\vspace{-3mm}
\label{fig:overview}
\end{figure*}

\section{Method}
\label{sec:method}
\label{sec:method-overview}
Our goal is to learn an animatable avatar from a collections of $N$ images $\left[\image_k\right]_{k=1}^N$ and the corresponding body pose in terms of skeleton joint angles $\left[\pose_k\right]_{k=1}^N$ that can stem from an off-the-shelf estimator~\cite{kolotouros2019learning_spin,kocabas2020vibe}, 
without using laser scans or surface tracking. 
We represent the human body as a neural field that moves with the input body pose. The neural field maps each 3D location to color and density to generate a
free-viewpoint synthetic image via volume rendering.
See~\figref{fig:overview} for an overview. 
Our approach consists of three stages that are trained end-to-end.
The first stage predicts a localized volumetric representation for each body part
with a Graph Neural Network (GNN). GNN has a limited receptive field and encodes only locally relevant pose information---which naturally leads to better disentangling
between body parts in the absence of surface priors.
This stage is independent of the query locations and is thus 
executed only once per frame.
Additional performance is gained by using a factorized volume and encouraging the 
volume bounds to be compact.
The second stage retrieves a feature code for each query point
by sampling volume features for all body parts that enclose the point and 
then aggregating the relevant ones using a separate network that predicts blend weights.
Finally, the third stage maps the resulting per-query feature code 
to the density and radiance at that location, followed by 
the volume rendering as in the original NeRF.

\subsection{Stage I: Part-disentangled Feature Computation}
\label{sec:method-gnn}
Given a pose $\pose=\veclist{\jrot}{24}$, where $\jrot_i\in \R^6$~\cite{zhou2019continuity} defines the rotation of the bone $i=1,2,\cdots,24$, we represent the body part attached to each bone $i$ with a coarse volume $\bonevolume$ (green boxes in Fig.~\ref{fig:overview}), predicted by a neural network~$\bodynet$,
\begin{equation}
    \veclist{\bonevolume}{24}=\bodynet(\pose).
\end{equation}
We design $\bodynet$ as a GNN operating on the skeleton graph with 
nodes initialized to the corresponding joint angles in $\pose$.
In practice, we use two graph convolutional layers followed by per-node 
2-layer MLPs. 
Because the human skeleton is irregular, we learn
individual MLP weights for every node. See the supplemental for additional details on the graph network.

\paragraph{Factorized volume.} 

A straight-forward way to represent a volume is via a dense voxel grid, which has cubic
complexity with respect to its resolution.
Instead, we propose to factorize each volume $V_i = (\bonefvolumex_i,\bonefvolumey_i,\bonefvolumez_i$) as one
vector $v_i \in \mR^{H\times M}$ for each 3D axis,
where $H$ is the voxel feature channel, and is $M$ the volume resolution. This is similar to \cite{peng2020convolutional} doing a factorization into 2D planes.

\newlength\voxretscale
\setlength\voxretscale{0.3\textwidth}
\setlength{\intextsep}{0pt}%
\begin{wrapfigure}[13]{r}{\voxretscale}
\centering
\setlength{\fboxrule}{0pt}%
\parbox[t]{\voxretscale}{%
\centering%
\fbox{\includegraphics[width=\voxretscale]{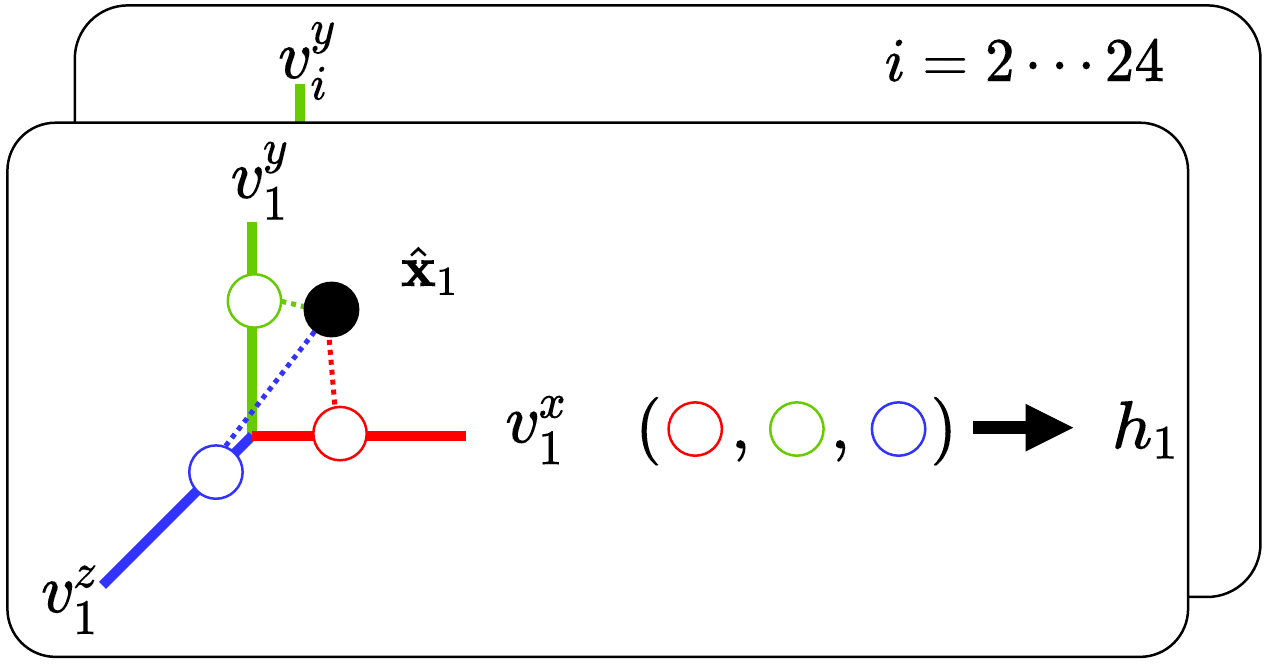}
}%
}
\caption{\small We retrieve the voxel feature by projecting $\localquery$ to the three axes and linearly interpolating the feature neighboring the projected location.
}
\label{fig:voxel-retrieval}
\end{wrapfigure}%

\figref{fig:voxel-retrieval}  shows how to retrieve a feature for a given 3D point $\localquery_i$ from the volume by projecting to each axis and interpolating,
\begin{equation}
    \voxelfeat^x_i=\bonefvolumexi\left[\bonefscalex_i\projected{x}\right]\in\mR^{H},%
\label{eq:voxel-projection}
\end{equation}
where $\bonefscalex_i$ is a learnable scaling factor to control the volume size along the x-axis, and $\bonefvolumexi\left[\cdot\right]$ returns the interpolated feature when the projected and scaled coordinate falls in $\left[-1,1\right]$, and $\mathbf{0}$ otherwise. The extraction for y and z axes follows the same procedure.

The GNN attaches one factorized volume to every bone $i$ and is computed only once for every pose. 
In~\secref{sec:exp-ablation}, we show that the factorized volumes compare favorably against full 3D volumes on short video sequences with sparse or single views, while having 2x lower parameter counts.

\subsection{Stage II: Global Feature Aggregation}
\label{sec:feat-agg}
Given a query location $\query\in\mR^{3}$ in global coordinates, the corresponding voxel feature can be retrieved by first mapping the 3D points to the bone-relative space of $i$ via the world-to-bone coordinates transformation $T(\jrot_i)$,
\begin{equation}
    \begin{bmatrix}
    \localquery_i \\ 1 
    \end{bmatrix} = T(\jrot_i) 
    \begin{bmatrix}
    \query \\ 1 
    \end{bmatrix},
\end{equation}
and then retrieving the factorized features with equation Eq.~\ref{eq:voxel-projection}. However, multiple volumes can overlap.

\paragraph{Windowed bounds.} To facilitate learning volume dimensions $s_x$ that adapt to the body shape and to mitigate seam artifacts, we apply a window function  
\begin{equation}
     w_i=\exp(-\alpha(\projected{x}^\beta+\projected{y}^\beta+\projected{z}^\beta))
 \end{equation}
 that attenuates the feature value $\voxelfeat_i = w_i\left[\voxelfeat^x_i,\voxelfeat^y_i,\voxelfeat^z_i\right]$ for $\localquery_i$ towards the boundary of the volume, with $\alpha=2$ and $\beta=6$ similar to~\cite{Lombardi21mixture}. Still, multiple volumes will overlap near joints and when body parts are in contact. Moreover, the overlap changes with varying skeleton pose, demanding for an explicit aggregation step.

\paragraph{Voxel Aggregation Network.}
\label{sec:method-van}
Since an $\query$ that is close to the body falls into multiple volumes, we employ a voxel aggregation network $\aggnet$ to decide which per-bone voxel features to pass on to the downstream neural field for rendering. We explore several strategies, and conduct ablation studies on each of the options. Our aggregation network $\aggnet$ consists of a graph layer followed by per-node 2-layer MLPs with a small network width (32 per layer). We predict the weight $\aggweight_i$ for the feature retrieved from bone $i$ and compute the aggregated features via
\begin{equation}
\aggweight_i = \aggnet_i(\voxelfeat_i), \text{~and aggregated feature~}\aggvoxelfeat=\sum_{i=1}^{24}p_i\voxelfeat_i.
\end{equation}

Below, we discuss the three strategies for computing the aggregation weights.

\paragraph{Concatenate.} Simply concatenating all features lets the network disentangle individual factors, which is prone to overfitting as no domain knowledge is used.

\paragraph{Softmax-OOB.} Instead of simply using Softmax to obtain sparse and normalized weights as in~\cite{deng2019nasa,noguchi2021narf}, we can make use of our volume representation to remove the influence of irrelevant bones
\begin{equation}
        \aggweight_i=\frac{(1-\oob_i)\exp(\agglogit_i)}{\sum_{j=1}^{24}(1-\oob_i)\exp(\agglogit_j)},
\end{equation}
where $\oob_i$ is the out-of-bound (OOB) indicator which equals to $1$ when $\localquery_i$ is not inside of $\bonevolume_i$. The potential caveat is that $\aggvoxelfeat$ is still susceptible to features from irrelevant volumes. 
For instance, \figref{fig:agg-strat-comp} shows that Softmax-OOB produces artifacts when the hand gets close to the chest.

\newlength\aggstratscale
\setlength\aggstratscale{0.29\textwidth}
\setlength{\intextsep}{0pt}%
\begin{wrapfigure}{r}{\aggstratscale}
\centering
\setlength{\fboxrule}{0pt}%
\parbox[t]{0.485\aggstratscale}{%
\centering%
\fbox{\includegraphics[width=0.25\aggstratscale,trim=195 260 235 130,clip]{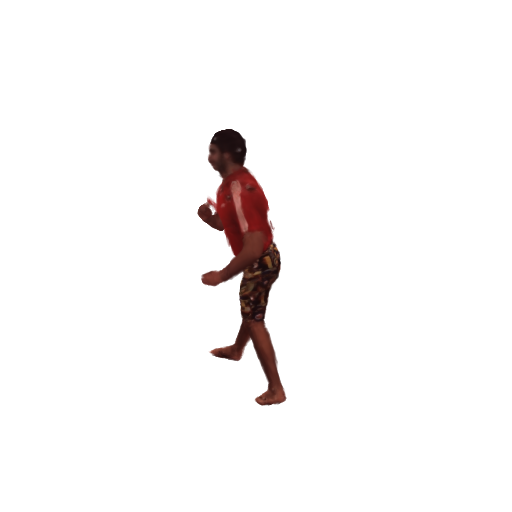}%
}\\%
{\scriptsize SoftmaxOOB}%
}
\hfill%
\parbox[t]{0.485\aggstratscale}{%
\centering%
\fbox{\includegraphics[width=0.25\aggstratscale,trim=195 260 235 130,clip]{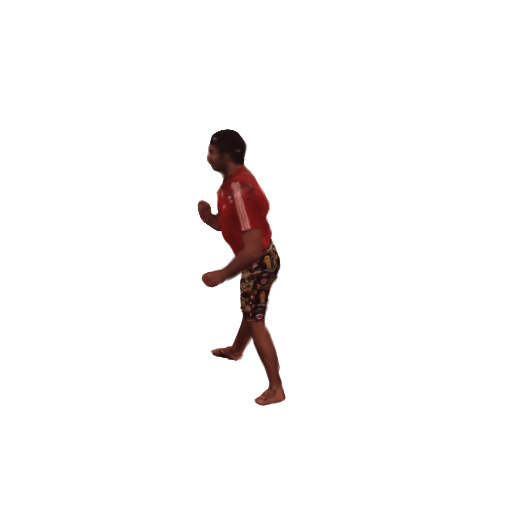}%
}\\%
{\scriptsize Soft-softmax}%
}%
\caption{\small The influence of aggregation strategies.}
\label{fig:agg-strat-comp}
\end{wrapfigure}

\paragraph{Soft-softmax} Due to the design of $\aggnet$, the output logit $\agglogit_i$ of bone $i$ is only dependent on itself. We can leverage this design to obtain the weight for each $\bonevolume_i$ independently and normalize their range to $[0,1]$ with a sigmoid function,
\begin{equation}
    \aggweight_i=(1-\oob_i)\cdot S(\agglogit_i),~\text{where~} S=\frac{1}{1+\exp(-\agglogit_i)}.
\end{equation}
To nevertheless ensure that aggregated features are in the same range irrespectively of the number of contributors, we introduce a \emph{soft-softmax} constraint 
\begin{equation}
    L_{\text{s}}=\sum_{\query}\left(\sum_{i=1}^{24} (1-\oob_i) p_i - l_{\query}\right)^2,
\end{equation}
that acts as a soft normalization factor opposed to the hard normalization in softmax. By setting $l_\query=1$ if $T_\query\cdot\density_\query > 0$ and 0 otherwise, the loss enforces the sum of weights of the activated bones to be close to $1$ when the downstream neural field has positive density prediction $\sigma$ (e.g., when $\query$ belong to the human body), and $0$ otherwise. The results is a compromise between an unweighted sum and softmax that attained the best generalization in our experiments. A representative improvement on softmax is shown in \figref{fig:agg-strat-comp}-right.

\subsection{Stage III: Neural Field and Volume Rendering}
\label{sec:method-nerf}
The aggregated features $\aggvoxelfeat$ contain the coarse, pose-dependent body features at location $\query$. 
To obtain high-quality human body, we learn a neural field $\neuralfield$ to predict the refined radiance $\radiance$ and density $\density$ for $\query$
\begin{equation}
    (\radiance,\density)=\neuralfield(\aggvoxelfeat,\viewdir),
\end{equation}
where $\viewdir\in\mR^{2}$ is the view direction. We can then render the image of the human subject by volume rendering as in the original NeRF~\cite{mildenhall2020nerf},
\begin{align}
\render(u,v;\pose) &= \sum^Q_{q=1} T_q(1-\exp(-\density_q\delta_q))\radiance_q,\quad T_q=\exp\left(-\sum_{j=1}^{q-1}\density_j\delta_j\right).
\label{eq:image fromation}
\end{align}
Given the pose $\pose$, the predicted image color at the 2D pixel location $\render(u,v;\pose)$ is computed by integrating the predicted color $\radiance_q$ of the $Q$ 3D samples along $\viewdir$. 
$\delta_q$ is the distance between neighboring samples, and $T_q$ represents the accumulated transmittance at sample $q$. 

\subsection{Training}
\label{sec:method-training}

Our model is directly supervised with ground truth images via photometric loss
\begin{equation}
    L_{\text{p}}=\sum_{(u,v)\in\image}\vert\render(u,v;\pose) - \image(u,v;\pose)\vert.
\end{equation}
We use L1 loss to avoid overfitting to appearance changes that cannot be explained by pose deformation alone. 
To prevent the per-bone volumes from growing too large and taking over other volumes, we employ a volume loss on the scaling factors as in ~\cite{Lombardi21mixture}
\begin{equation}
    L_{\text{v}} = \sum_{i=1}^{24}(\bonefscalex\cdot\bonefscaley\cdot\bonefscalez).
\end{equation}
For Soft-softmax in~\secref{sec:method-van}, we further regularize the output weights via the self-supervised loss $L_{\text{s}}$.

To summarize, the training objective of our approach is
\begin{equation}
    L=L_{\text{p}}+\lambda_{\text{v}} L_{\text{v}} + \lambda_{\text{s}} L_{\text{s}}.
\end{equation}
We set both $\lambda_{\text{v}}$ and $\lambda_{\text{s}}$ to $0.001$ for all experiments. See the supplemental for more implementation details.

\section{Experiments}
\label{sec:experiments}
In the following, we evaluate the improvements upon the most recent surface-free neural body model A-NeRF~\cite{su2021anerf}, and compare against recent model-based solutions NeuralBody~\cite{peng2020neuralbody} and Anim-NeRF~\cite{peng2021animatable}. %
An ablation study further quantifies the improvement of using the proposed aggregation function, local GNN features, and factorized volumes over simpler and more complex~\cite{Lombardi21mixture} alternatives, including the effects on model capacity.
The supplemental materials provide additional quantitative and qualitative results, including videos of retargeting applications.

\parag{Metrics and protocols.} Our goal is to analyze the quality of synthesizing novel views and separately the rendering of previously unseen poses. %
We quantify improvements by PSNR, SSIM\cite{wang2004ssim}, and perceptual metrics KID~\cite{binkowski2018demystifying,parmar2021cleanfid} and LPIPS~\cite{zhang2018perceptual} that are resilient to slight misalignments. All scores are computed over frames withheld from training: (1) Novel view synthesis is evaluated on multi-view datasets by learning the body model from a subset of cameras with the remaining ones used as the test set, i.e. rendering the same pose from the unseen view, and (2) novel pose synthesis quality is measured by training on the first part of a video and testing on the latter frames, given their corresponding 3D pose as input. This assumes that only the pose changes as the person moves. Hence, view-dependent illumination changes in (1) but stays similar in (2).

As the background image is not our focus, we report scores on tight bounding boxes either provided by the dataset or computed from the 3D poses.
\newlength\qnvpscale
\setlength\qnvpscale{0.160\linewidth}
\ifblurface
\newcommand{\qnvppost}{_blur}
\else
\newcommand{\qnvppost}{}
\fi
\newcommand{\qnvppath}{static/figure/experiments/h36m_novelview/}
\newcommand{\qnvpaps}{S5/} %
\newcommand{\qnvpapi}{009\qnvppost}
\newcommand{\qnvpbps}{S7/} %
\newcommand{\qnvpbpi}{009\qnvppost}
\newcommand{\qnvpcps}{S11/} %
\newcommand{\qnvpcpi}{007\qnvppost}
\begin{figure*}[t]
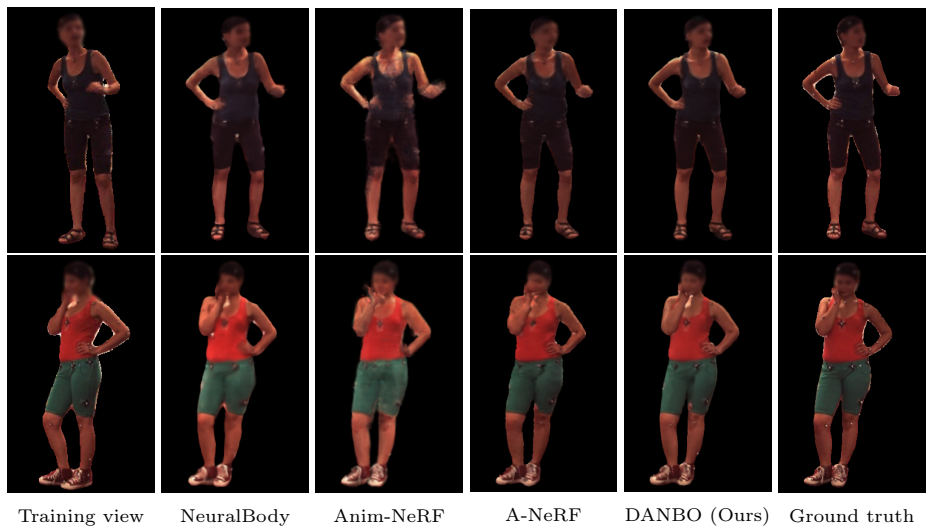

\setlength{\fboxrule}{0pt}%
\setlength{\fboxsep}{0pt}%
\parbox[t]{\qnvpscale}{%
\centering%
\fbox{\includegraphics%
[width=\qnvpscale,trim=551 430 258 250,clip]%
{\qnvppath\qnvpaps GTv_p\qnvpapi}%
}\\%
\fbox{\includegraphics%
[width=\qnvpscale,trim=466 430 336 250,clip]%
{\qnvppath\qnvpbps GTv_p\qnvpbpi}}\\%
{\scriptsize Training view}%
}%
\hfill%
\parbox[t]{\qnvpscale}{%
\centering %
\fbox{\includegraphics%
[width=\qnvpscale,trim=392 430 367 170,clip]%
{\qnvppath\qnvpaps NeuralBody_p\qnvpapi}%
}\\%
\fbox{\includegraphics%
[width=\qnvpscale,trim=342 410 397 170,clip]%
{\qnvppath\qnvpbps NeuralBody_p\qnvpbpi}%
}\\%
{\scriptsize  NeuralBody}%
}%
\hfill%
\parbox[t]{\qnvpscale}{%
\centering %
\fbox{\includegraphics%
[width=\qnvpscale,trim=392 430 367 170,clip]%
{\qnvppath\qnvpaps Anim-NeRF_p\qnvpapi}%
}\\%
\fbox{\includegraphics%
[width=\qnvpscale,trim=342 410 397 170,clip]%
{\qnvppath\qnvpbps Anim-NeRF_p\qnvpbpi}%
}\\%
{\scriptsize  Anim-NeRF}%
}%
\hfill%
\parbox[t]{\qnvpscale}{%
\centering %
\fbox{\includegraphics%
[width=\qnvpscale,trim=392 430 367 170,clip]%
{\qnvppath\qnvpaps A-NeRF_p\qnvpapi}%
}\\%
\fbox{\includegraphics%
[width=\qnvpscale,trim=342 410 397 170,clip]%
{\qnvppath\qnvpbps A-NeRF_p\qnvpbpi}%
}\\%
{\scriptsize  A-NeRF}%
}%
\hfill%
\parbox[t]{\qnvpscale}{%
\centering%
\fbox{\includegraphics%
[width=\qnvpscale,trim=392 430 367 170,clip]%
{\qnvppath\qnvpaps Ours_p\qnvpapi}%
}\\%
\fbox{\includegraphics%
[width=\qnvpscale,trim=342 410 397 170,clip]%
{\qnvppath\qnvpbps Ours_p\qnvpbpi}%
}\\%
{\scriptsize \ourapproach{} (Ours)}%
}%
\hfill%
\parbox[t]{\qnvpscale}{%
\centering%
\fbox{\includegraphics%
[width=\qnvpscale,trim=392 430 367 170,clip]%
{\qnvppath\qnvpaps GT_p\qnvpapi}%
}\\%
\fbox{\includegraphics%
[width=\qnvpscale,trim=342 410 397 170,clip]%
{\qnvppath\qnvpbps GT_p\qnvpbpi}}\\%
{\scriptsize  Ground truth}%
}%
\centering%
\caption{\textbf{Novel-view synthesis results on Human3.6M~\cite{Ionescu14b}.}~\ourapproach{} renders more complete limbs and clearer facial features than the baselines.
}
\label{fig:exp-novel-view}
\end{figure*}
\begin{table}[!t]
\caption{\textbf{Novel-view synthesis comparisons on Human3.6M~\cite{Ionescu14b}}. The disentangled feature enables~\ourapproach{} to achieve better novel view synthesis.
}
\centering
\resizebox{1.0\linewidth}{!}{
\setlength{\tabcolsep}{3pt}
\begin{tabular}{lcccccccccccccccc}
\toprule
& \multicolumn{4}{c}{NeuralBody~\cite{peng2020neuralbody}}  & \multicolumn{4}{c}{Anim-NeRF~\cite{peng2021animatable}} & \multicolumn{4}{c}{A-NeRF~\cite{su2021anerf}} & \multicolumn{4}{c}{\tbf{\ourapproach{} (Ours)}} \\
 \midrule
& PSNR~$\uparrow$& SSIM~$\uparrow$& KID~$\downarrow$& LPIPS~$\downarrow$& PSNR~$\uparrow$& SSIM~$\uparrow$& KID~$\downarrow$& LPIPS~$\downarrow$& PSNR~$\uparrow$& SSIM~$\uparrow$& KID~$\downarrow$& LPIPS~$\downarrow$& PSNR~$\uparrow$& SSIM~$\uparrow$& KID~$\downarrow$& LPIPS~$\downarrow$\\
\rowcolor{Gray}
S1 & 22.88& 0.897& 0.048& 0.153& 22.74& 0.896& 0.106& 0.156& 23.93& 0.912& 0.042& 0.153& \tbf{23.95}& \tbf{0.916}& \tbf{0.033}& \tbf{0.148}\\
S5 & 24.61& 0.917& 0.033& 0.146& 23.40& 0.895& 0.087& 0.151& 24.67& 0.919& 0.036& 0.147& \tbf{24.86}& \tbf{0.924}& \tbf{0.029}& \tbf{0.142}\\
\rowcolor{Gray}
S6 & 22.83& 0.888& 0.050& 0.146& 22.85& 0.871& 0.113& 0.151& 23.78& 0.887& 0.051& 0.164& \tbf{24.54}& \tbf{0.903}& \tbf{0.035}& \tbf{0.143}\\
S7 & 23.17& 0.915& 0.043& 0.134& 21.97& 0.891& 0.054& 0.140& 24.40& 0.917& \tbf{0.025}& 0.139& \tbf{24.45}& \tbf{0.920}& 0.028& \tbf{0.131}\\
\rowcolor{Gray}
S8 & 21.72& 0.894& 0.071& 0.177& 22.82& 0.900& 0.095& 0.178& 22.70& 0.907& 0.086& 0.196& \tbf{23.36}& \tbf{0.917}& \tbf{0.068}& \tbf{0.173}\\
S9 & 24.29& 0.911& \tbf{0.035}& 0.141& 24.86& 0.911& 0.057& 0.145& 25.58& 0.916& 0.039& 0.150& \tbf{26.15}& \tbf{0.925}& 0.040& \tbf{0.137}\\
\rowcolor{Gray}
S11 & 23.70& 0.896& 0.080& 0.155& 24.76& 0.907& 0.077& 0.158& 24.38& 0.905& \tbf{0.057}& 0.164& \tbf{25.58}& \tbf{0.917}& 0.060& \tbf{0.153}\\
\midrule
Avg & 23.31& 0.903& 0.051& 0.150& 23.34& 0.896& 0.084& 0.154& 24.21& 0.909& 0.048& 0.159& \tbf{24.70}& \tbf{0.917}& \tbf{0.042}& \tbf{0.146} \\
    \bottomrule
\end{tabular}
\label{tab:exp-h36m-novel-view}
}
\end{table}

\parag{Datasets.}
We compare our~\ourapproach{} 
using the established benchmarks for neural bodies, covering indoor and outdoor, and single and multi-view settings:
\begin{itemize}
    \item \textbf{Human3.6M}\footnote{Meta did not have access to the Human3.6M dataset.}~\cite{Ionescu14b,Ionescu11}: We follow the same evaluation protocol as in Anim-NeRF~\cite{peng2021animatable}, with a total of 7 subjects for evaluation. The foreground maps are computed using~\cite{gong2018instance}.
    \item \textbf{MonoPerfCap}~\cite{xu2018monoperfcap} features multiple outdoor sequences, recorded using a single high-resolution camera. 
    We use the same two sequences and setting as in A-NeRF~\cite{su2021anerf}: Weipeng\_outdoor and Nadia\_outdoor with 1151 and 1635 images, respectively, of resolution $1080\times1920$. Human and camera pose is estimated by SPIN~\cite{kolotouros2019learning_spin} and refined with \cite{su2021anerf}. Foreground masks are obtained by DeepLabv3~\cite{chen2017rethinking}. 
\end{itemize}
We further include the challenging motion such as dancing and gymnastic poses from Mixamo~\cite{mixamo} and Surreal+CMU-Mocap dataset~\cite{varol17_surreal,CMUMOCAP} for motion retargeting (detailed in the supplemental). In total, we evaluate on 9 different subjects and 11 different sequences. %
\newlength\qnphmpscale
\setlength\qnphmpscale{0.0982\linewidth}
\ifblurface
\newcommand{\qnphmpost}{_blur}
\else
\newcommand{\qnphmpost}{}
\fi
\newcommand{\qnphmpath}{static/figure/experiments/h36m_unseen/}
\newcommand{\qnphmaps}{S1/} %
\newcommand{\qnphmapi}{000\qnphmpost}
\newcommand{\qnphmbps}{S6/} %
\newcommand{\qnphmbpi}{001\qnphmpost}
\newcommand{\qnphmcps}{S9/} %
\newcommand{\qnphmcpi}{004\qnphmpost}
\newcommand{\qnphmdps}{S11/} %
\newcommand{\qnphmdpi}{004\qnphmpost}
\begin{figure*}[t]
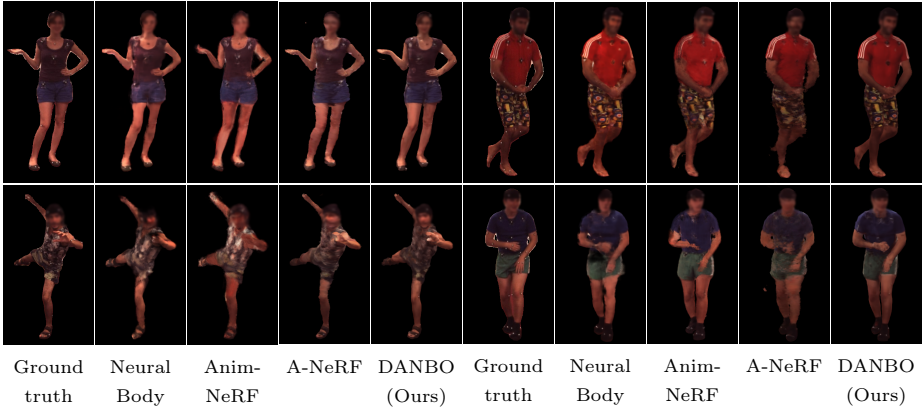

\setlength{\fboxsep}{0pt}%
\setlength{\fboxrule}{0pt}%
\parbox[t]{\qnphmpscale}{%
\centering%
\fbox{\includegraphics%
[width=\qnphmpscale,trim=340 375 410 125,clip]%
{\qnphmpath\qnphmaps GT_p\qnphmapi}%
}\\%
\fbox{\includegraphics%
[width=\qnphmpscale,trim=368 325 305 100,clip]%
{\qnphmpath\qnphmbps GT_p\qnphmbpi}}\\%
{\scriptsize Ground truth}%
}%
\hfill%
\parbox[t]{\qnphmpscale}{%
\centering %
\fbox{\includegraphics%
[width=\qnphmpscale,trim=340 375 410 125,clip]%
{\qnphmpath\qnphmaps NeuralBody_p\qnphmapi}%
}\\%
\fbox{\includegraphics
[width=\qnphmpscale,trim=368 325 305 100,clip]%
{\qnphmpath\qnphmbps NeuralBody_p\qnphmbpi}%
}\\%
{\scriptsize Neural Body}%
}%
\hfill%
\parbox[t]{\qnphmpscale}{%
\centering %
\fbox{\includegraphics%
[width=\qnphmpscale,trim=340 375 410 125,clip]%
{\qnphmpath\qnphmaps Anim-NeRF_p\qnphmapi}%
}\\%
\fbox{\includegraphics%
[width=\qnphmpscale,trim=368 325 305 100,clip]%
{\qnphmpath\qnphmbps Anim-NeRF_p\qnphmbpi}%
}\\%
{\scriptsize Anim-NeRF}%
}%
\hfill%
\parbox[t]{\qnphmpscale}{%
\centering %
\fbox{\includegraphics%
[width=\qnphmpscale,trim=340 375 410 124,clip]%
{\qnphmpath\qnphmaps A-NeRF_p\qnphmapi}%
}\\%
\fbox{\includegraphics%
[width=\qnphmpscale,trim=368 325 305 100,clip]%
{\qnphmpath\qnphmbps A-NeRF_p\qnphmbpi}%
}\\%
{\scriptsize A-NeRF}%
}%
\hfill%
\parbox[t]{\qnphmpscale}{%
\centering%
\fbox{\includegraphics%
[width=\qnphmpscale,trim=340 375 410 124,clip]%
{\qnphmpath\qnphmaps Ours_p\qnphmapi}%
}\\%
\fbox{\includegraphics%
[width=\qnphmpscale,trim=368 325 305 100,clip]%
{\qnphmpath\qnphmbps Ours_p\qnphmbpi}%
}\\%
{\scriptsize \ourapproach{} (Ours)}%
}%
\hfill%
\parbox[t]{\qnphmpscale}{%
\centering%
\fbox{\includegraphics%
[width=\qnphmpscale,trim=340 375 409 125,clip]%
{\qnphmpath\qnphmcps GT_p\qnphmcpi}}\\%
\fbox{\includegraphics%
[width=\qnphmpscale,trim=287 390 450 150,clip]%
{\qnphmpath\qnphmdps GT_p\qnphmdpi}}\\%
{\scriptsize Ground truth}%
}%
\hfill%
\parbox[t]{\qnphmpscale}{%
\centering %
\fbox{\includegraphics%
[width=\qnphmpscale,trim=340 375 409 125,clip]%
{\qnphmpath\qnphmcps NeuralBody_p\qnphmcpi}}\\%
\fbox{\includegraphics%
[width=\qnphmpscale,trim=287 390 450 150,clip]%
{\qnphmpath\qnphmdps NeuralBody_p\qnphmdpi}}\\%
{\scriptsize Neural Body}%
}%
\hfill%
\parbox[t]{\qnphmpscale}{%
\centering %
\fbox{\includegraphics%
[width=\qnphmpscale,trim=340 375 409 125,clip]%
{\qnphmpath\qnphmcps Anim-NeRF_p\qnphmcpi}}\\%
\fbox{\includegraphics%
[width=\qnphmpscale,trim=287 390 450 150,clip]%
{\qnphmpath\qnphmdps Anim-NeRF_p\qnphmdpi}}\\%
{\scriptsize Anim-NeRF}%
}%
\hfill%
\parbox[t]{\qnphmpscale}{%
\centering %
\fbox{\includegraphics%
[width=\qnphmpscale,trim=340 375 409 124,clip]%
{\qnphmpath\qnphmcps A-NeRF_p\qnphmcpi}}\\%
\fbox{\includegraphics%
[width=\qnphmpscale,trim=288 390 450 150,clip]%
{\qnphmpath\qnphmdps A-NeRF_p\qnphmdpi}}\\%
{\scriptsize A-NeRF}%
}%
\hfill%
\parbox[t]{\qnphmpscale}{%
\centering%
\fbox{\includegraphics%
[width=\qnphmpscale,trim=340 375 409 124,clip]%
{\qnphmpath\qnphmcps Ours_p\qnphmcpi}}\\%
\fbox{\includegraphics%
[width=\qnphmpscale,trim=288 390 450 150,clip]%
{\qnphmpath\qnphmdps Ours_p\qnphmdpi}}\\%
{\scriptsize  \ourapproach{} (Ours)}%
}%
\centering%
\caption{\textbf{Unseen pose synthesis on Human3.6M~\cite{Ionescu14b} test split}. Our disentangled representation enables~\ourapproach{} to generate plausible geometry and deformation for held-out testing poses, and achieve better visual quality than both surface-free and surface-based baselines. Note that, unlike Anim-NeRF~\cite{peng2021animatable}, we do not require test-time finetuning for unseen poses.
}%
\label{fig:exp-h36m-novel-pose}
\end{figure*}
\newlength\qrtpscale
\setlength\qrtpscale{0.16\linewidth}
\ifblurface%
\newcommand{\qrtpost}{_blur}
\else
\newcommand{\qrtpost}{}
\fi
\newcommand{\qrtpath}{static/figure/experiments/motion_retarget/}
\newcommand{\qrtpa}{ND}
\newcommand{\qrtpb}{WP}
\newcommand{\qrtpc}{S8}
\newcommand{\qrtpd}{S6}
\newcommand{\qrtpe}{S11}
\newcommand{\qrtpf}{S7}
\begin{figure}[!t]
\setlength{\fboxrule}{0pt}%
\setlength{\fboxsep}{0pt}%
\parbox[t]{\qrtpscale}{
\centering
\fbox{\includegraphics
[width=\qrtpscale,trim=150 100 120 130,clip]%
{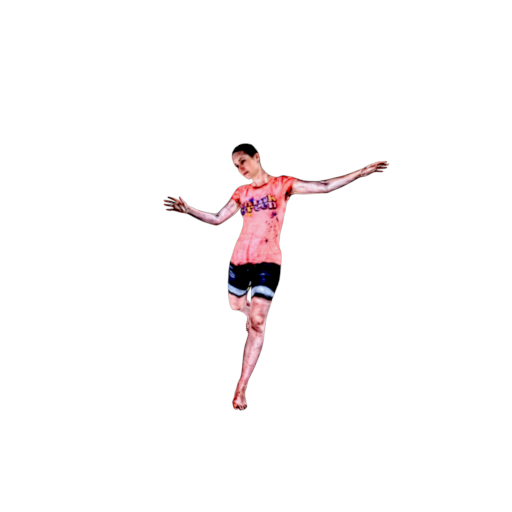}}\\%
\fbox{\includegraphics
[width=\qrtpscale,trim=120 120 190 150,clip]%
{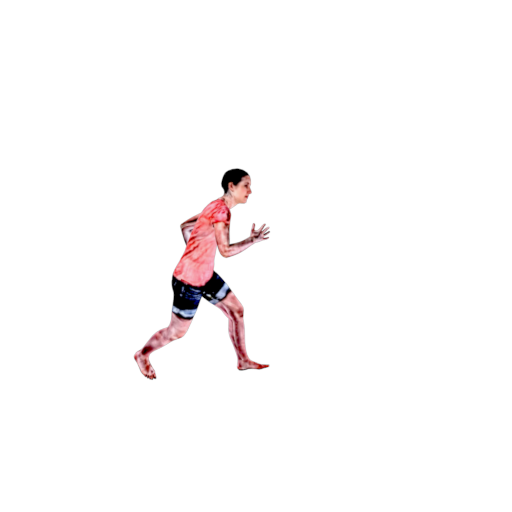}}\\%
\fbox{\includegraphics
[width=\qrtpscale,trim=400 260 300 340,clip]%
{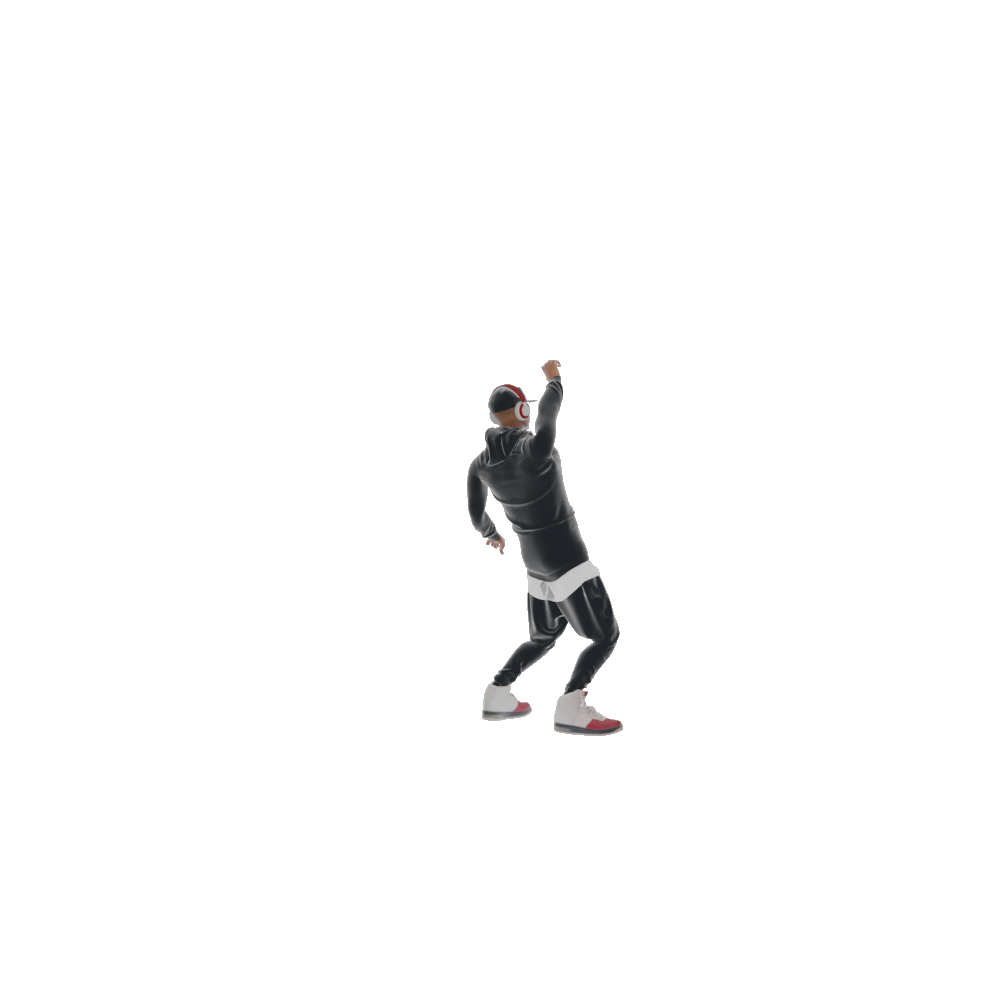}}\\%
\fbox{\includegraphics
[width=\qrtpscale,trim=350 290 350 345,clip]%
{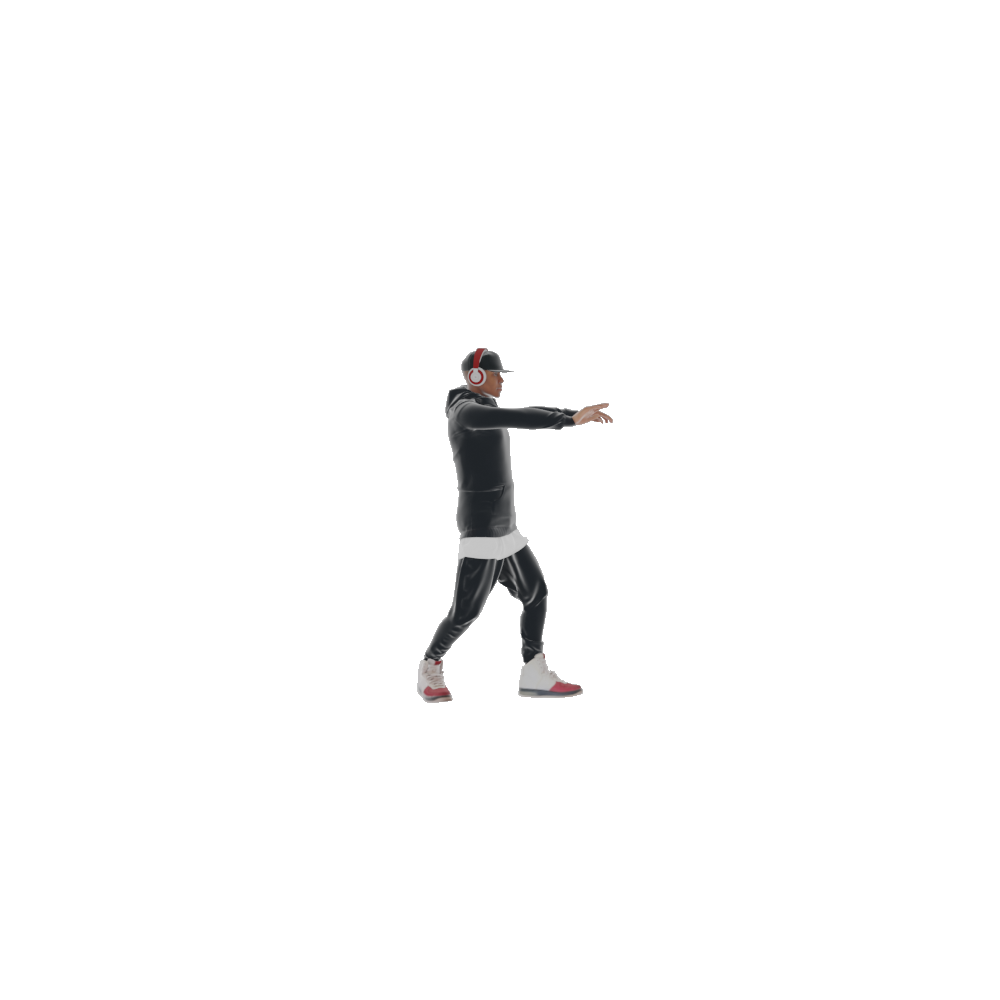}}\\%
\fbox{\includegraphics
[width=\qrtpscale,trim=470 290 270 400,clip]%
{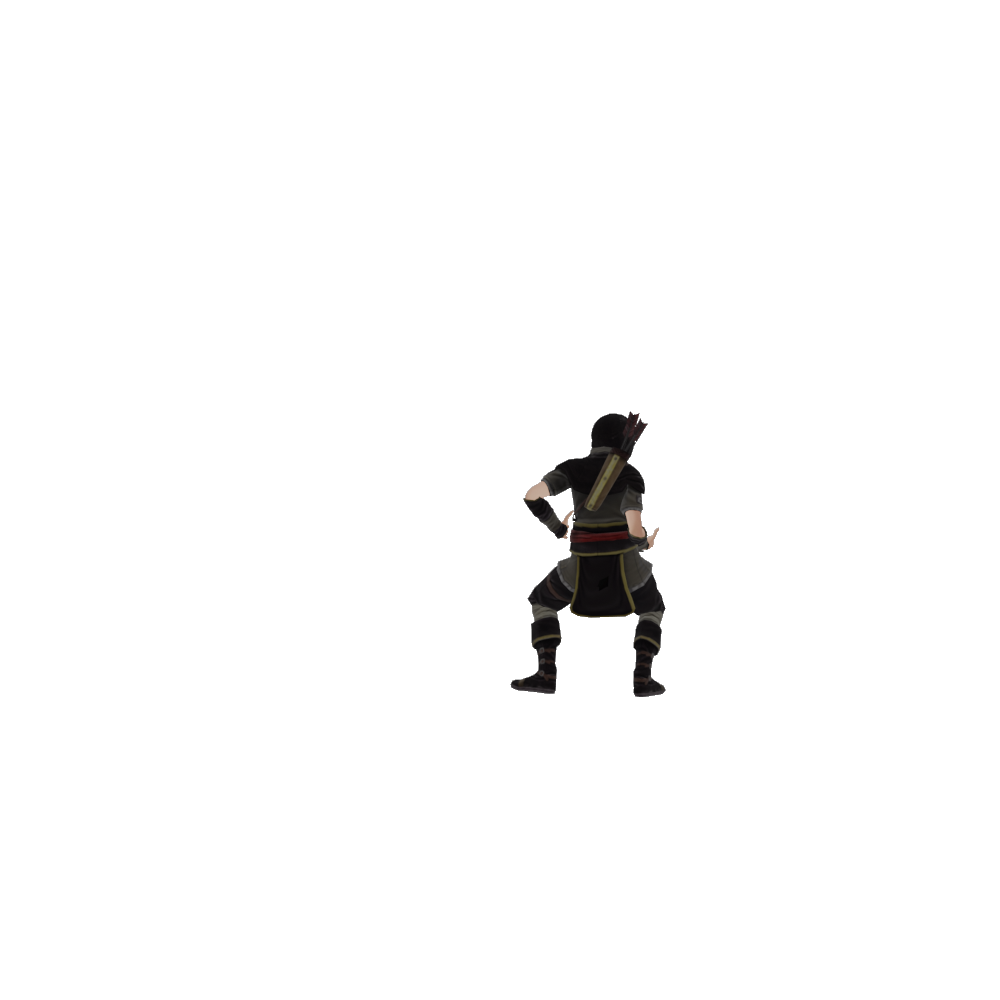}}\\%
{\scriptsize Source}
}%
\hfill%
\parbox[t]{\qrtpscale}{
\centering %
\fbox{\includegraphics
[width=\qrtpscale,trim=280 145 220 275,clip]%
{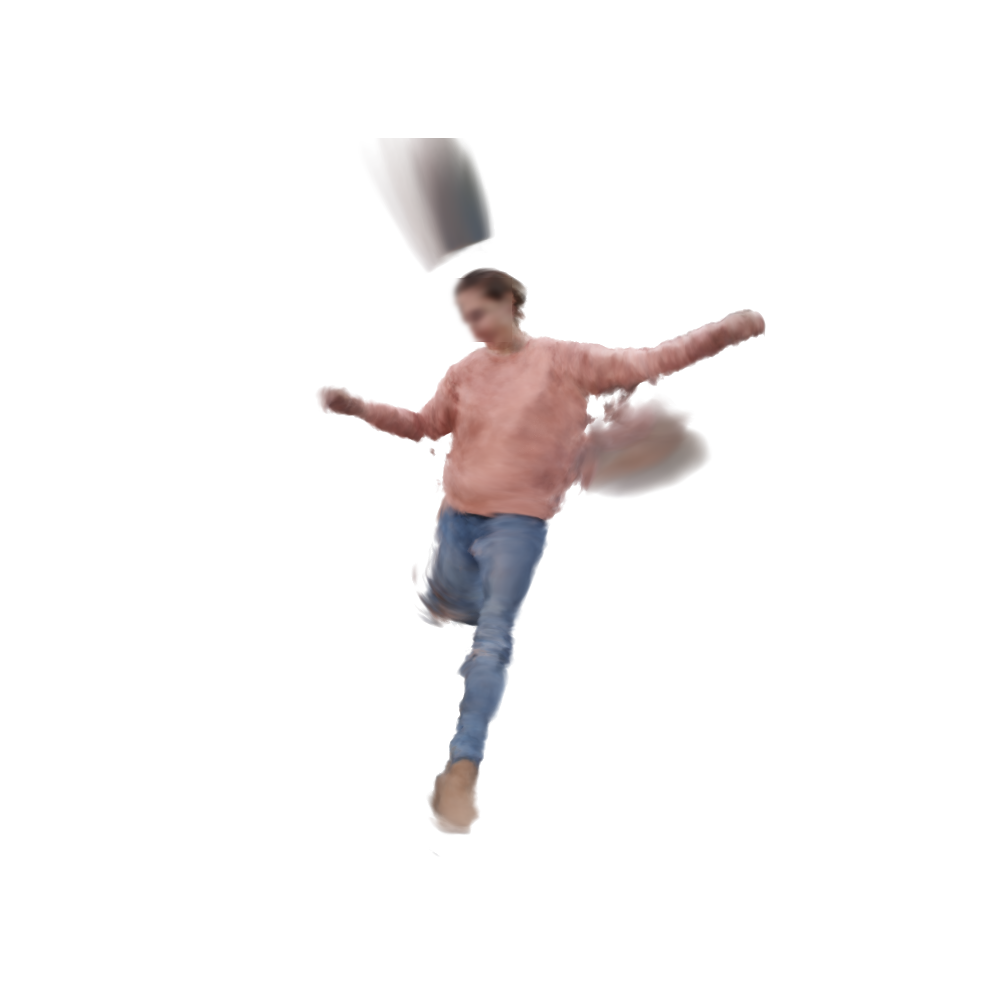}}\\%
\fbox{\includegraphics
[width=\qrtpscale,trim=240 240 380 300,clip]%
{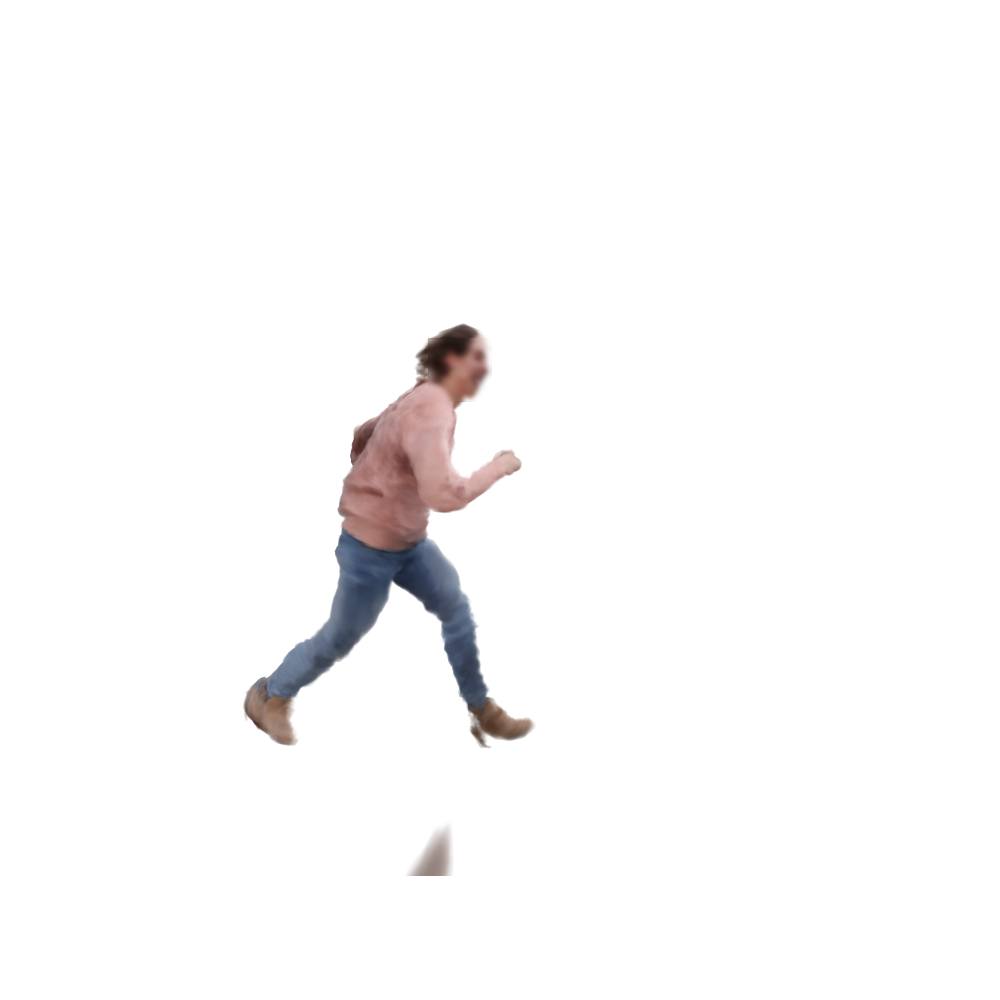}}\\%
\fbox{\includegraphics
[width=\qrtpscale,trim=400 260 300 345,clip]%
{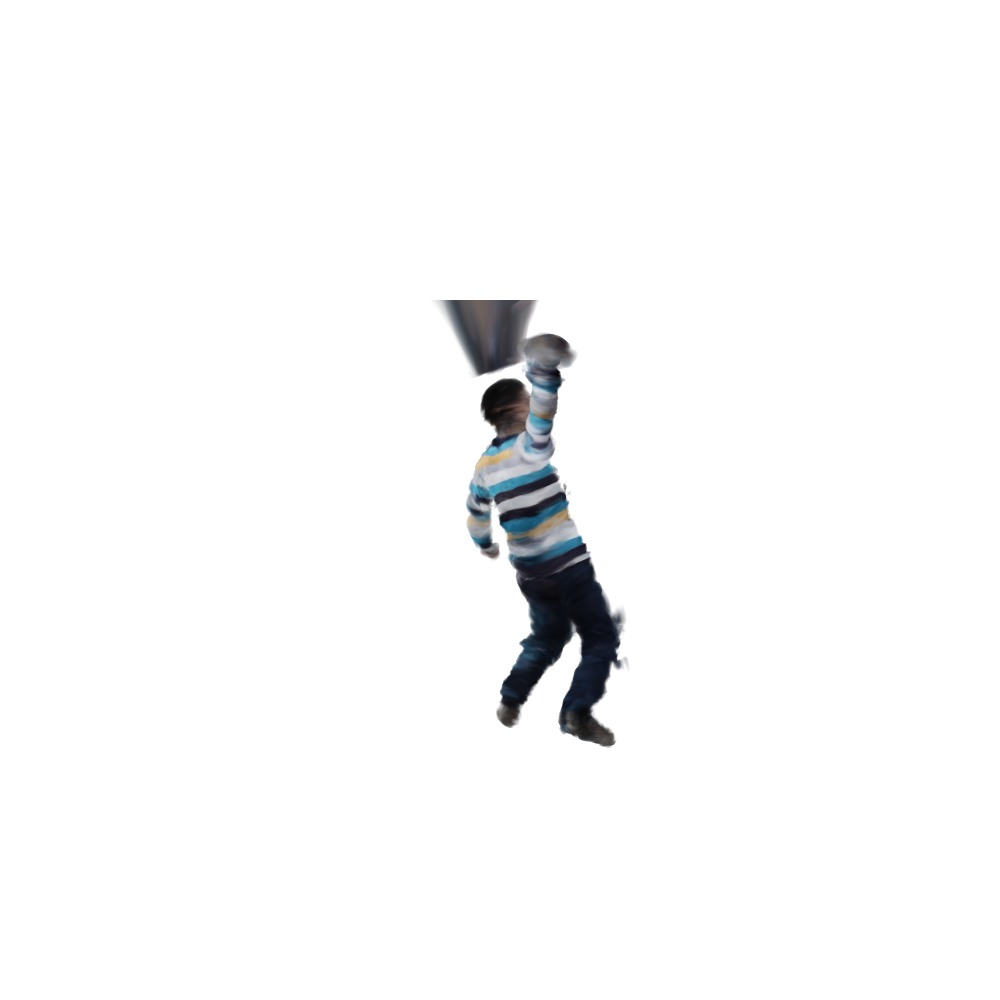}}\\%
\fbox{\includegraphics
[width=\qrtpscale,trim=350 290 350 345,clip]%
{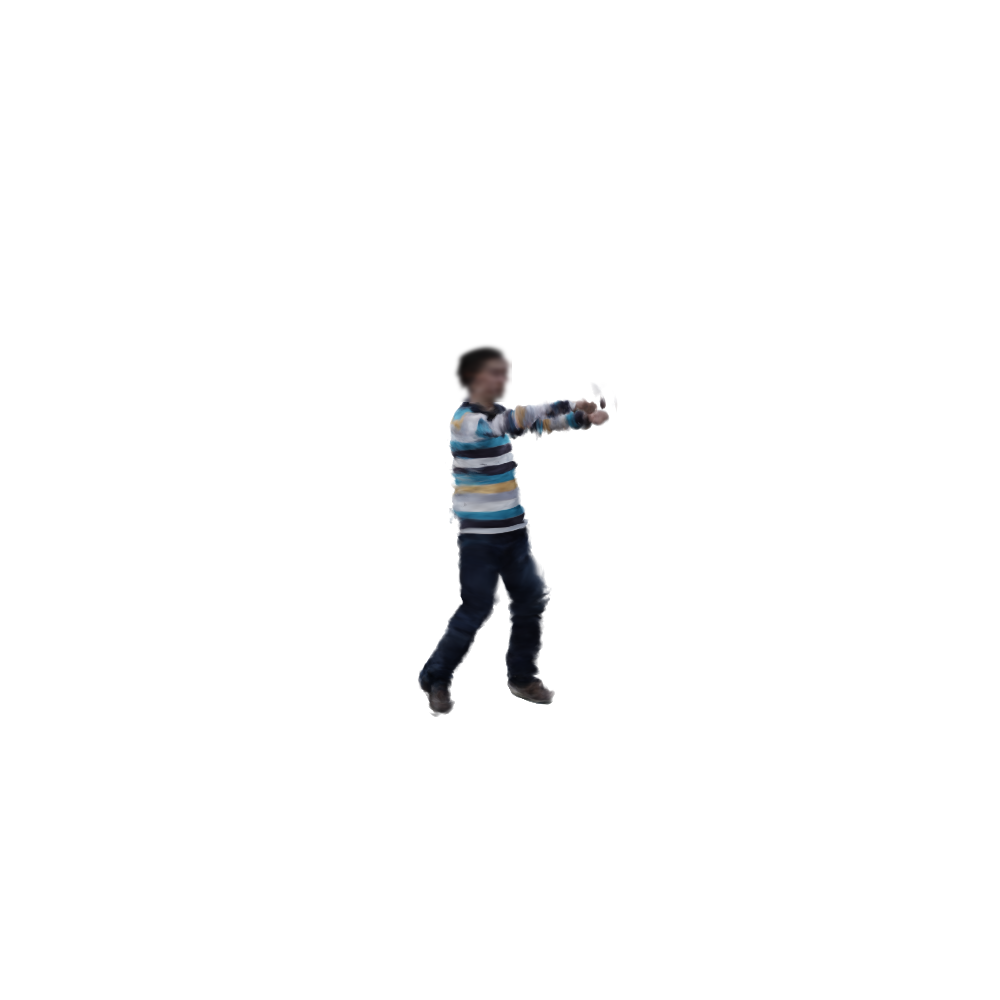}}\\%
\fbox{\includegraphics
[width=\qrtpscale,trim=475 320 275 380,clip]%
{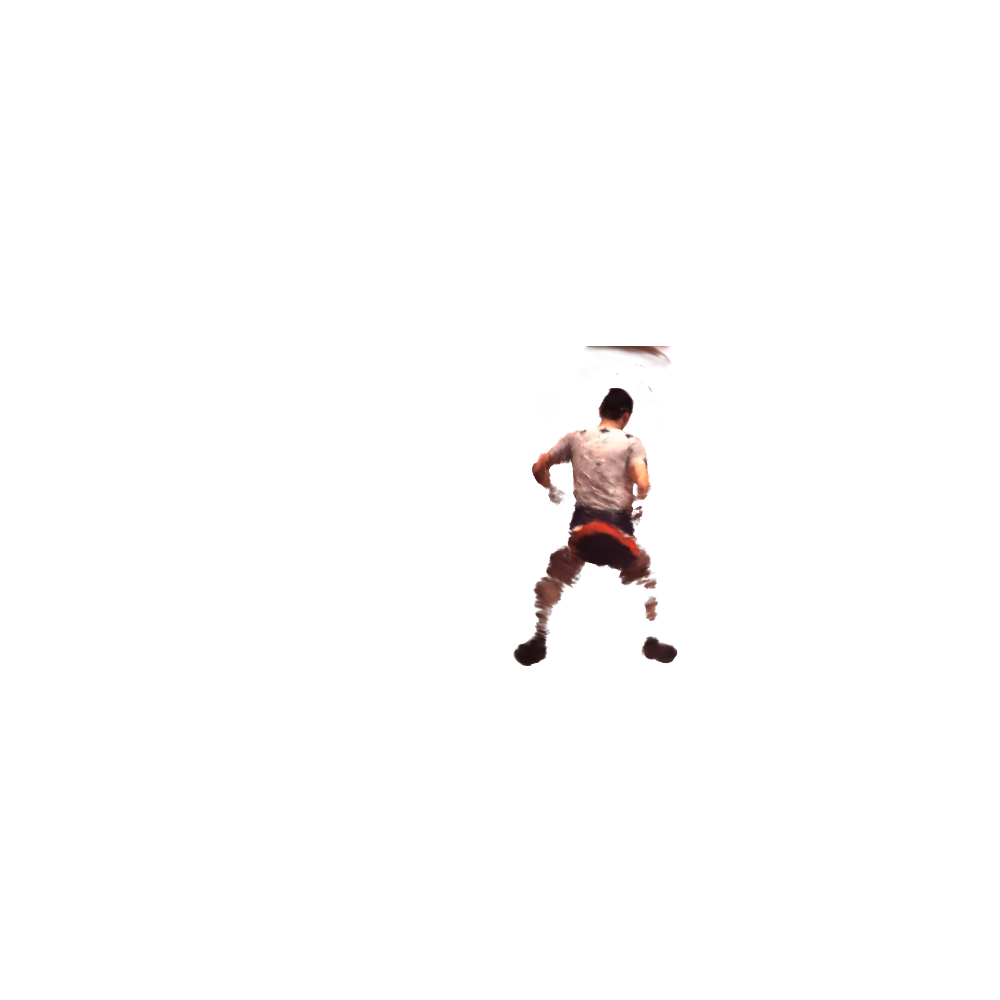}}\\%
{\scriptsize A-NeRF}%
}
\hfill%
\parbox[t]{\qrtpscale}{
\centering
\fbox{\includegraphics
[width=\qrtpscale,trim=280 145 220 275,clip]%
{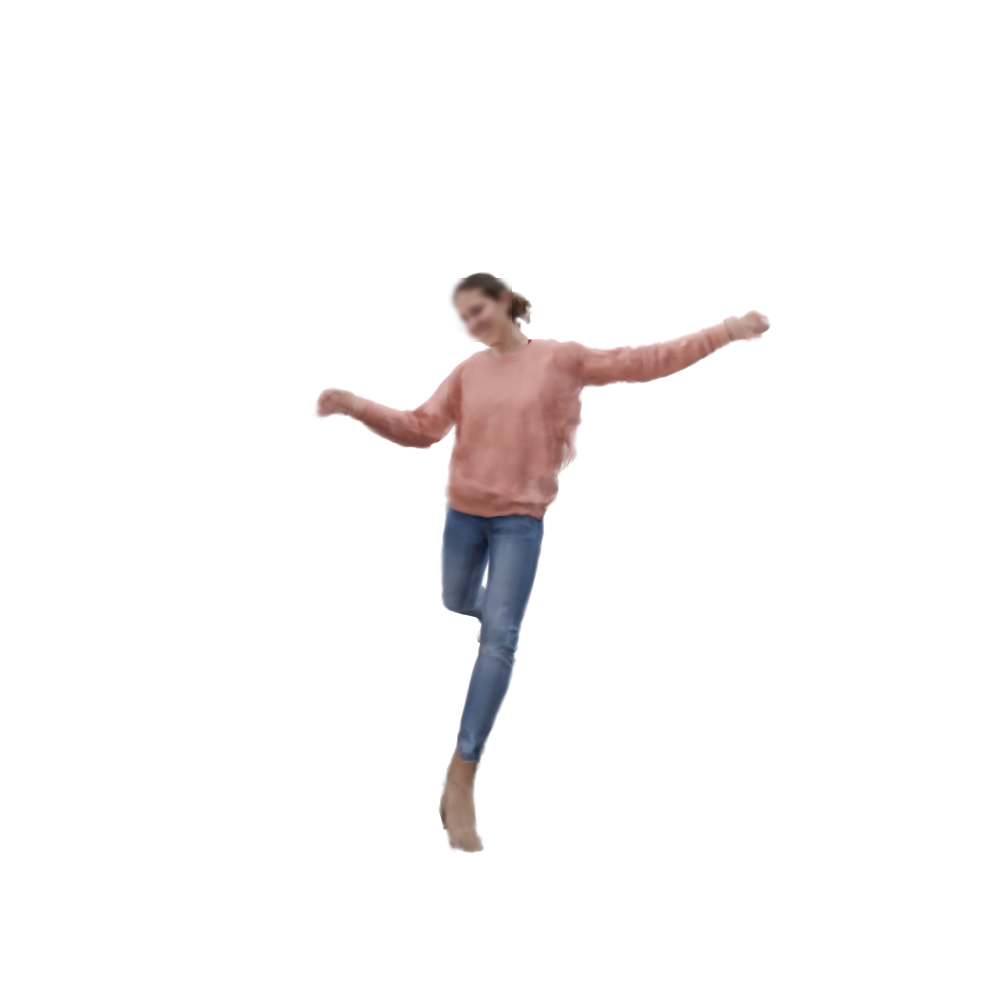}}\\%
\fbox{\includegraphics
[width=\qrtpscale,trim=240 240 380 300,clip]%
{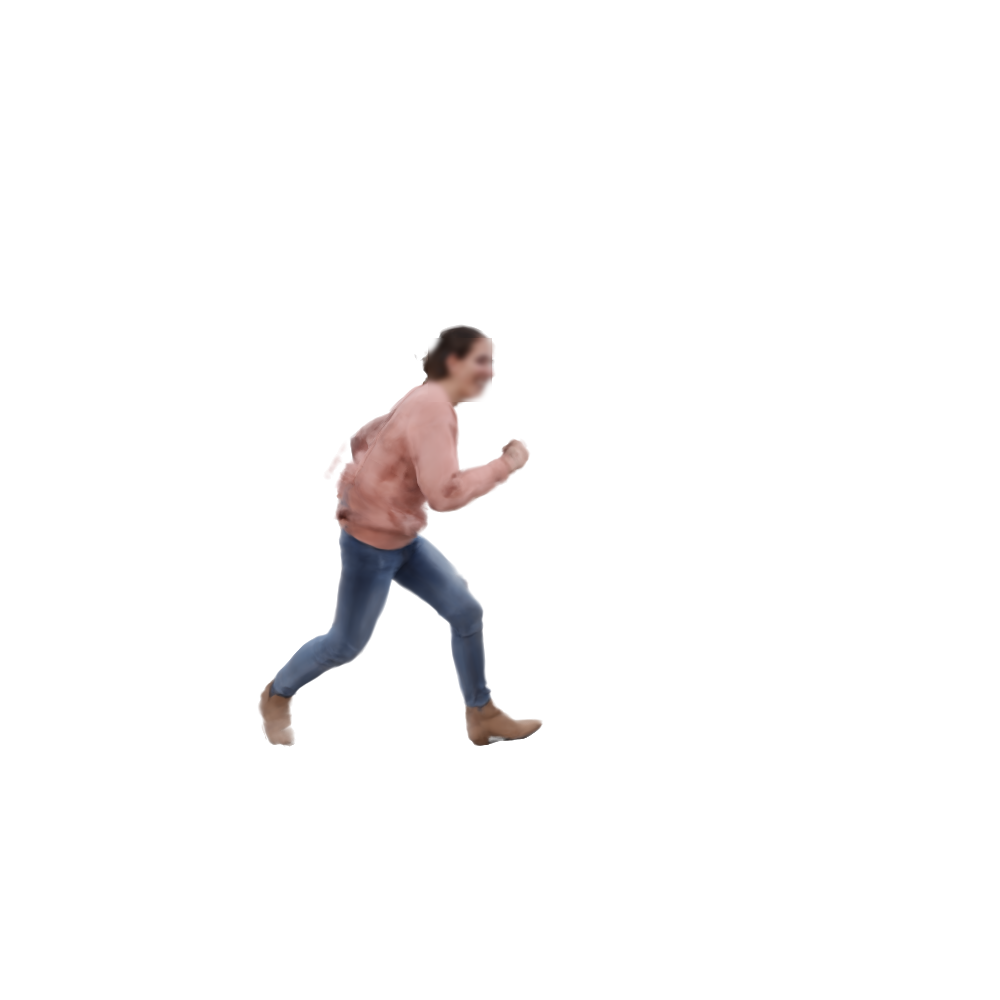}}\\%
\fbox{\includegraphics
[width=\qrtpscale,trim=400 260 300 345,clip]%
{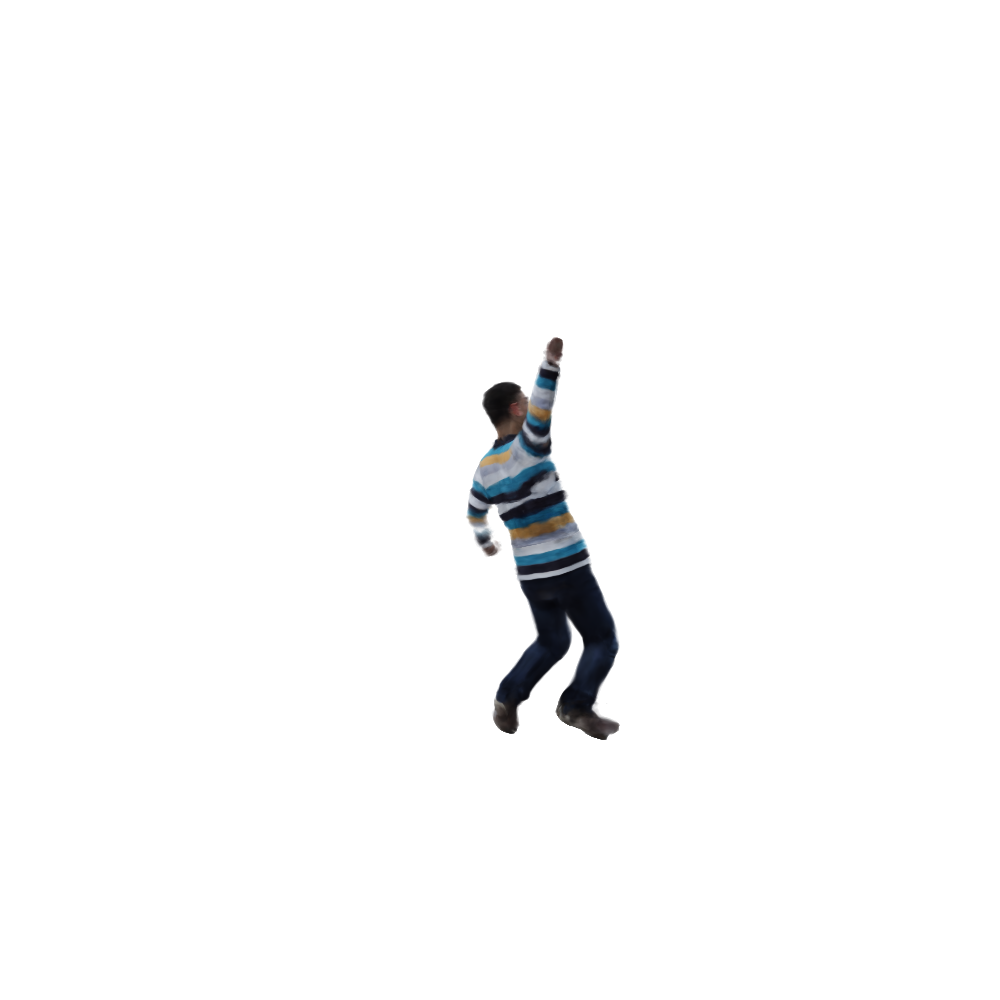}}\\%
\fbox{\includegraphics
[width=\qrtpscale,trim=350 290 350 345,clip]%
{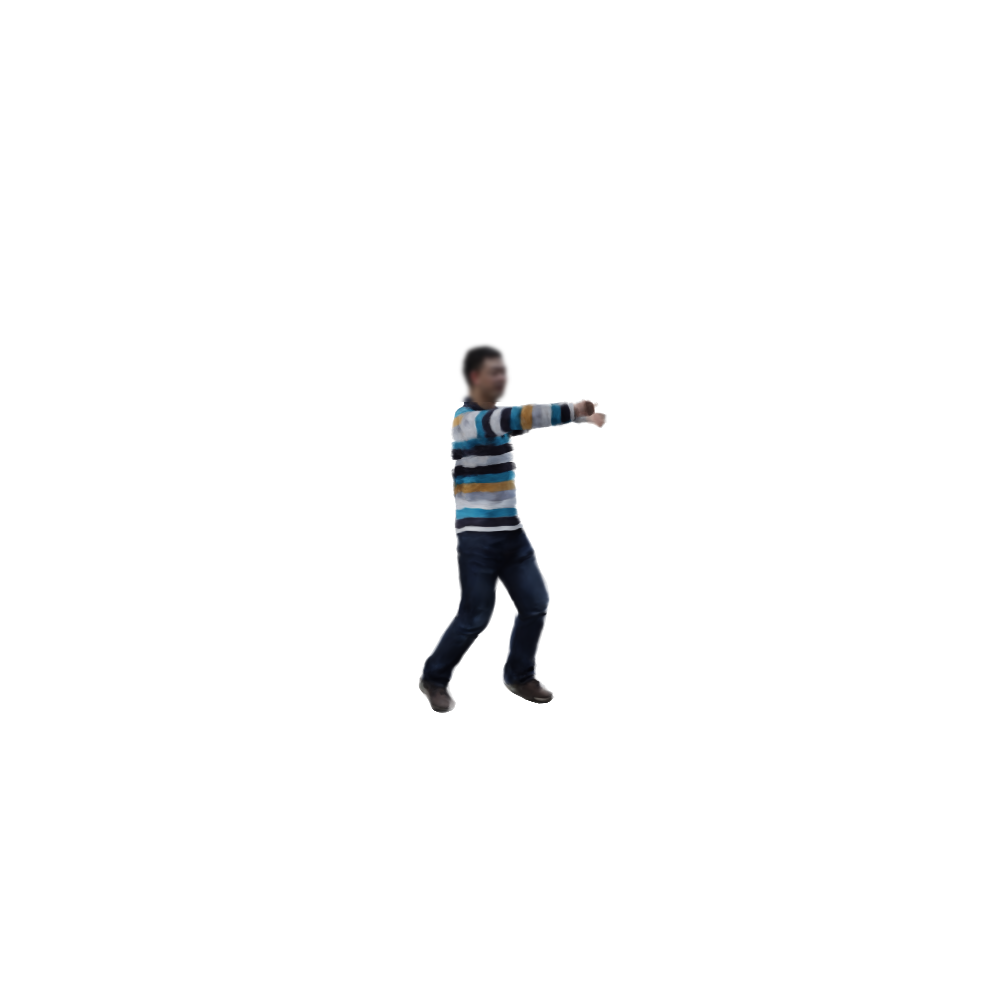}}\\%
\fbox{\includegraphics
[width=\qrtpscale,trim=475 320 275 380,clip]%
{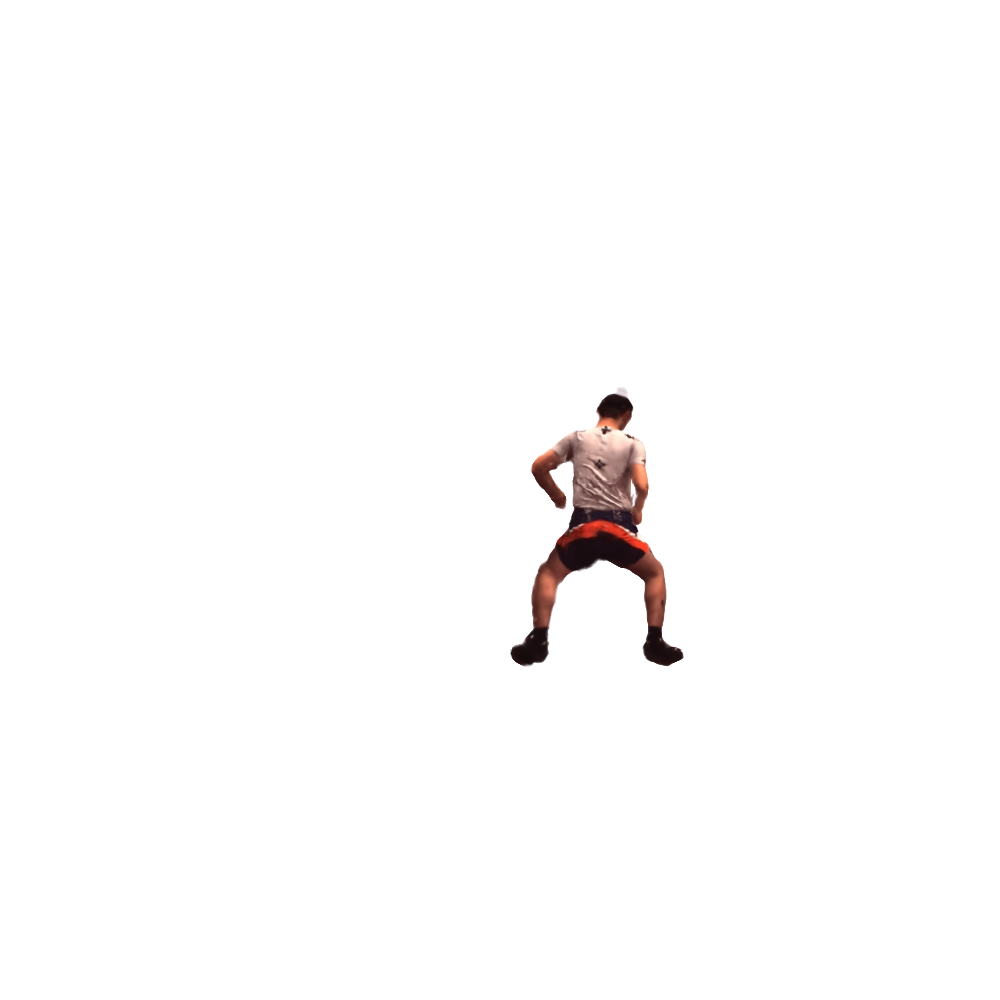}}\\%
{\scriptsize \ourapproach{} (Ours) }%
}%
\parbox[t]{\qrtpscale}{
\centering
\fbox{\includegraphics
[width=\qrtpscale,trim=180 100 100 150,clip]%
{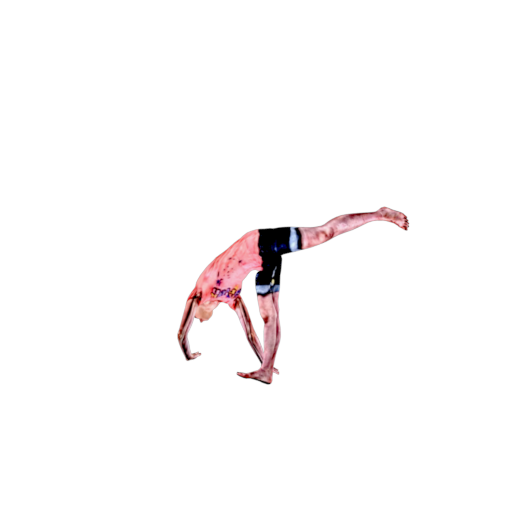}}\\%
\fbox{\includegraphics
[width=\qrtpscale,trim=210 160 150 160,clip]%
{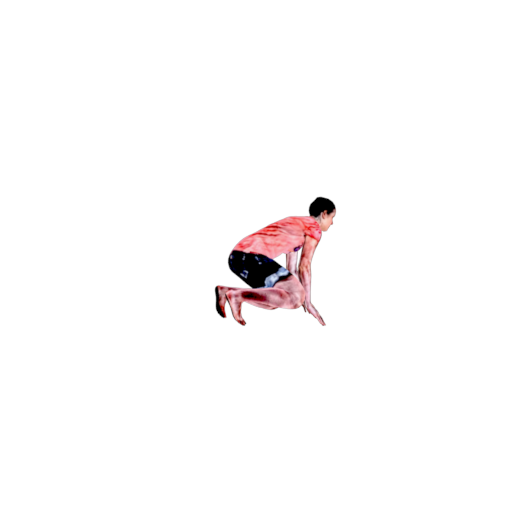}}\\%
\fbox{\includegraphics
[width=0.88\qrtpscale,trim=750 300 650 100,clip]%
{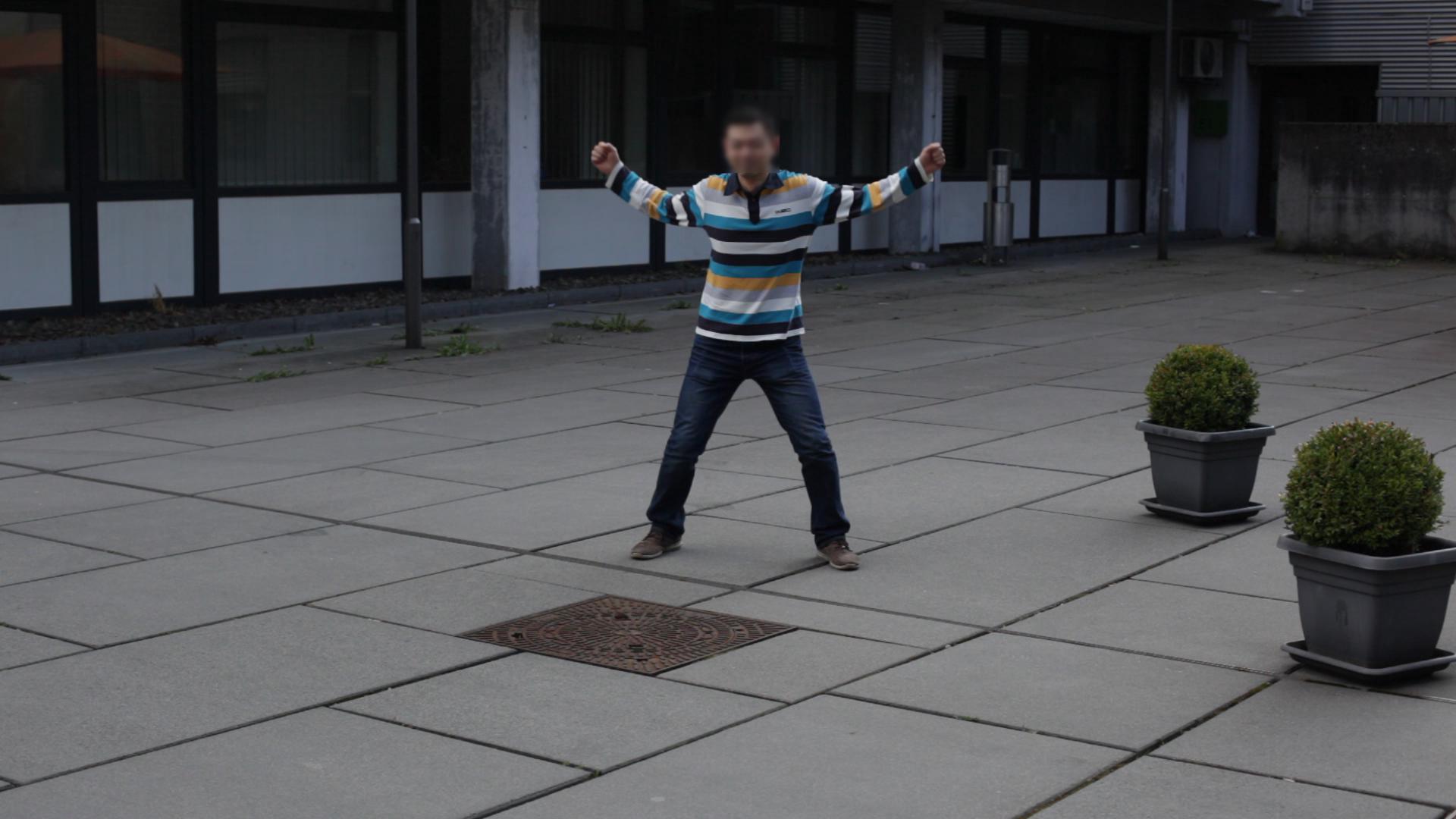}}\\%
\fbox{\includegraphics
[width=0.88\qrtpscale,trim=580 280 750 0,clip]%
{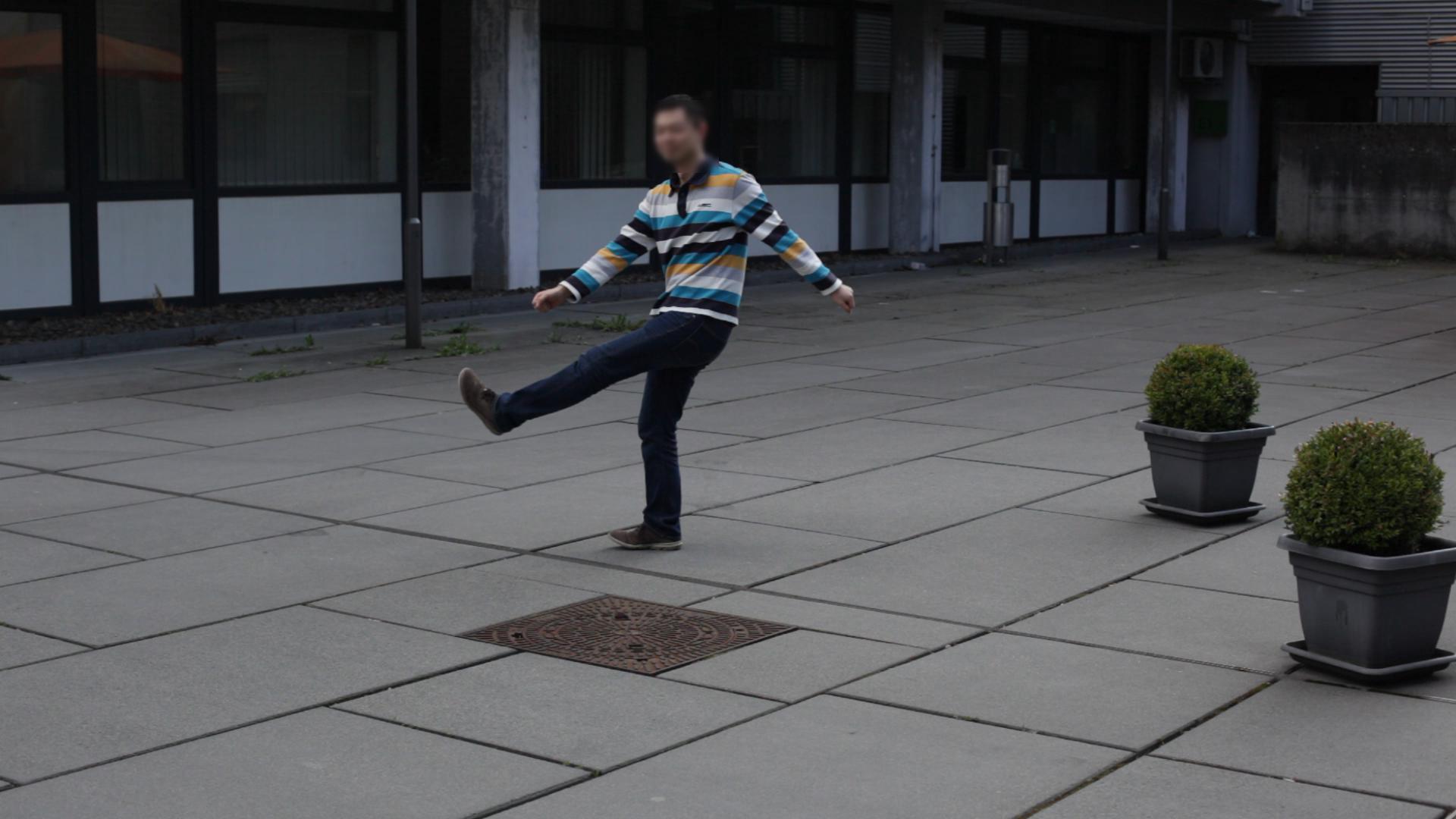}}\\%
\fbox{\includegraphics
[width=0.85\qrtpscale,trim=590 160 800 100,clip]%
{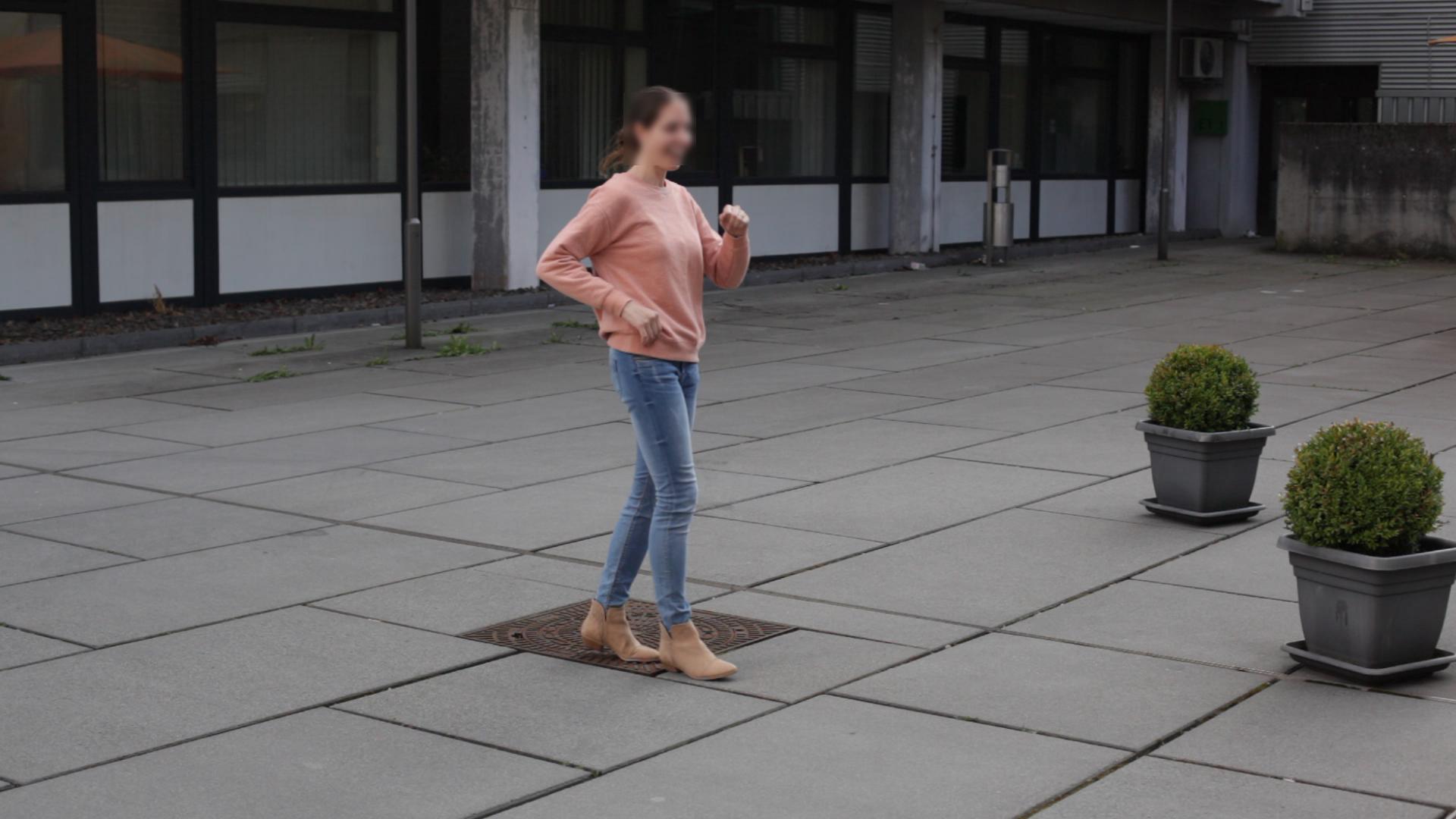}}\\%
{\scriptsize Source}
}%
\hfill%
\parbox[t]{\qrtpscale}{
\centering %
\fbox{\includegraphics
[width=\qrtpscale,trim=360 200 200 300,clip]%
{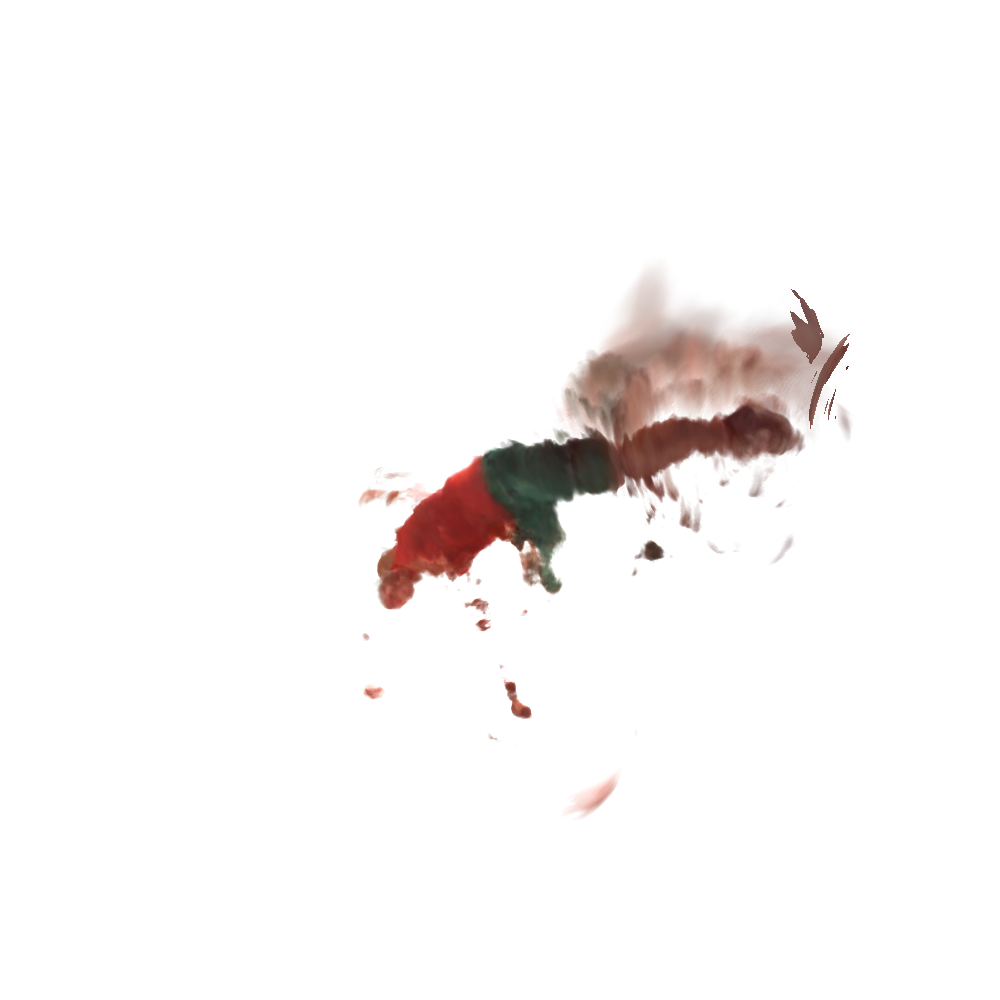}}\\%
\fbox{\includegraphics
[width=\qrtpscale,trim=410 323 310 320,clip]%
{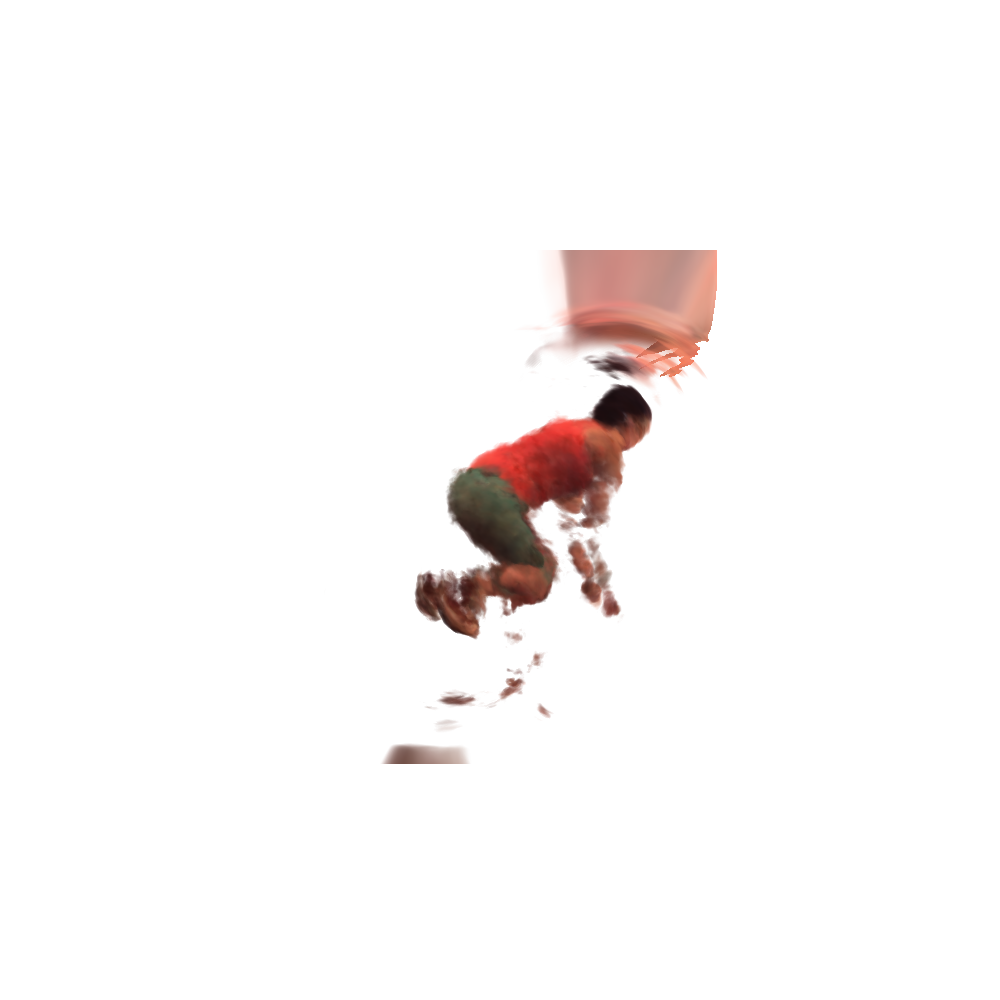}}\\%
\fbox{\includegraphics
[width=\qrtpscale,trim=370 410 310 214,clip]%
{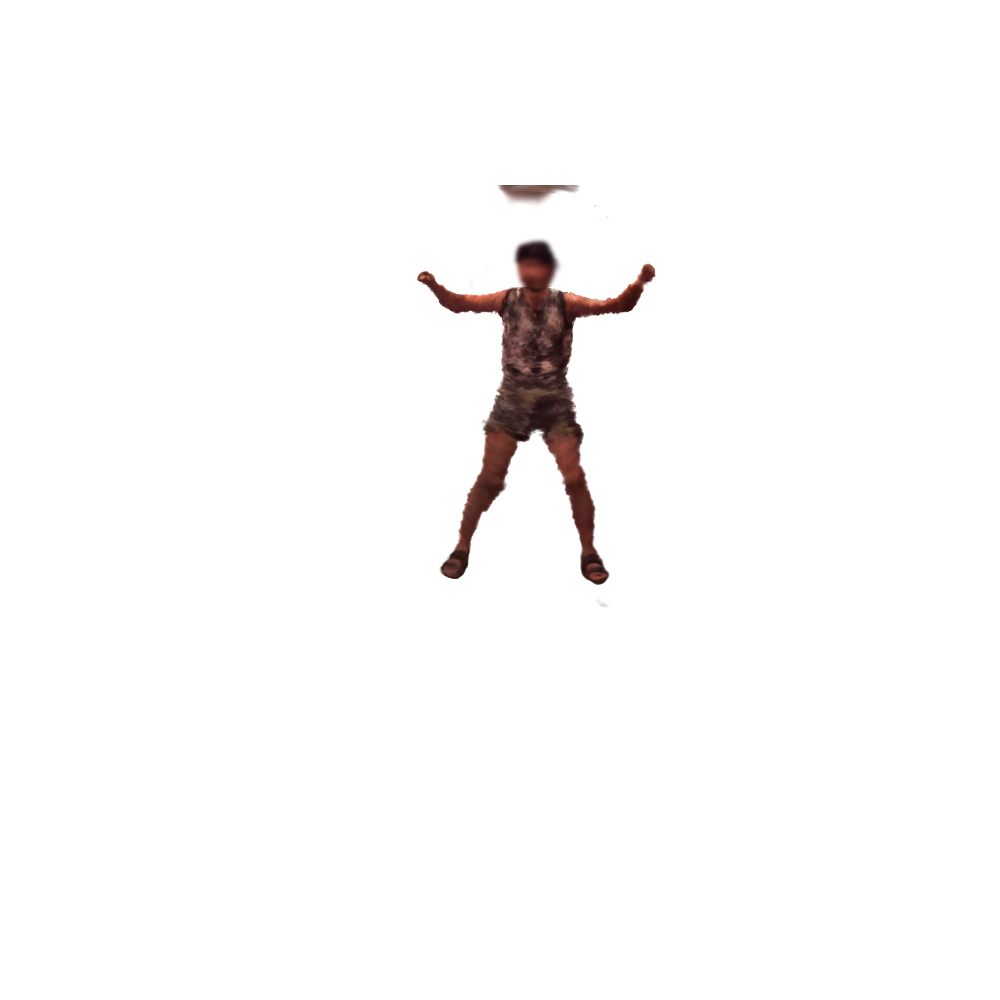}}\\%
\fbox{\includegraphics
[width=\qrtpscale,trim=280 420 400 200,clip]%
{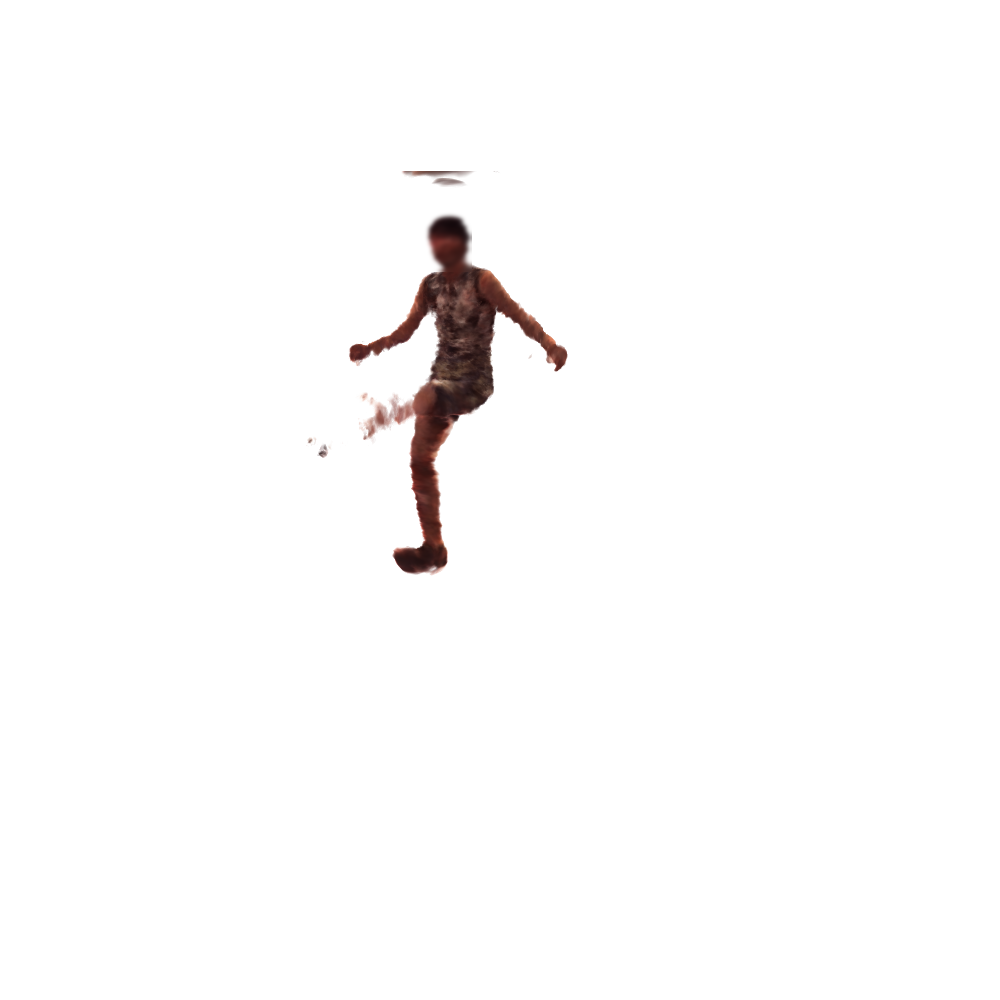}}\\%
\fbox{\includegraphics
[width=\qrtpscale,trim=250 320 400 220,clip]%
{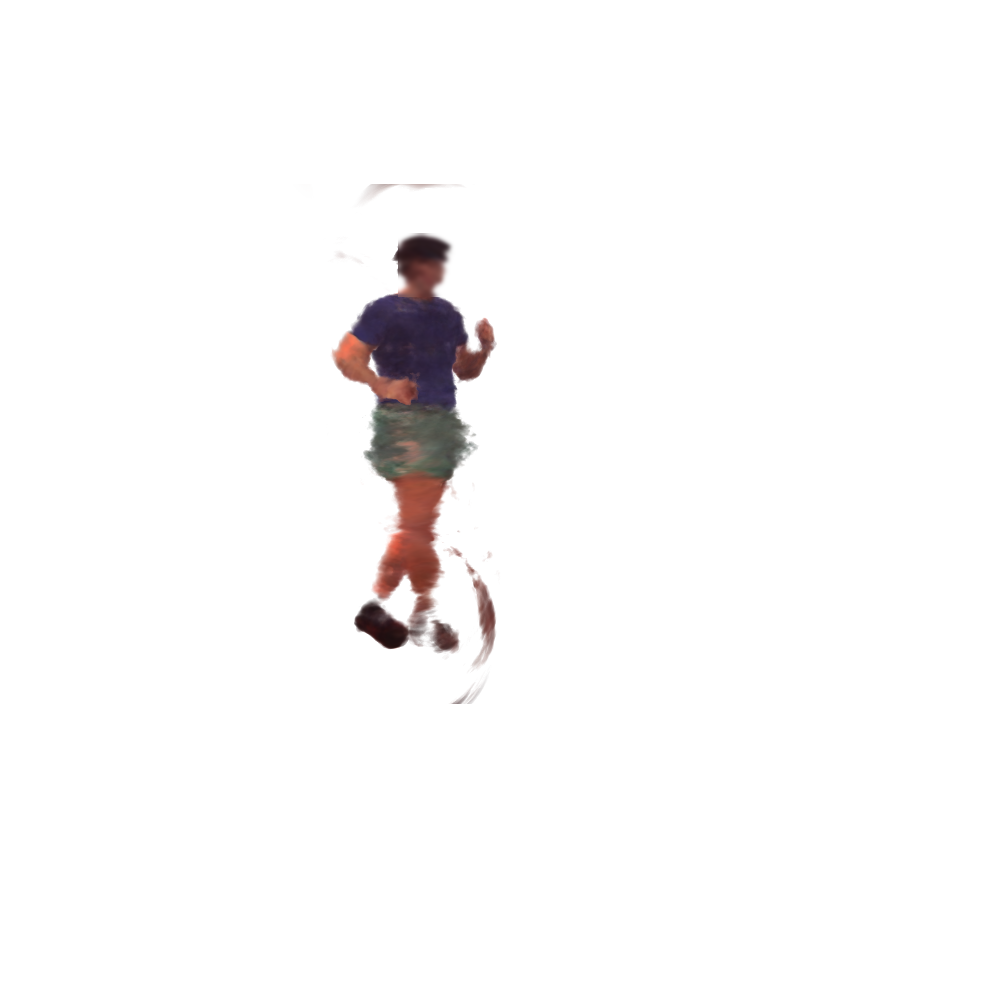}}\\%
{\scriptsize A-NeRF}%
}
\hfill%
\parbox[t]{\qrtpscale}{
\centering
\fbox{\includegraphics
[width=\qrtpscale,trim=352 200 208 300,clip]%
{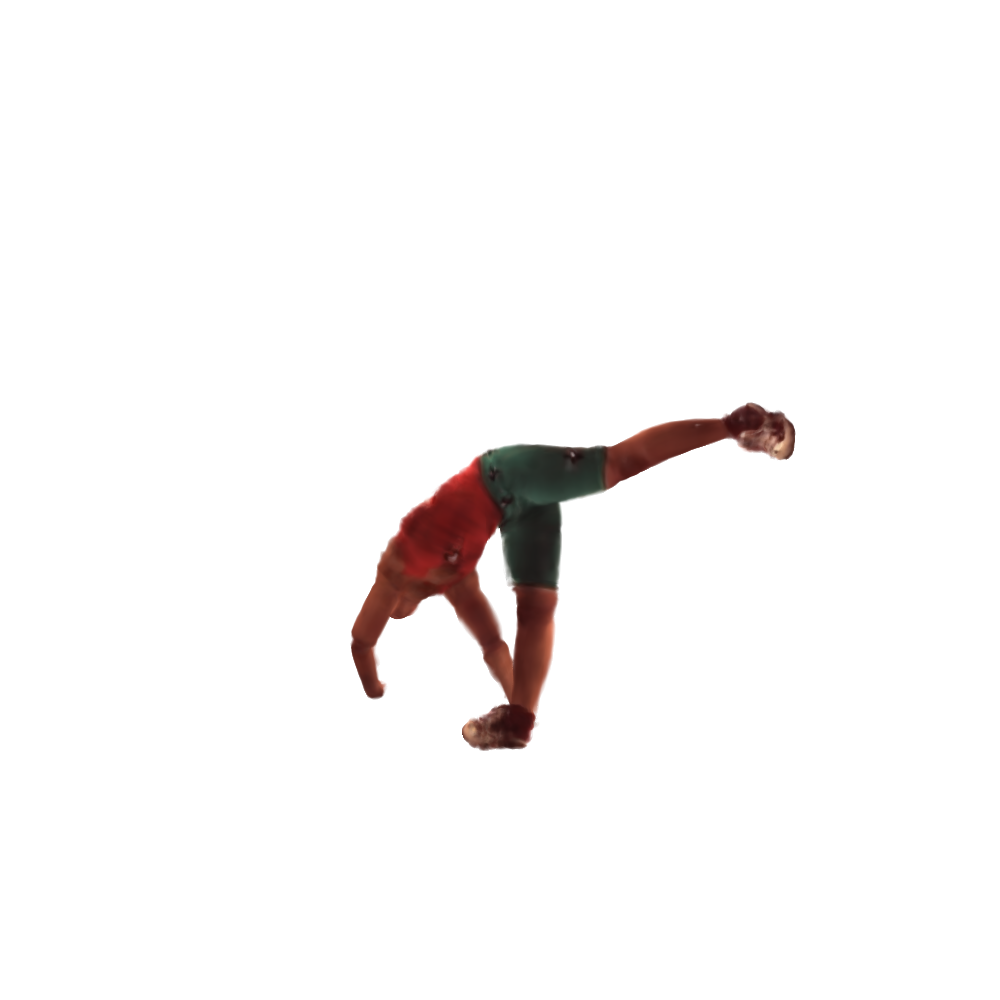}}\\%
\fbox{\includegraphics
[width=\qrtpscale,trim=410 323 310 320,clip]%
{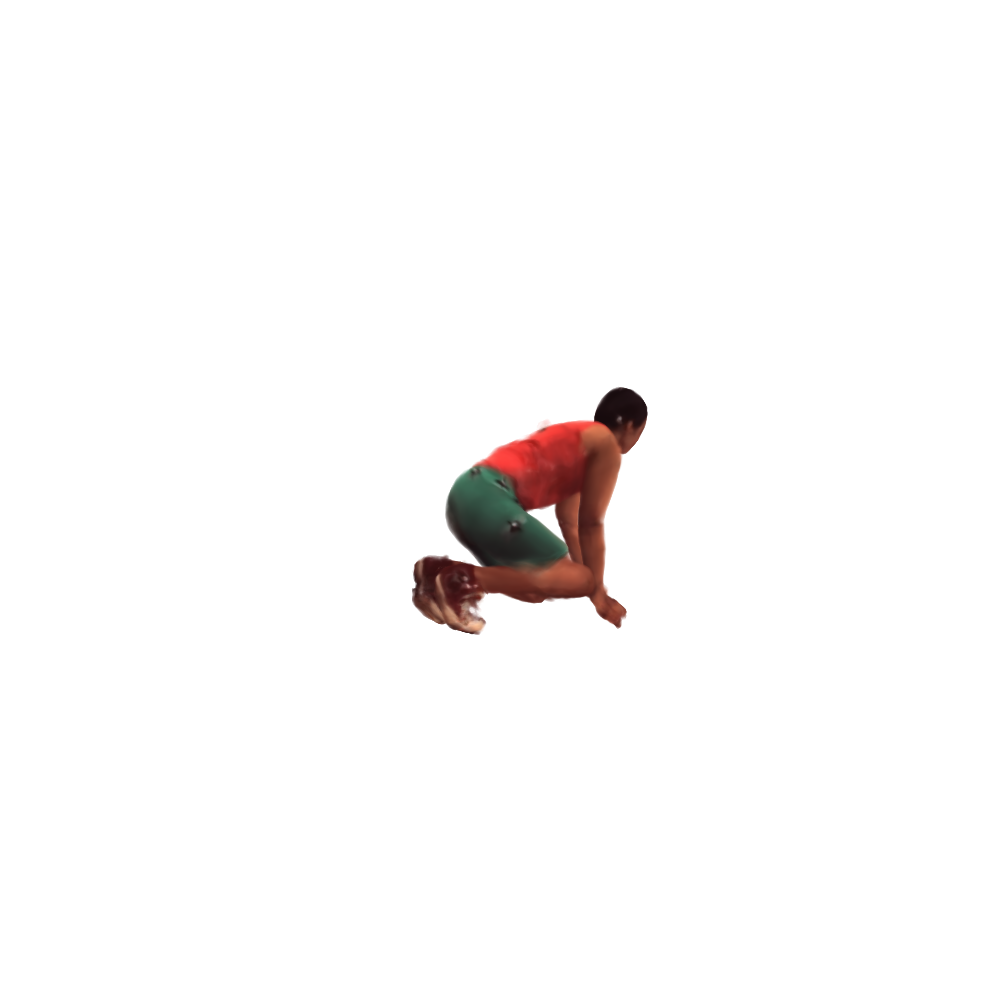}}\\%
\fbox{\includegraphics
[width=\qrtpscale,trim=370 410 310 214,clip]%
{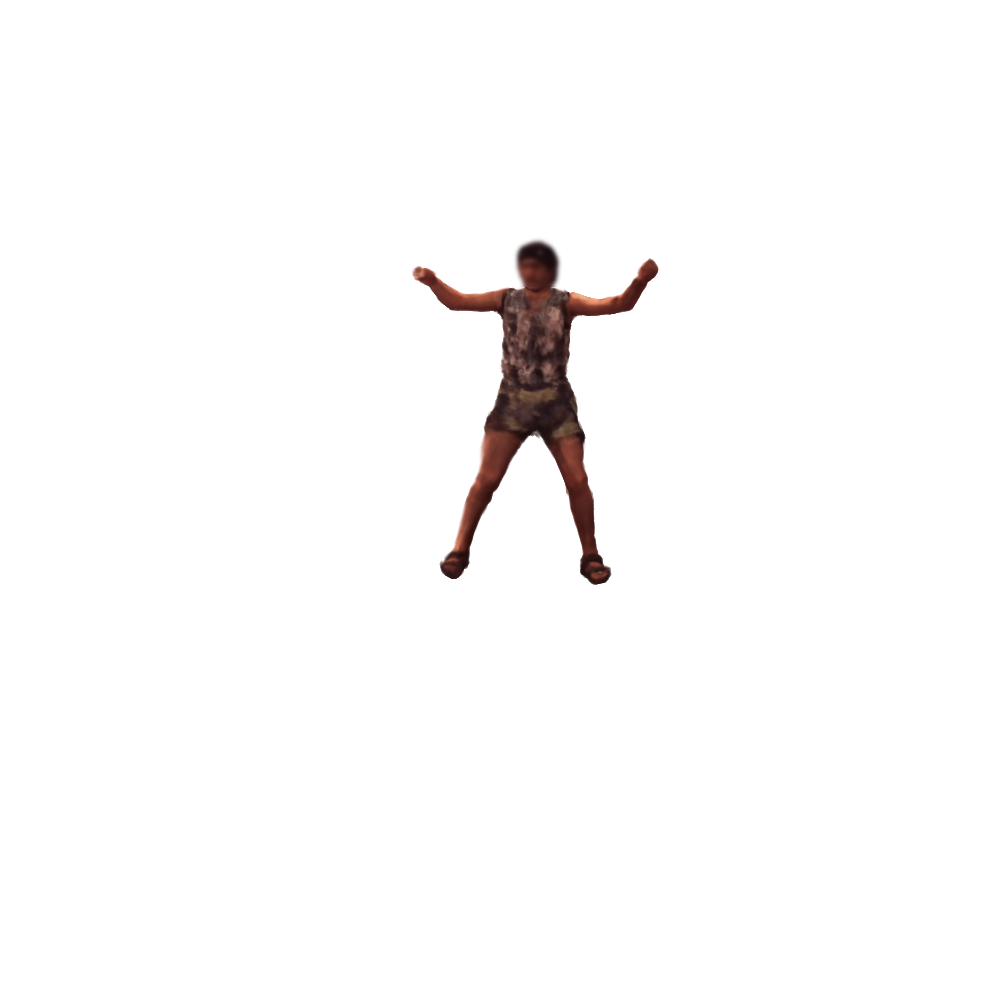}}\\%
\fbox{\includegraphics
[width=\qrtpscale,trim=280 420 400 200,clip]%
{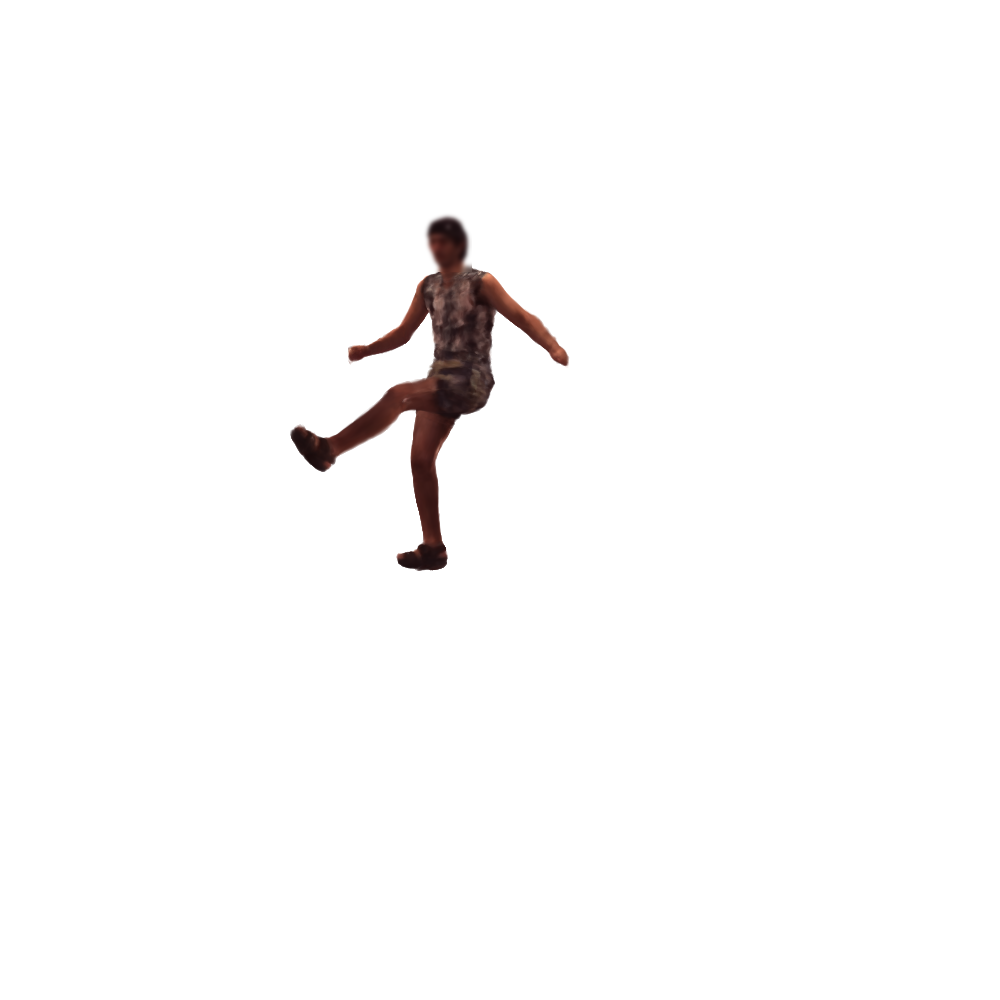}}\\%
\fbox{\includegraphics
[width=\qrtpscale,trim=250 320 400 220,clip]%
{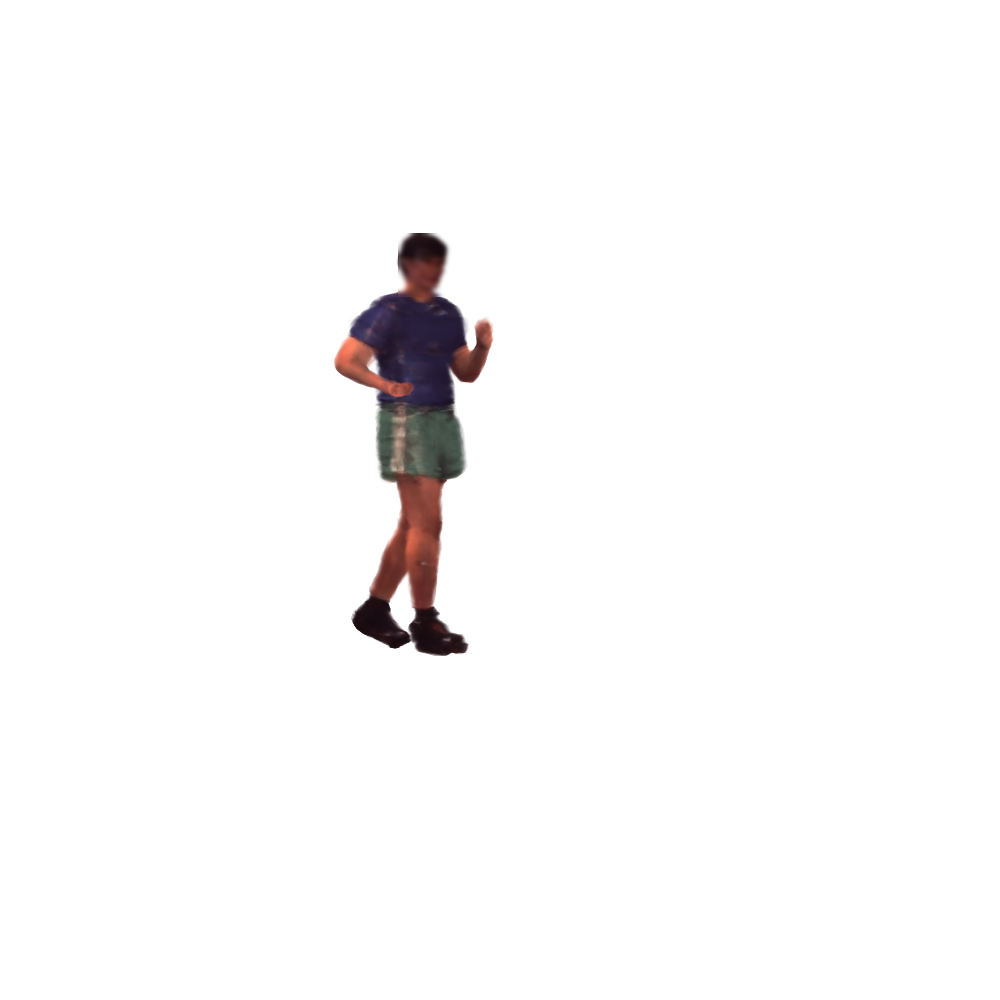}}\\%
{\scriptsize \ourapproach{} (Ours) }%
}%
\centering%
\caption{\textbf{Motion retargeting on Mixamo~\cite{mixamo} and Surreal~\cite{CMUMOCAP,varol17_surreal} dataset} with body models trained on various subjects. \ourapproach{} shows better robustness and generalization than the surface-free approach A-NeRF.
}
\label{fig:exp-motion-retarget}
\end{figure}
\newlength\qgepscale
\setlength\qgepscale{0.145\linewidth}
\newcommand{\qgeppath}{static/figure/supp/mesh}
\newcommand{\qgepaps}{S7}%
\newcommand{\qgepbps}{S9}%
\newcommand{\qgepcps}{S5}%
\newcommand{\qgepdps}{S11}%
\ifblurface
\newcommand{\qgepost}{_blur}
\else
\newcommand{\qgetpost}{}
\fi
\begin{figure}[!t]
\setlength{\fboxsep}{0pt}%
\setlength{\fboxrule}{0pt}%
\parbox[t]{\qgepscale}{%
\centering%
\fbox{\includegraphics%
[width=\qgepscale,trim=265 390 405 210,clip]%
{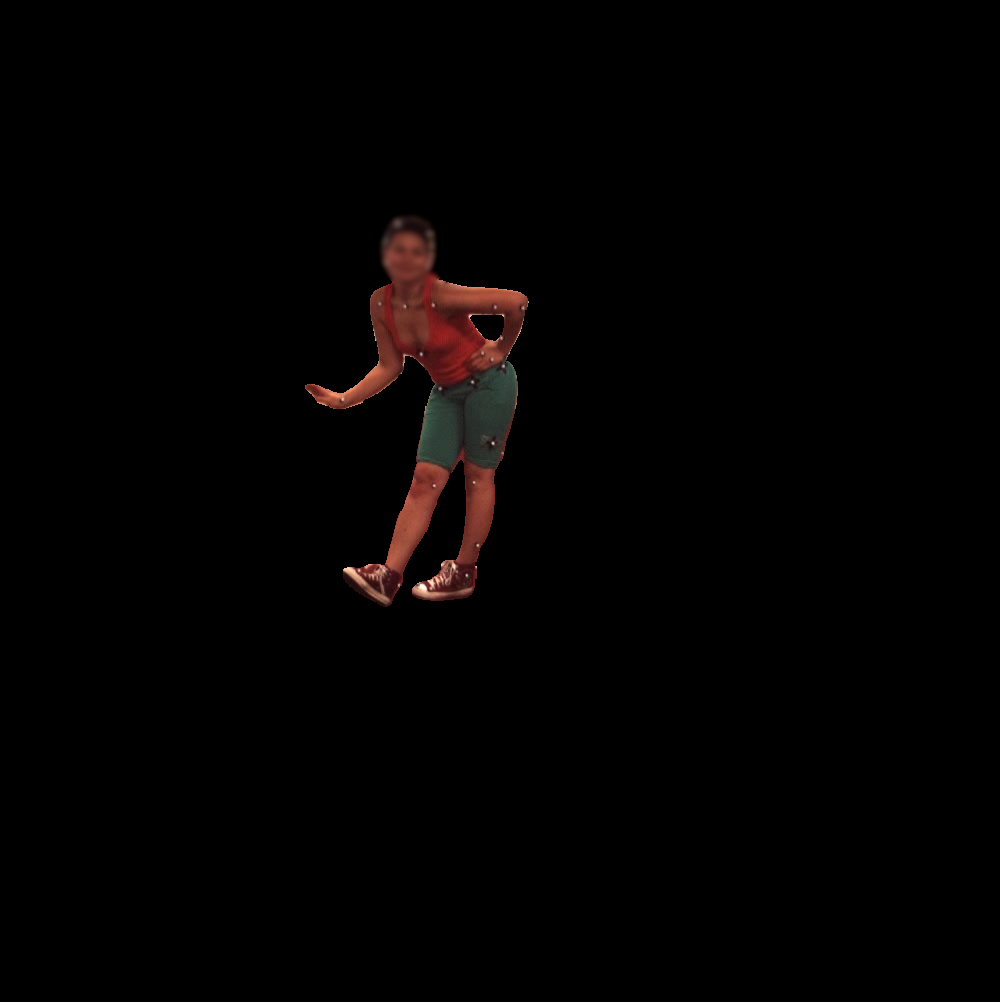}}\\%
{\small  Reference}\\%
}%
\hfill%
\parbox[t]{\qgepscale}{ %
\centering %
\fbox{\includegraphics
[width=\qgepscale,trim=120 60 120 140,clip]%
{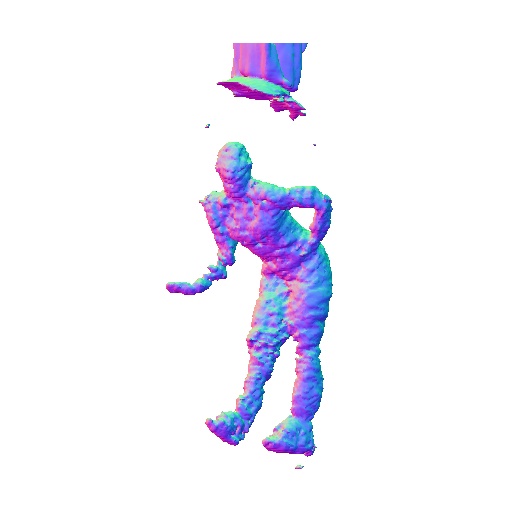}}\\%
{\small A-NeRF}\\%
}%
\hfill%
\parbox[t]{\qgepscale}{ %
\centering
\fbox{\includegraphics
[width=\qgepscale,trim=65 30 65 45,clip]%
{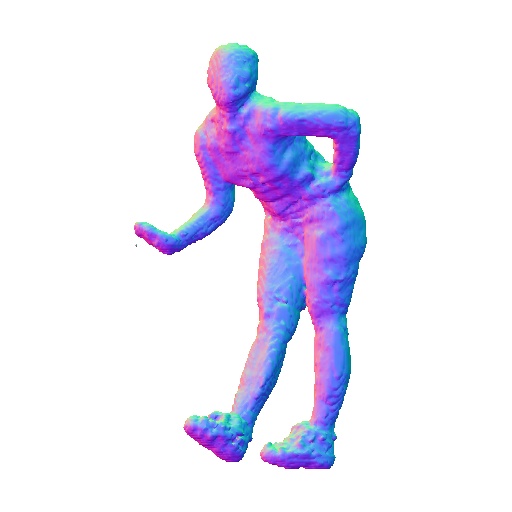}}\\%
{\small Ours}\\%
}%
\hfill%
\parbox[t]{\qgepscale}{ %
\centering
\fbox{\includegraphics%
[width=\qgepscale,trim=265 430 405 170,clip]%
{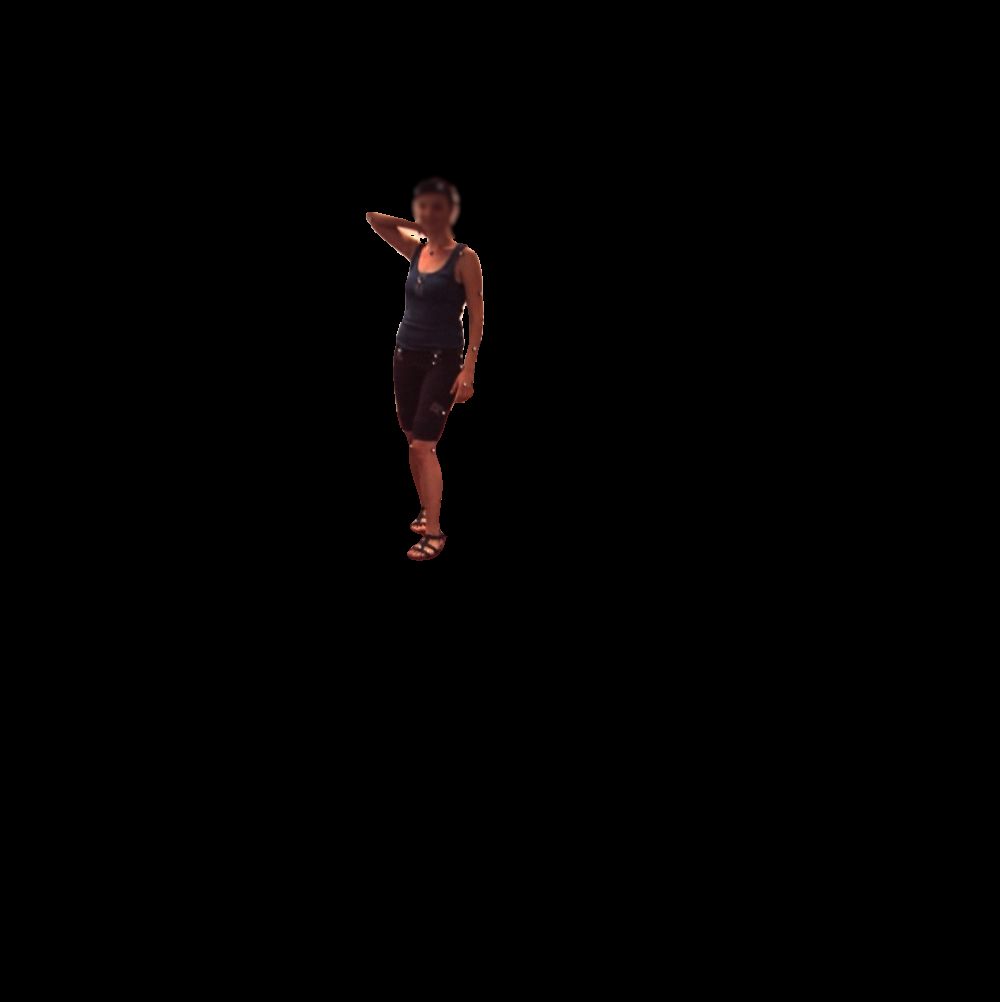}}\\%
{\small  Reference}\\%
}%
\hfill%
\parbox[t]{\qgepscale}{ %
\centering
\fbox{\includegraphics
[width=\qgepscale,trim=100 40 100 80,clip]%
{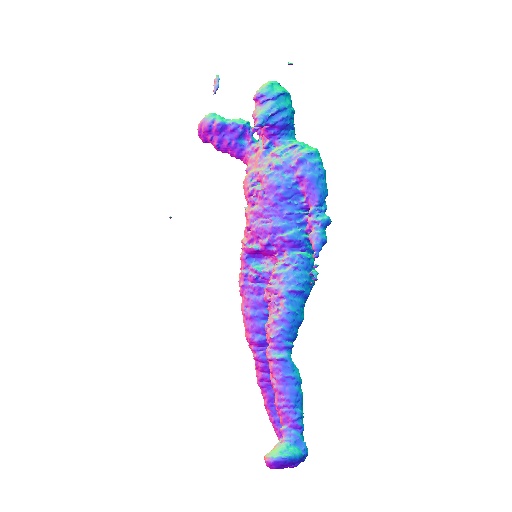}}\\%
{\small A-NeRF}\\%
}%
\hfill%
\parbox[t]{\qgepscale}{ %
\centering
\fbox{\includegraphics
[width=\qgepscale,trim=80 40 90 25,clip]%
{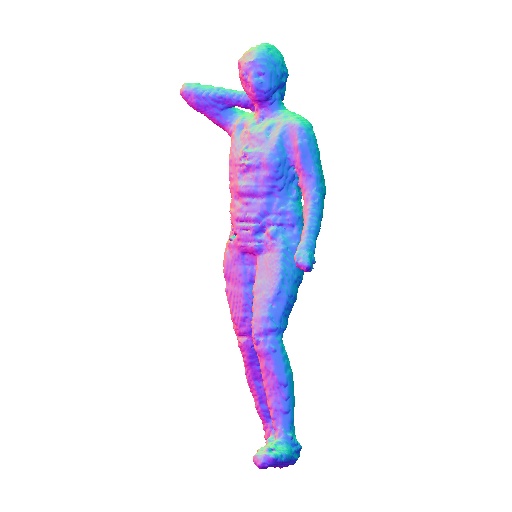}}\\%
{\small Ours}\\%
}%
\centering%
\caption{\textbf{\ourapproach{} better preserves body geometry, showing a less noisy surface than A-NeRF.} We extract the isosurface using Marching cubes~\cite{lorensen1987marching} with voxel resolution 256. See the supplemental for more results.}
\label{fig:exp-geometry}
\end{figure}
\newlength\qnabltpscale
\setlength\qnabltpscale{0.120\linewidth}
\newcommand{\qnabltpath}{static/figure/experiments/ablation/}
\newcommand{\qnabltaps}{S1/} %
\newcommand{\qnabltapi}{000}
\newcommand{\qnabltbps}{S6/} %
\newcommand{\qnabltbpi}{001}
\newcommand{\qnabltcps}{S9/} %
\newcommand{\qnabltcpi}{004}
\newcommand{\qnabltdps}{S11/} %
\newcommand{\qnabltdpi}{004}
\ifblurface
\newcommand{\qnabltpost}{_blur}
\else
\newcommand{\qnabltpost}{}
\fi
\begin{figure}[h]
\setlength{\fboxrule}{0pt}%
\setlength{\fboxsep}{0pt}%
\parbox[t]{\qnabltpscale}{
\centering
\fbox{\includegraphics
[width=\qnabltpscale,trim=350 625 450 145,clip]
{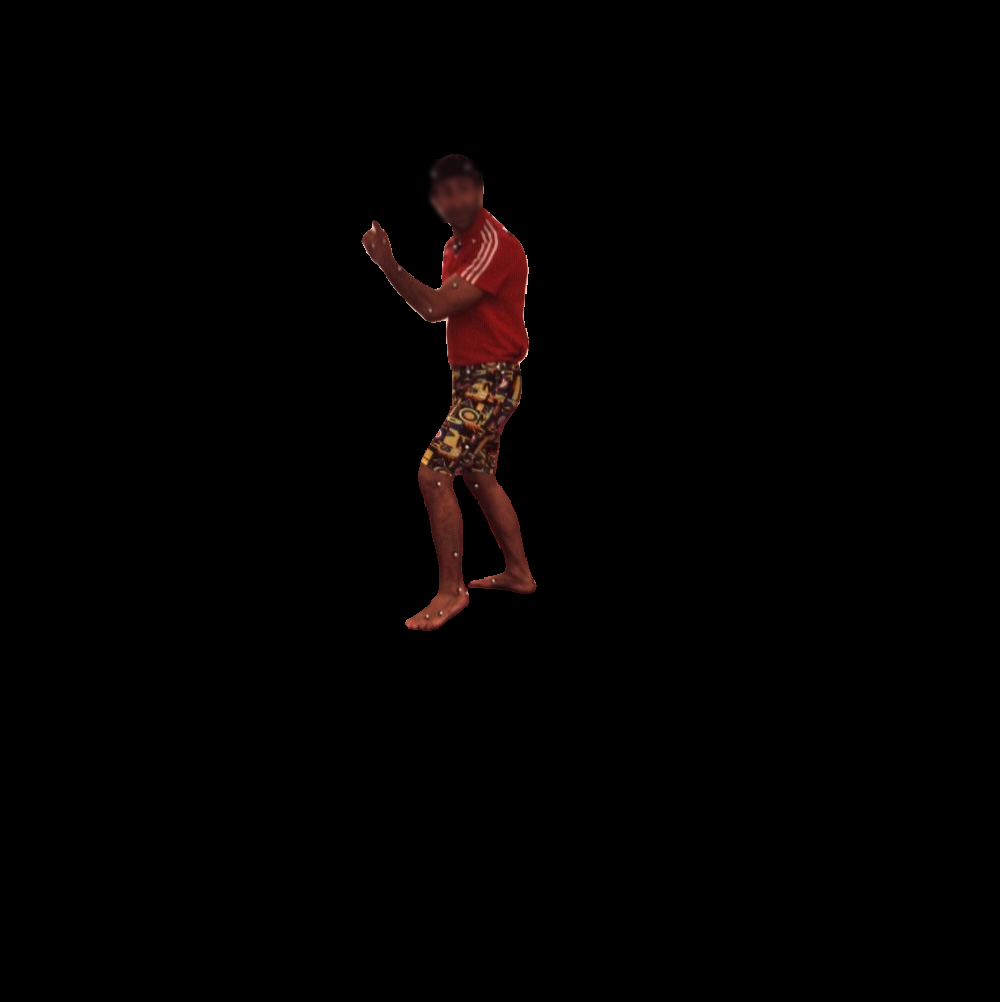}
}\\%
\fbox{\includegraphics
[width=\qnabltpscale,trim=490 635 320 145,clip]
{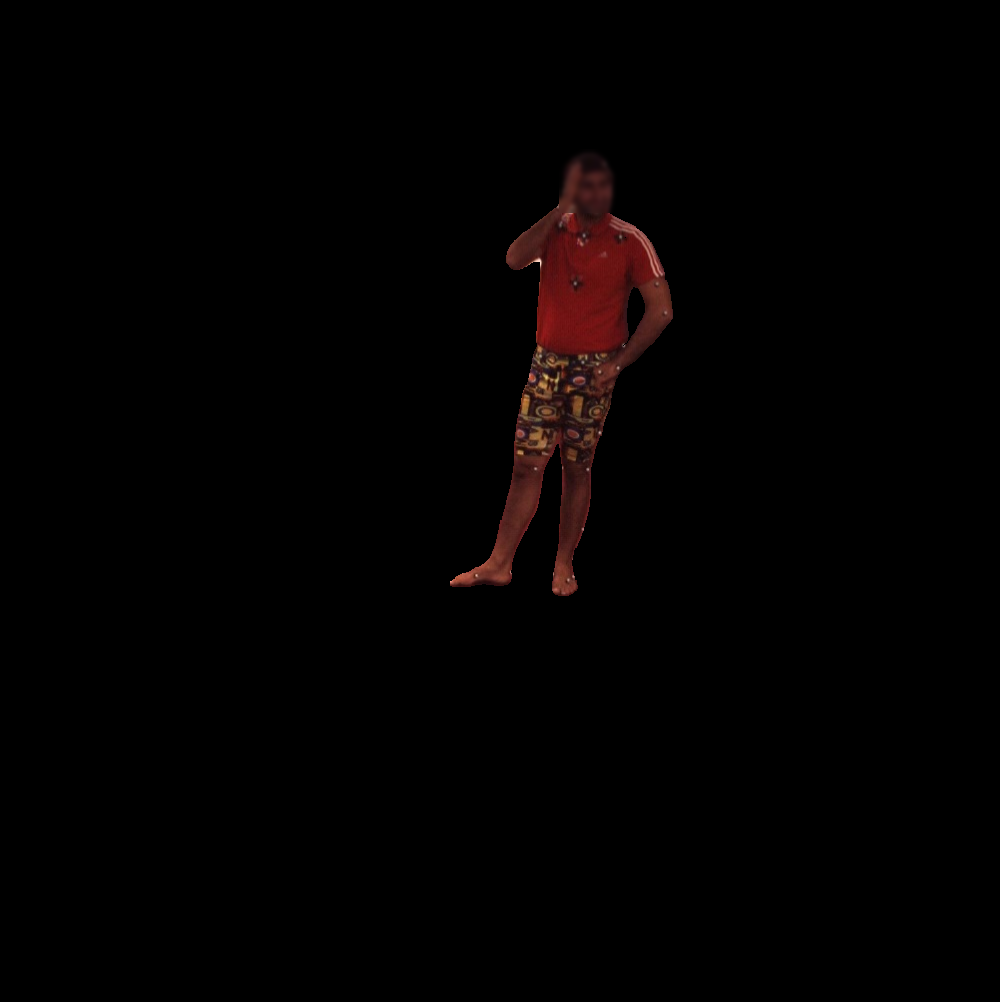}
}\\%
{\scriptsize Ground Truth}%
}%
\hfill%
\parbox[t]{\qnabltpscale}{
\centering
\fbox{\includegraphics
[width=\qnabltpscale,trim=350 625 450 145,clip]
{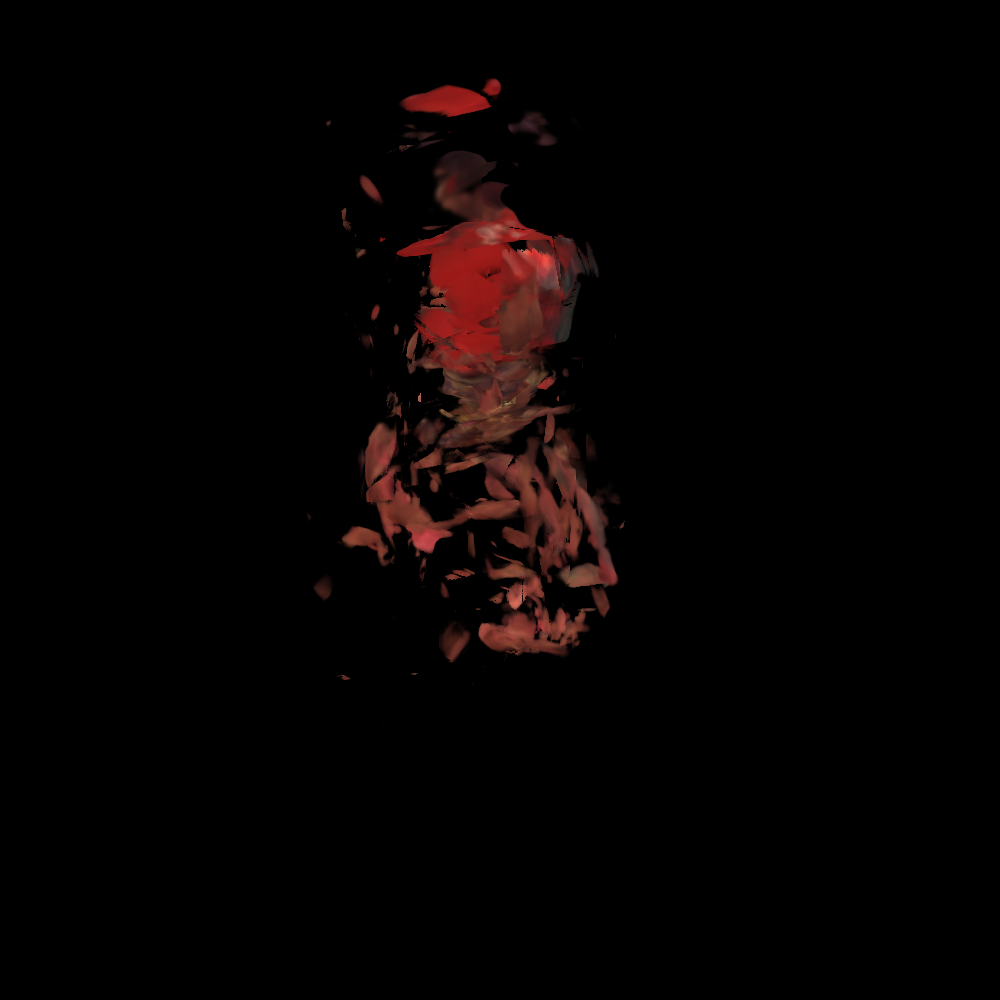}
}\\%
\fbox{\includegraphics
[width=\qnabltpscale,trim=490 635 320 145,clip]
{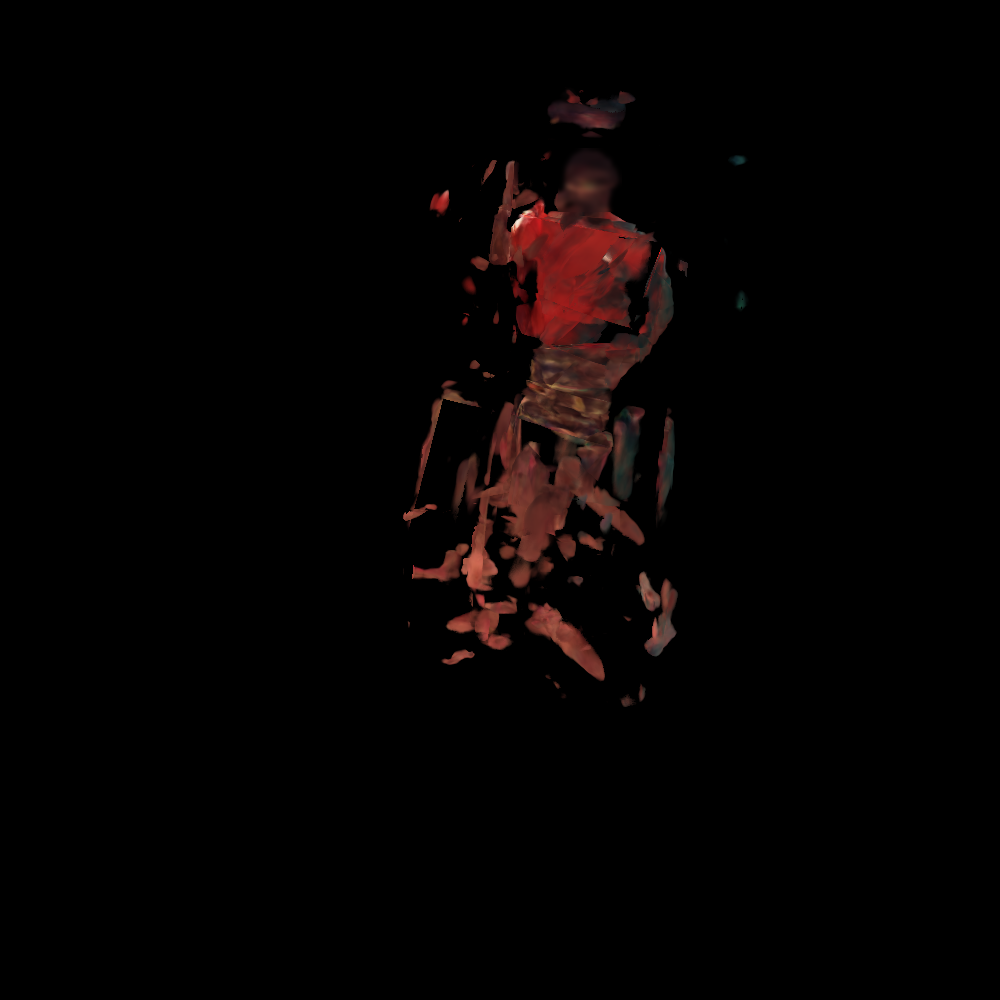}
}\\%
{\scriptsize w/o aggregation}%
}%
\hfill%
\parbox[t]{\qnabltpscale}{
\centering
\fbox{\includegraphics
[width=\qnabltpscale,trim=350 625 450 145,clip]
{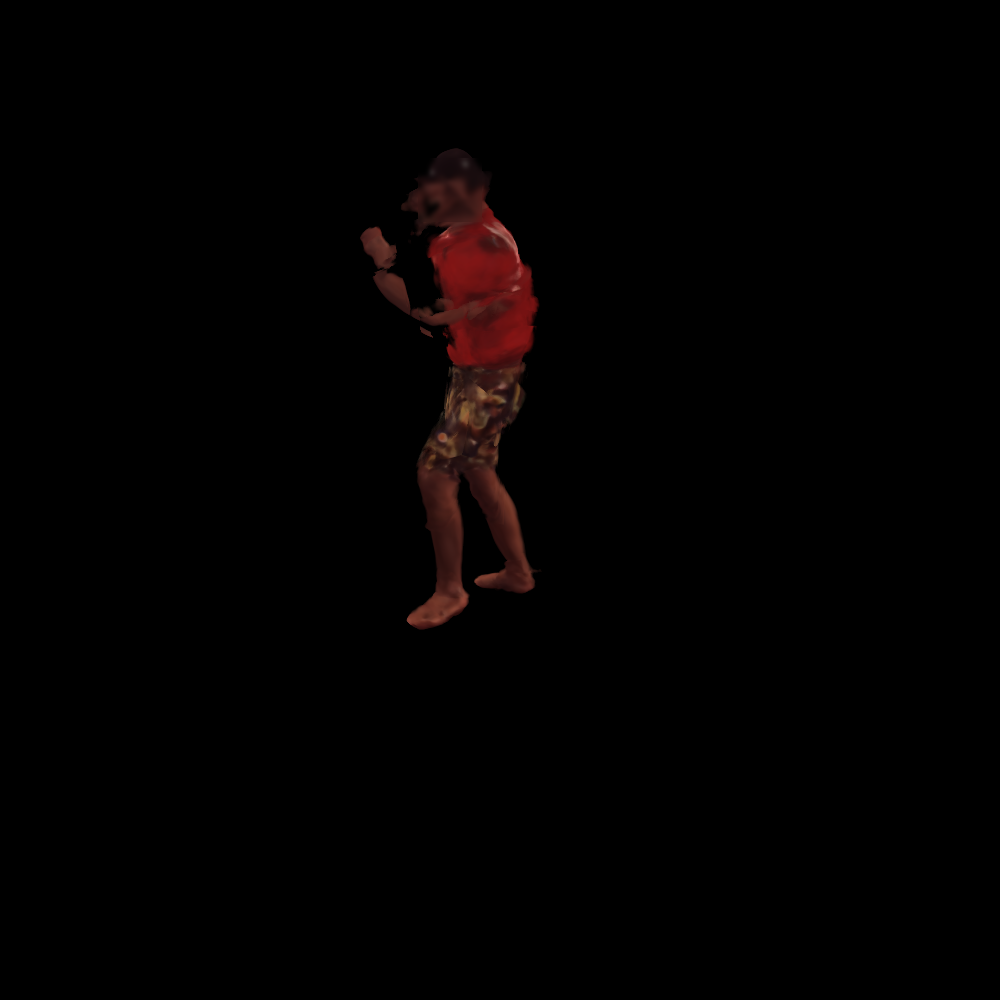}
}\\%
\fbox{\includegraphics
[width=\qnabltpscale,trim=490 635 320 145,clip]
{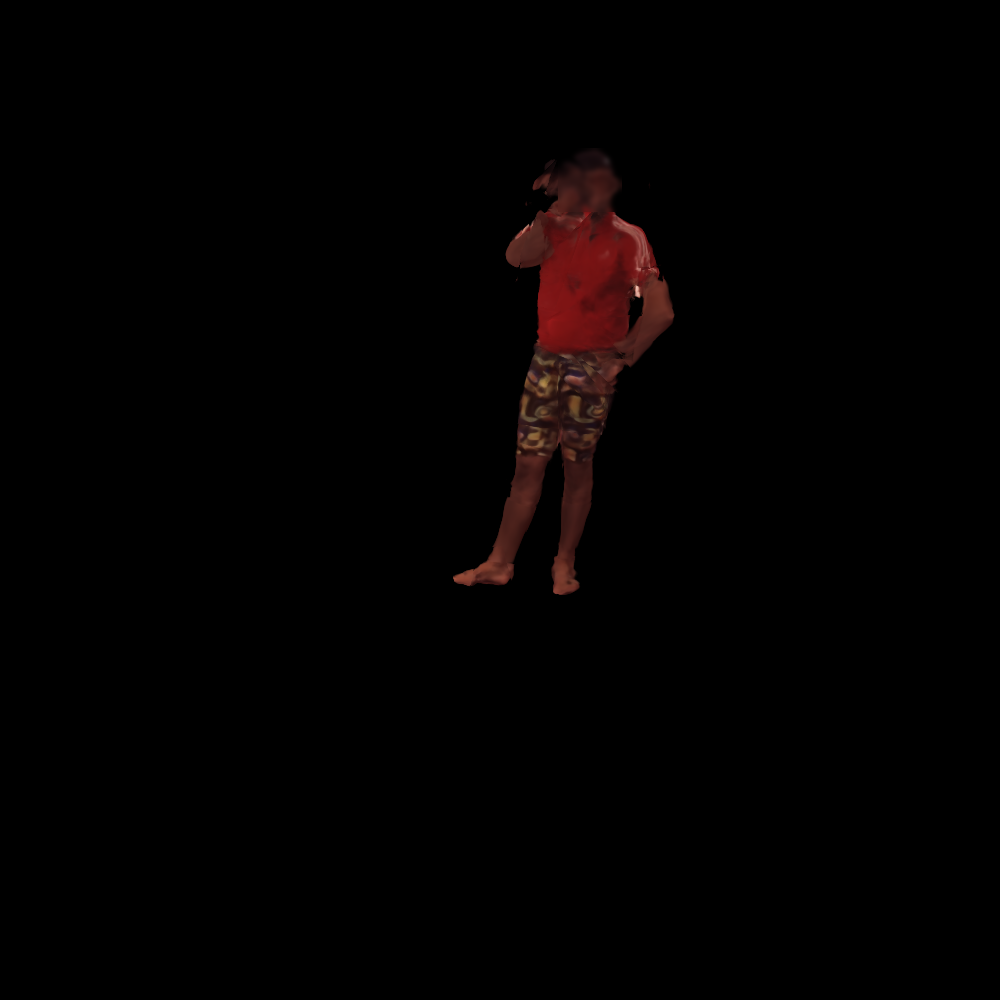}
}\\%
{\scriptsize w/o volume}%
}%
\hfill%
\parbox[t]{\qnabltpscale}{
\centering
\fbox{\includegraphics
[width=\qnabltpscale,trim=350 625 450 145,clip]
{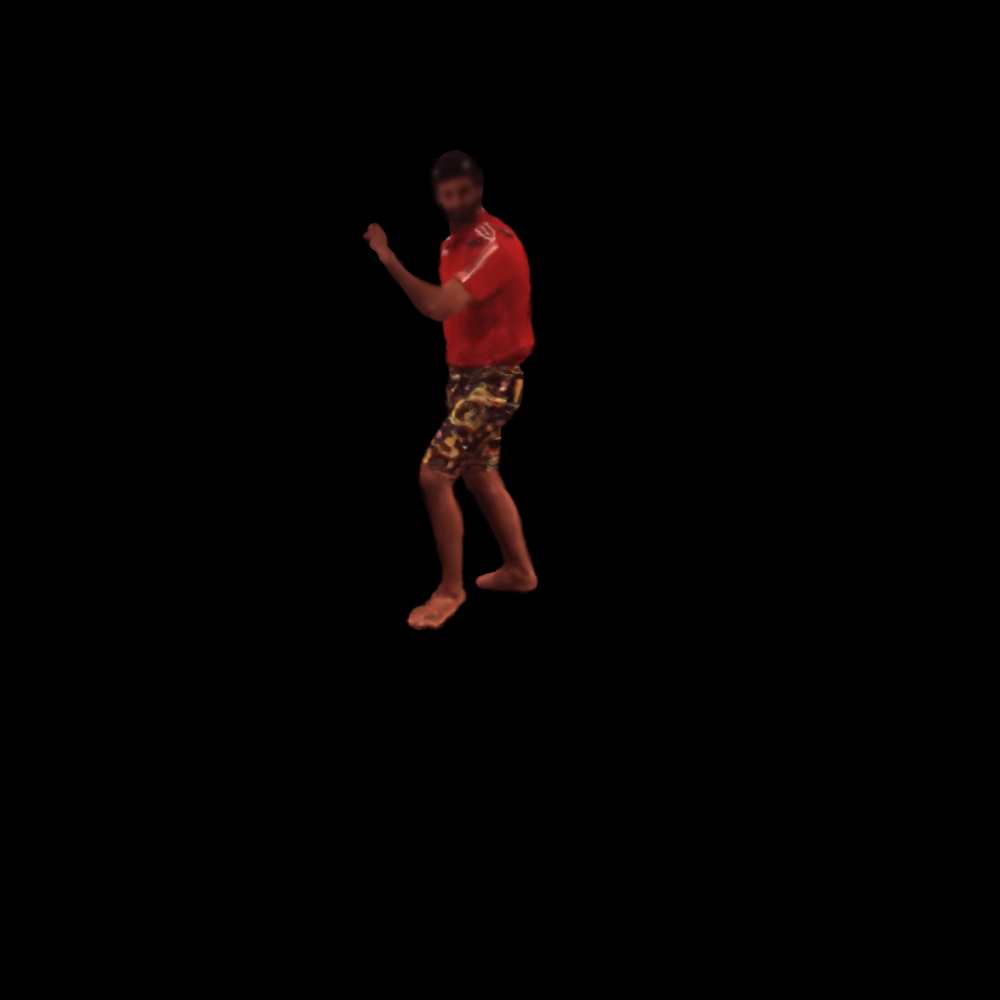}
}\\%
\fbox{\includegraphics
[width=\qnabltpscale,trim=490 635 320 145,clip]
{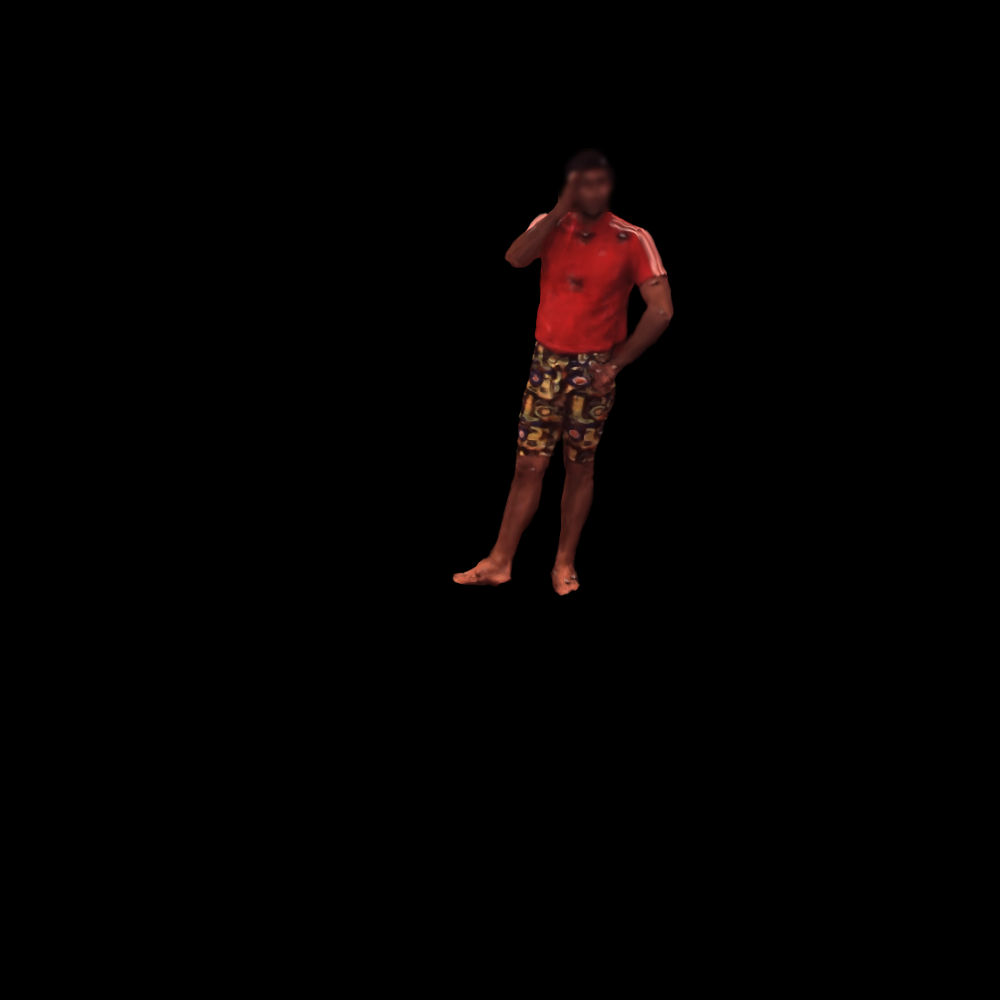}
}\\%
{\scriptsize Softmax}%
}%
\hfill%
\parbox[t]{\qnabltpscale}{
\centering
\fbox{\includegraphics
[width=\qnabltpscale,trim=350 625 450 145,clip]
{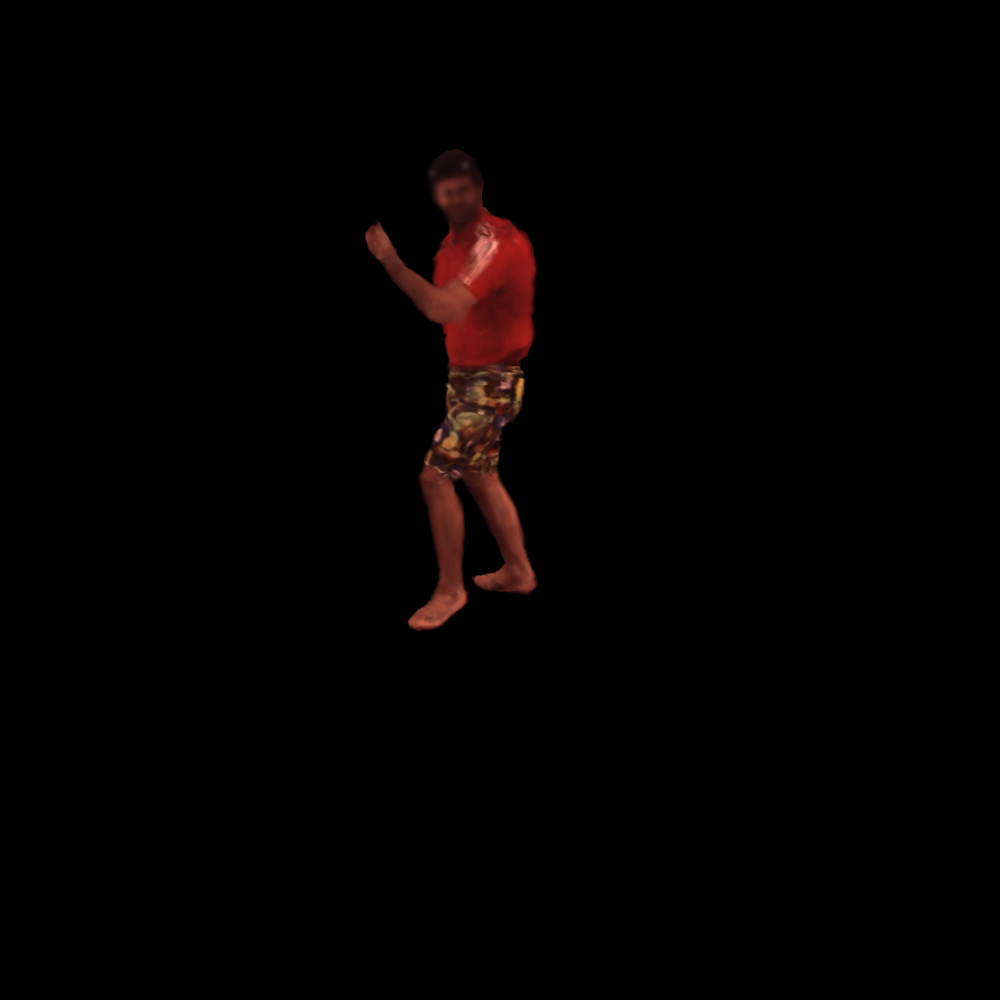}
}\\%
\fbox{\includegraphics
[width=\qnabltpscale,trim=490 635 320 145,clip]
{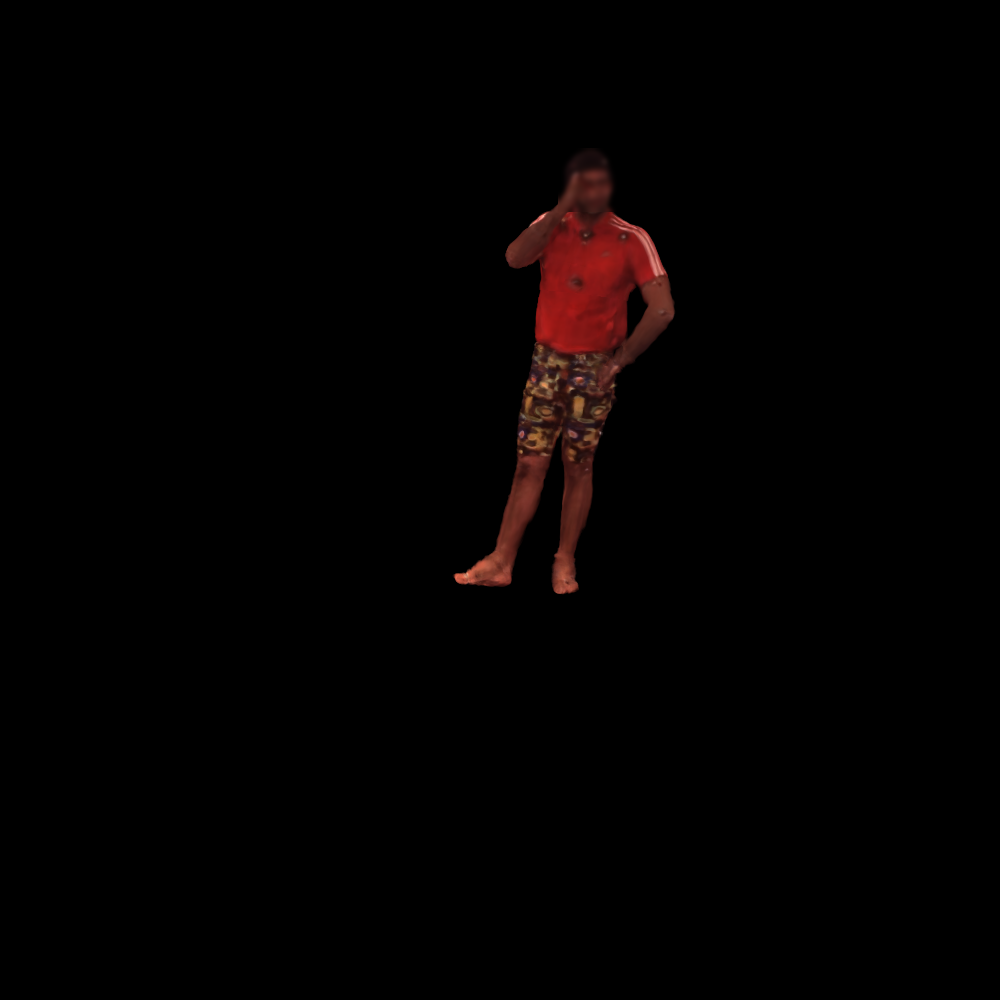}
}\\%
{\scriptsize Softmax-OOB}%
}%
\hfill%
\parbox[t]{\qnabltpscale}{
\centering
\fbox{\includegraphics
[width=\qnabltpscale,trim=350 625 450 145,clip]
{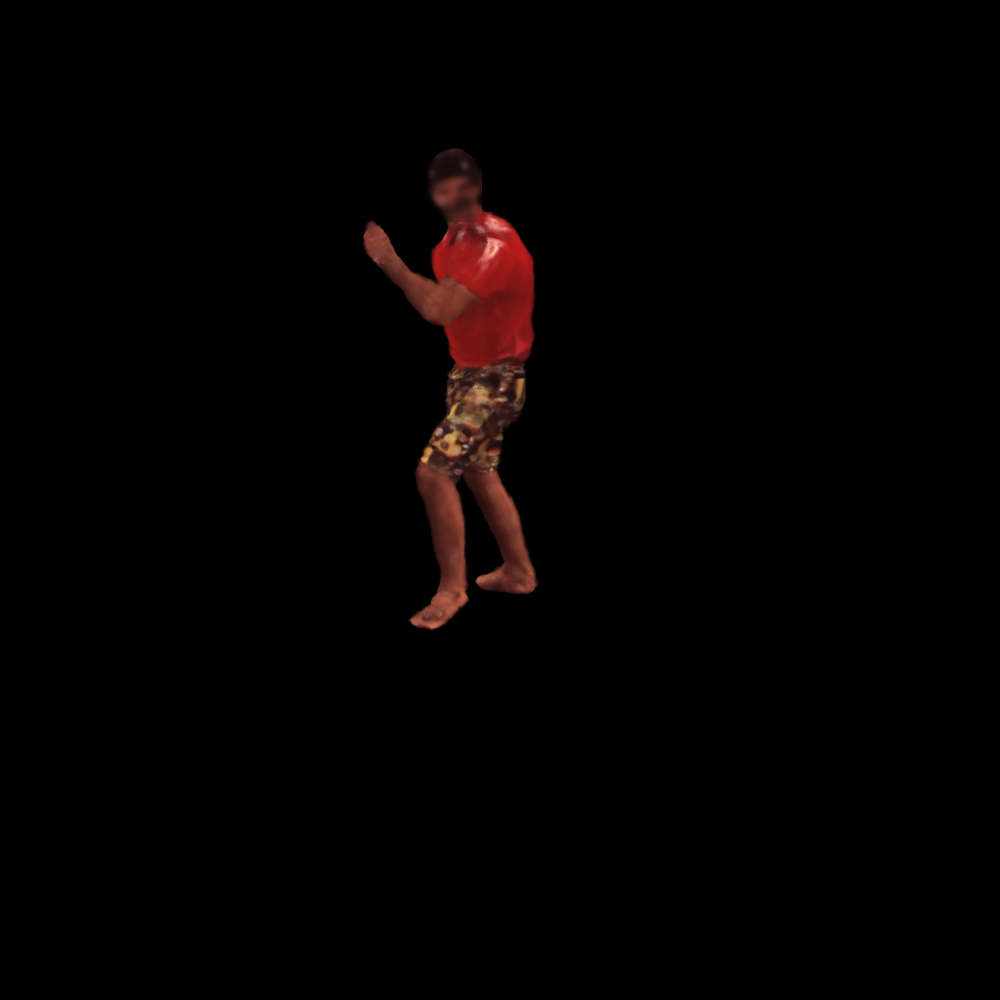}
}\\%
\fbox{\includegraphics
[width=\qnabltpscale,trim=490 635 320 145,clip]
{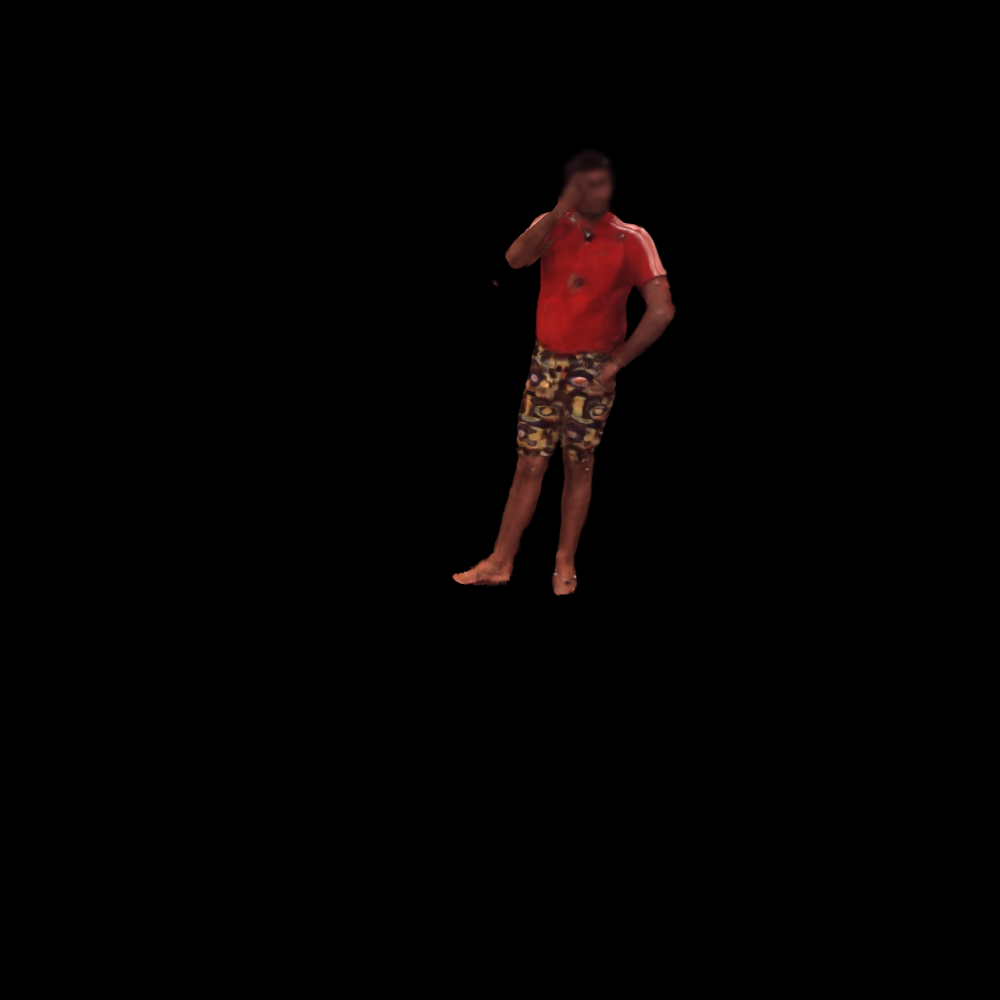}
}\\%
{\scriptsize Sum-OOB}%
}%
\hfill%
\parbox[t]{\qnabltpscale}{
\centering
\fbox{\includegraphics
[width=\qnabltpscale,trim=350 625 450 145,clip]
{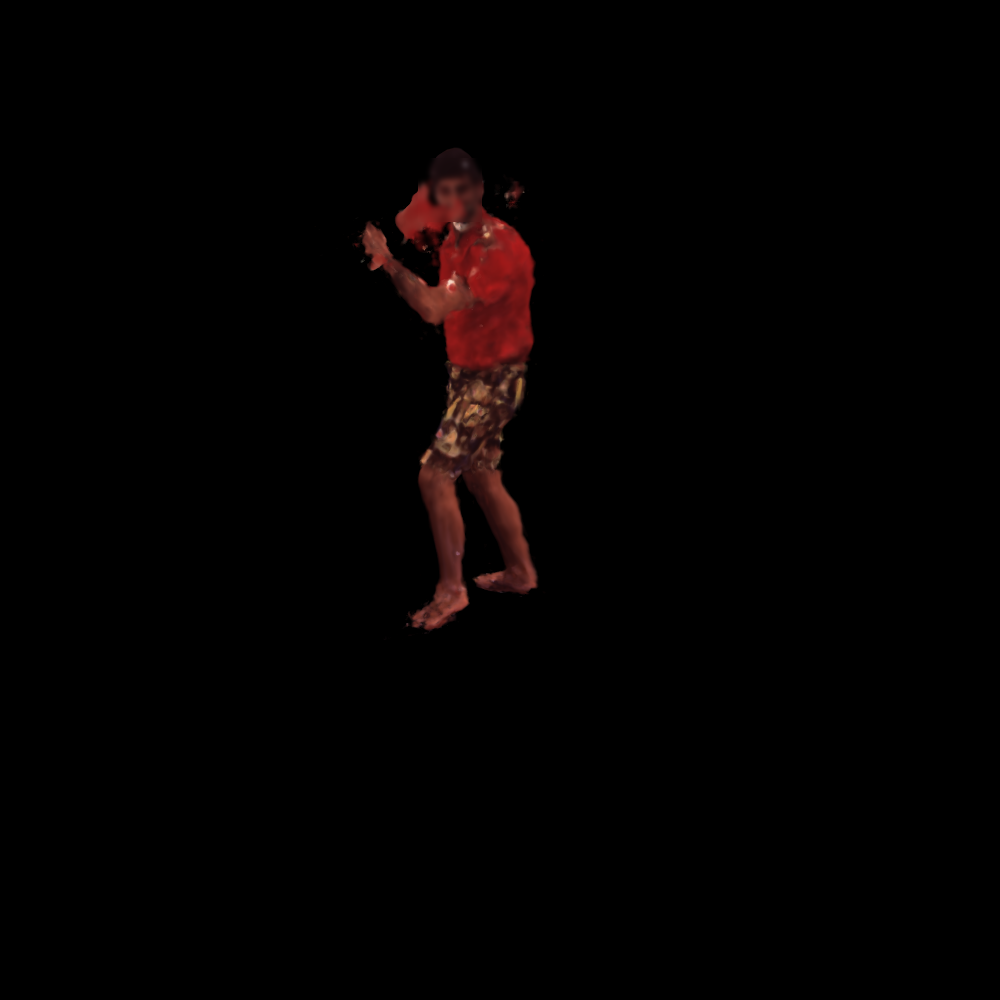}
}\\%
\fbox{\includegraphics
[width=\qnabltpscale,trim=490 635 320 145,clip]
{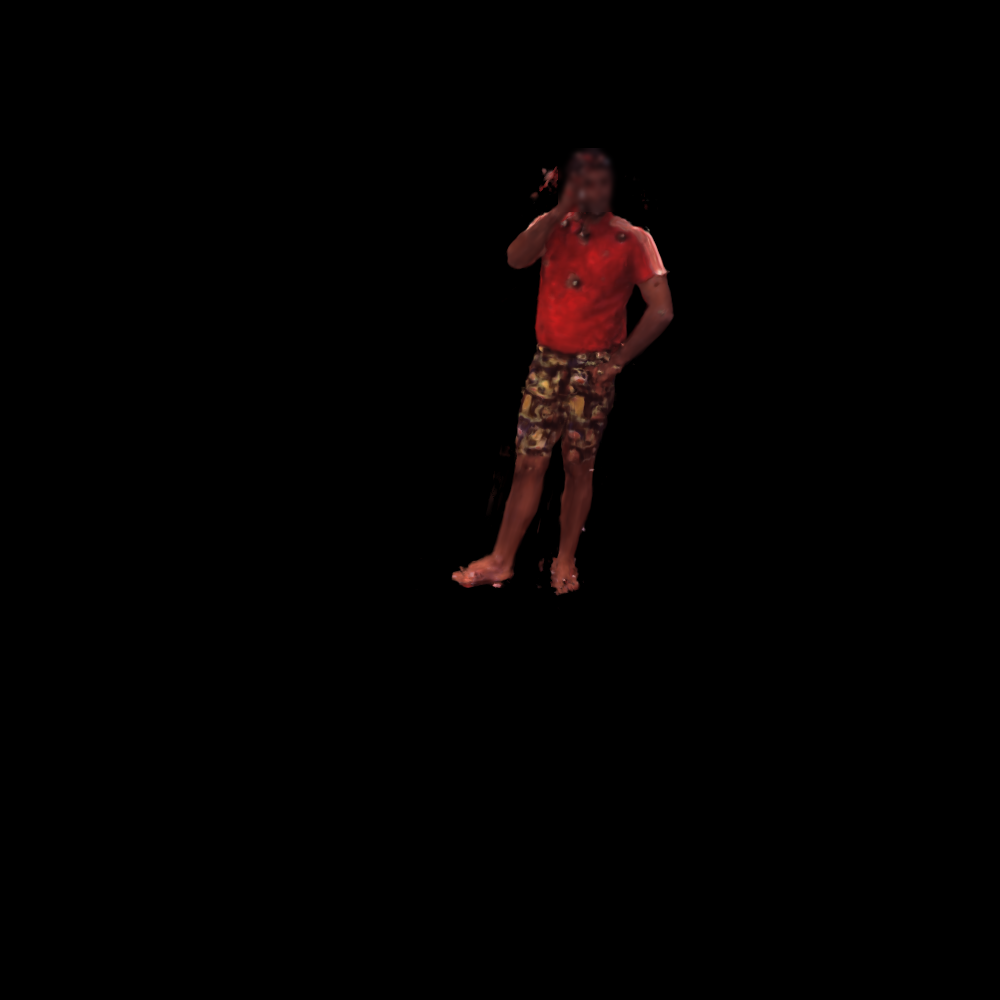}
}\\%
{\scriptsize SlabConv}%
}%
\hfill%
\parbox[t]{\qnabltpscale}{
\centering
\fbox{\includegraphics
[width=\qnabltpscale,trim=350 625 450 145,clip]
{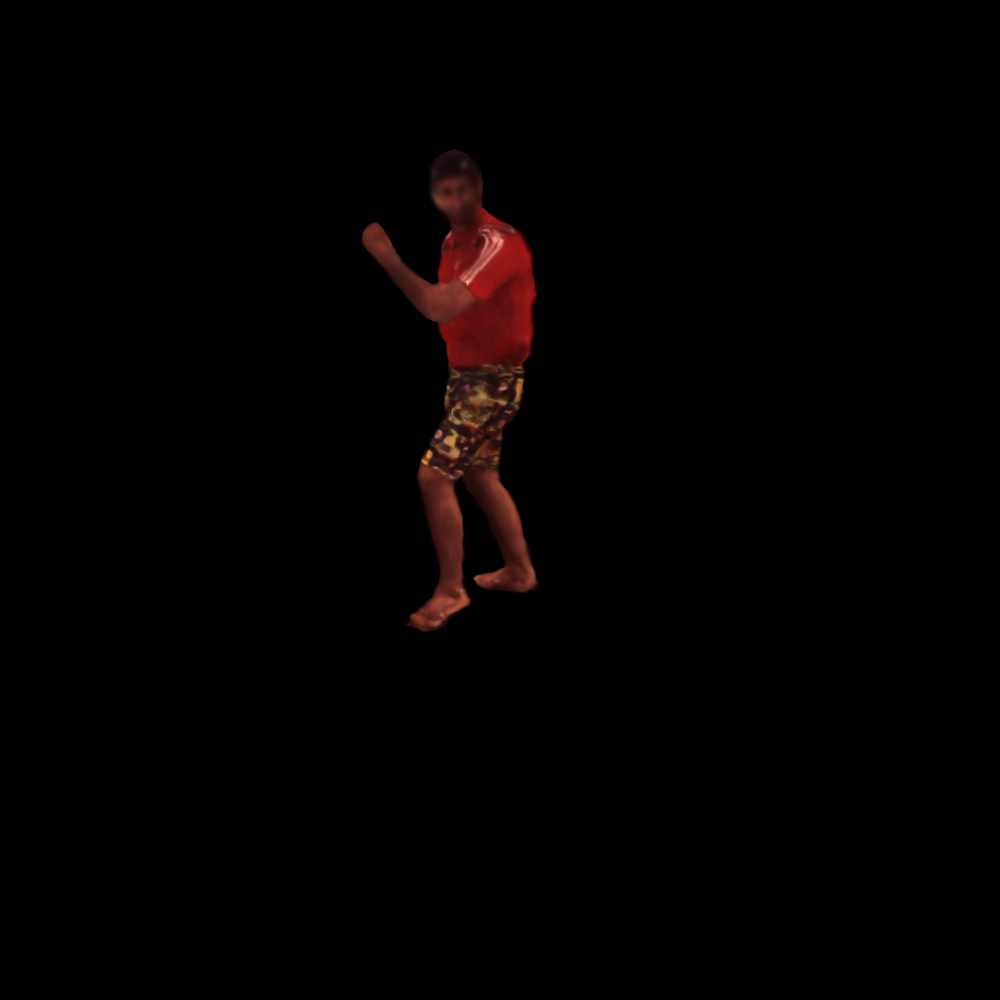}
}\\%
\fbox{\includegraphics
[width=\qnabltpscale,trim=490 635 320 145,clip]
{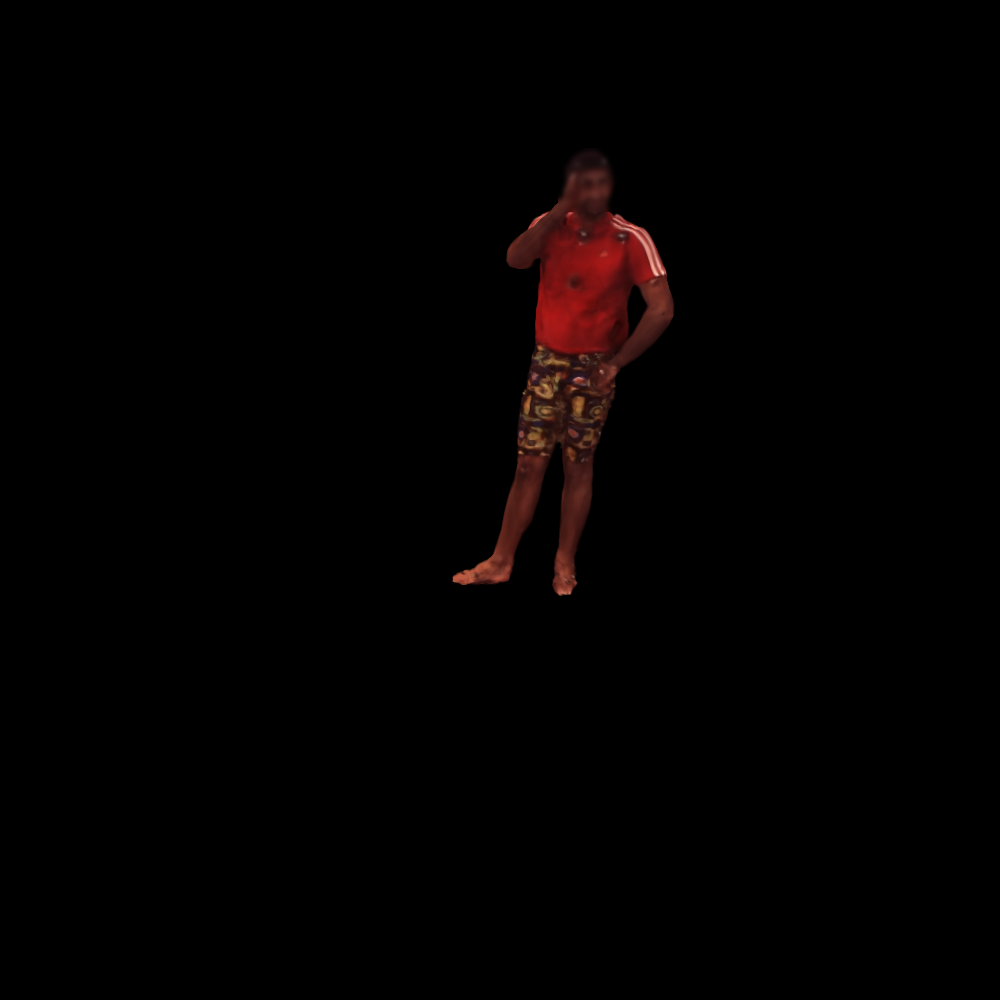}
}\\%
{\scriptsize Ours}%
}%
\centering%
\caption{\textbf{Ablation study on Human3.6M~\cite{Ionescu14b} test split} novel pose (top row) and novel view (bottom row). Our proposed designs together achieve better results with less distortion on the body parts, particularly in the limbs and face.
}%
\label{fig:exp-ablation}
\end{figure}
\subsection{Novel View Synthesis}
\label{sec:exp-novel-view}
View synthesis of poses seen during training is simpler as the interplay between body parts is observable. Hence, our explicit disentanglement of body parts is less crucial but still beneficial. Compared to the
baselines,
higher detail is present and body shape is better preserved, such as visible at %
facial features and arm contours in~\figref{fig:exp-novel-view}. %
Anim-NeRF shows slightly distorted arms and cloud-like artifacts, potentially caused by incorrectly estimated deformation fields.
\tabref{tab:exp-h36m-novel-view} verifies these improvements %
on the test set of Anim-NeRF~\cite{peng2021animatable}. 

\subsection{Unseen Pose Synthesis and Animation}
\begin{table}[!t]
\caption{\textbf{Novel-pose synthesis comparisons on Human3.6M~\cite{Ionescu14b} (row 1-8) and MonoPerfCap~\cite{xu2018monoperfcap} (row 9-11)} . Our part-disentangled design enables \ourapproach{} to generalize better to unseen poses with superior perceptual qualities.}
\centering
\resizebox{1.0\linewidth}{!}{
\setlength{\tabcolsep}{3pt}
\begin{tabular}{lcccccccccccccccc}
\toprule
 & \multicolumn{4}{c}{NeuralBody~\cite{peng2020neuralbody}}  & \multicolumn{4}{c}{Anim-NeRF~\cite{peng2021animatable}} & \multicolumn{4}{c}{A-NeRF~\cite{su2021anerf}} & \multicolumn{4}{c}{\tbf{\ourapproach{} (Ours)}} \\
 \midrule
 & PSNR~$\uparrow$& SSIM~$\uparrow$& KID~$\downarrow$& LPIPS~$\downarrow$& PSNR~$\uparrow$& SSIM~$\uparrow$& KID~$\downarrow$& LPIPS~$\downarrow$& PSNR~$\uparrow$& SSIM~$\uparrow$& KID~$\downarrow$& LPIPS~$\downarrow$& PSNR~$\uparrow$& SSIM~$\uparrow$& KID~$\downarrow$& LPIPS~$\downarrow$\\
\rowcolor{Gray}
S1 & 22.10& 0.878& 0.110& 0.140& 21.37& 0.868& 0.163& 0.141& 22.67& 0.883& 0.178& 0.143& \tbf{23.03}& \tbf{0.895}& \tbf{0.081}& \tbf{0.135}\\
S5 & 23.52& 0.897& \tbf{0.039}& 0.151& 22.29& 0.875& 0.123& 0.155& 22.96& 0.888& 0.081& 0.157& \tbf{23.66}& \tbf{0.903}& 0.049& \tbf{0.147}\\
\rowcolor{Gray}
S6 & 23.42& 0.892& 0.095& 0.165& 22.59& 0.884& 0.131& 0.172& 22.77& 0.869& 0.169& 0.198& \tbf{24.57}& \tbf{0.906}& \tbf{0.052}& \tbf{0.158}\\
S7 & 22.59& 0.893& 0.046& 0.140& 22.22& 0.878& 0.066& 0.143& 22.80& 0.880& 0.059& 0.152& \tbf{23.08}& \tbf{0.897}& \tbf{0.036}& \tbf{0.136}\\
\rowcolor{Gray}
S8 & 20.94& 0.876& 0.137& 0.173& 21.78& 0.882& 0.107& 0.172& 21.95& 0.886& 0.142& 0.203& \tbf{22.60}& \tbf{0.904}& \tbf{0.092}& \tbf{0.167}\\
S9 & 23.05& 0.885& 0.043& 0.141& 23.73& 0.886& 0.068& 0.141& 24.16& 0.889& 0.074& 0.152& \tbf{24.79}& \tbf{0.904}& \tbf{0.042}& \tbf{0.136}\\
\rowcolor{Gray}
S11 & 23.72& 0.884& 0.060& 0.148& 23.92& 0.889& 0.087& 0.149& 23.40& 0.880& 0.079& 0.164& \tbf{24.57}& \tbf{0.901}& \tbf{0.040}& \tbf{0.144}\\
\midrule
Avg & 22.76& 0.886& 0.076& 0.151& 22.56& 0.880& 0.106& 0.153& 22.96& 0.882& 0.112& 0.167& \tbf{23.76}& \tbf{0.902}& \tbf{0.056}& \tbf{0.146} \\
\bottomrule
\toprule
Nadia &-&-&-&-&-&-&-&-& \tbf{24.88} & \tbf{0.931} & 0.048 & 0.115& 24.44 & 0.921 & \tbf{0.026} & \tbf{0.111}\\
\rowcolor{Gray}
Weipeng &-&-&-&-&-&-&-&-& \tbf{22.45} & \tbf{0.893} & 0.039 & 0.125  & 22.07 & 0.885 & \tbf{0.024} & \tbf{0.117}\\
\midrule
Avg &-&-&-&-&-&-&-&-& \tbf{23.67} & \tbf{0.912}  & 0.044  & 0.120 & 23.25 & 0.903  & \tbf{0.025} & \tbf{0.114} \\
\bottomrule
\end{tabular}
\label{tab:exp-h36m-novel-pose}
}
\end{table}

\begin{table}[t]
\begin{minipage}[t]{0.315\linewidth}%

\newlength\tabmodscale
\setlength\tabmodscale{1.0\textwidth}
\setlength{\intextsep}{0pt}%
\caption{\small Ablation on each of the proposed modules.}
\centering
\resizebox{1.0\tabmodscale}{!}{
\setlength{\tabcolsep}{3pt}
\begin{tabular}{lcc}
\toprule
Method variant  &   PSNR~$\uparrow$ & SSIM~$\uparrow$   \\ 
\midrule
\rowcolor{Gray}
Ours w/o aggregation  & 17.08 & 0.627 \\
Ours w/o volume  & 24.24 & 0.892 \\
\rowcolor{Gray}
 Ours w/o GNN & 23.87 & 0.896 \\
Ours full  & \tbf{24.38} & 0.\tbf{899} \\
 \bottomrule
\end{tabular}
\label{tab:exp-ablation-modules}
}
\end{minipage}%
\hfill%
\begin{minipage}[t]{0.350\linewidth}%
\newlength\tabaggscale
\setlength\tabaggscale{0.4\textwidth}
\setlength{\intextsep}{0pt}%
\caption{\small Ablation on different aggregation methods.}
\centering
\resizebox{0.745\linewidth}{!}{
\setlength{\tabcolsep}{3pt}
\begin{tabular}{lcc}
\toprule
Aggregation methods  &   PSNR~$\uparrow$ & SSIM~$\uparrow$   \\ 
\midrule
\rowcolor{Gray}
Softmax  & 23.80 & 0.896 \\
Softmax-OOB  & 24.00 & 0.897 \\
\rowcolor{Gray}
Sum-OOB  & 23.22 & 0.890 \\
Sigmoid-OOB & 23.75 & 0.896 \\
\rowcolor{Gray}
Soft-softmax (Ours)  & \tbf{24.38} & \tbf{0.899} \\
 \bottomrule
\end{tabular}
\label{tab:exp-ablation-aggregate}
}
\end{minipage}%
\hfill%
\begin{minipage}[t]{0.323\linewidth}%
\newlength\tabarchscale
\setlength\tabarchscale{0.4\textwidth}
\setlength{\intextsep}{0pt}%
\caption{\small Ablation study of different coarse volumes.}
\centering
\resizebox{1.0\linewidth}{!}{
\setlength{\tabcolsep}{3pt}
\begin{tabular}{lcc}
\toprule
Volume type  &   PSNR~$\uparrow$ & SSIM~$\uparrow$   \\ 
\midrule
\rowcolor{Gray}
3D Volume (SlabConv)~\cite{Lombardi21mixture}  & 24.17 & 0.892 \\
Factorized Volume (Ours)  & \tbf{24.38} & \tbf{0.899} \\
 \bottomrule
\end{tabular}
\label{tab:exp-ablation-vol}
}
\end{minipage}%
\end{table}
\label{sec:exp-novel-pose}
Rendering of unseen poses tests how well the learned pose-dependent deformations generalize. ~\figref{fig:exp-h36m-novel-pose} shows how differences are most prominent on limbs and faces.
\ourapproach~achieves better rendering quality and retains more consistent geometric details, generalizing well to both held out poses and out-of-distribution poses (see~\figref{fig:exp-motion-retarget}).
\tabref{tab:exp-h36m-novel-pose} reports the quantitative results. DANBO consistently outperforms other baselines on Human3.6M. On MonoPerCap, the high-frequency details generated by DANBO yield lower PSNR and SSIM scores, as they penalize slightly misaligned details more than the overly smoothed results by A-NeRF. The perceptual metrics properly capture DANNO's significant quality improvement by 43\% on KID and 5\% on LPIPS. We attribute the boost in generalization and visual quality to the improved localization via graph neural networks as well as the Soft-softmax that outperforms the default softmax baseline as used in \cite{deng2019nasa,noguchi2021narf}. The ablation study below provides further insights.

Manual animation and driving of virtual models, e.g., in VR, requires such pose synthesis, for which ~\figref{fig:exp-motion-retarget} provides animation examples. Note that no quantitative evaluation is possible here as no ground truth reference image is available in this mode. Note also that the more difficult outdoor sequences are trained from a monocular video, a setting supported only by few existing approaches. %
The qualitative examples validate that ~\ourapproach{} achieves better rendering quality on 
most
subjects, and the poses generated by \ourapproach{} are sharper, with more consistent body parts, and suffer from less floating artifacts.

\subsection{Geometry Comparisons} 
\label{exp:geo-qual}
To further validate that \ourapproach{} improves the body shape reconstruction, we visualize the learned body geometry of \ourapproach{} and A-NeRF on unseen poses of the Human3.6M~\cite{Ionescu14b} dataset in~\figref{fig:exp-geometry}. \ourapproach{} captures better body shapes and per-part geometry despite also being surface-free. A-NeRF suffers from missing and shrinking body parts, and predicts noisy density near the body surface. Besides the improved completeness, \ourapproach{} shows a smoother surface, which we attribute to our coarse per-bone volumetric representation.

\subsection{Ablation Study}
\label{sec:exp-ablation}
We conduct the ablation study on S9 of Human3.6M using the same splits as before. To speed up iteration cycles, we reduce the training iterations to 100k, and use every other pose in the training set from the default configuration.
We furthermore decreased the factorized volume resolution to $M=10$.
\figref{fig:exp-ablation} shows results on both novel pose and novel view for all variants.
\parag{Proposed Modules.}
In~\tabref{tab:exp-ablation-modules}, we report how each of our proposed modules contributes to the final performance. 
For Ours w/o learned aggregation, we simply concatenate all the retrieved voxel features as inputs to the NeRF network, which is similar to A-NeRF but using GNN features. 
This leads to poor generalization, and the w/o aggregation model predicts many floating artifacts in empty space.
For Ours w/o volume, the GNN predicts per-bone feature vector instead of factorized volumes. 
In this variant, the aggregation network takes as input $\localquery_i$ to predict per-bone weights. 
The feature to neural field $\neuralfield$ is the aggregated GNN feature and local coordinates. 
While the w/o volume variant achieves comparable results, the model suffers from overfitting, and produces distorted results on joints. In sum, both our aggregation network and per-bone volume designs provide useful inductive biases for learning robust and expressive models.

\parag{Aggregation Strategy.} In~\tabref{tab:exp-ablation-aggregate},
we show the evaluation results with different aggregation methods in~\secref{sec:method-van}. Note that the Softmax variant is equivalent to NARF~\cite{noguchi2021narf} with our GNN backbone. Strategy with out-of-bound handling shows better robustness to unseen poses, with our Soft-softmax aggregation works better than Softmax-OOB, 
and the unweighted variant SUM-OOB being the worst.
\parag{Choice of Volume Representation.}
In~\tabref{tab:exp-ablation-vol}, we show the results of using both our factorized volumes, and full 3D volume predicted using SlabConv~\cite{Lombardi21mixture}. We observe that SlabConv, while capturing finer texture details as the model is more expressive, is prone to noises in the empty space. We conclude that more views and pose data are required for using SlabConv as the volume representation.
\section{Limitations and Discussion}
\newlength\figlimitscale
\setlength\figlimitscale{0.4\textwidth}
\setlength{\intextsep}{0pt}%
\newlength\figlimitimscale
\setlength\figlimitimscale{0.32\figlimitscale}
\ifblurface
\newcommand{\figlimitpost}{_blur}
\else
\newcommand{\figlimitpost}{}
\fi
\begin{wrapfigure}{r}{\figlimitscale}
\centering
\setlength{\fboxrule}{0pt}%
\parbox[t]{\figlimitimscale}{%
\centering%
\fbox{\includegraphics[width=\figlimitimscale,trim=330 600 385 130,clip]{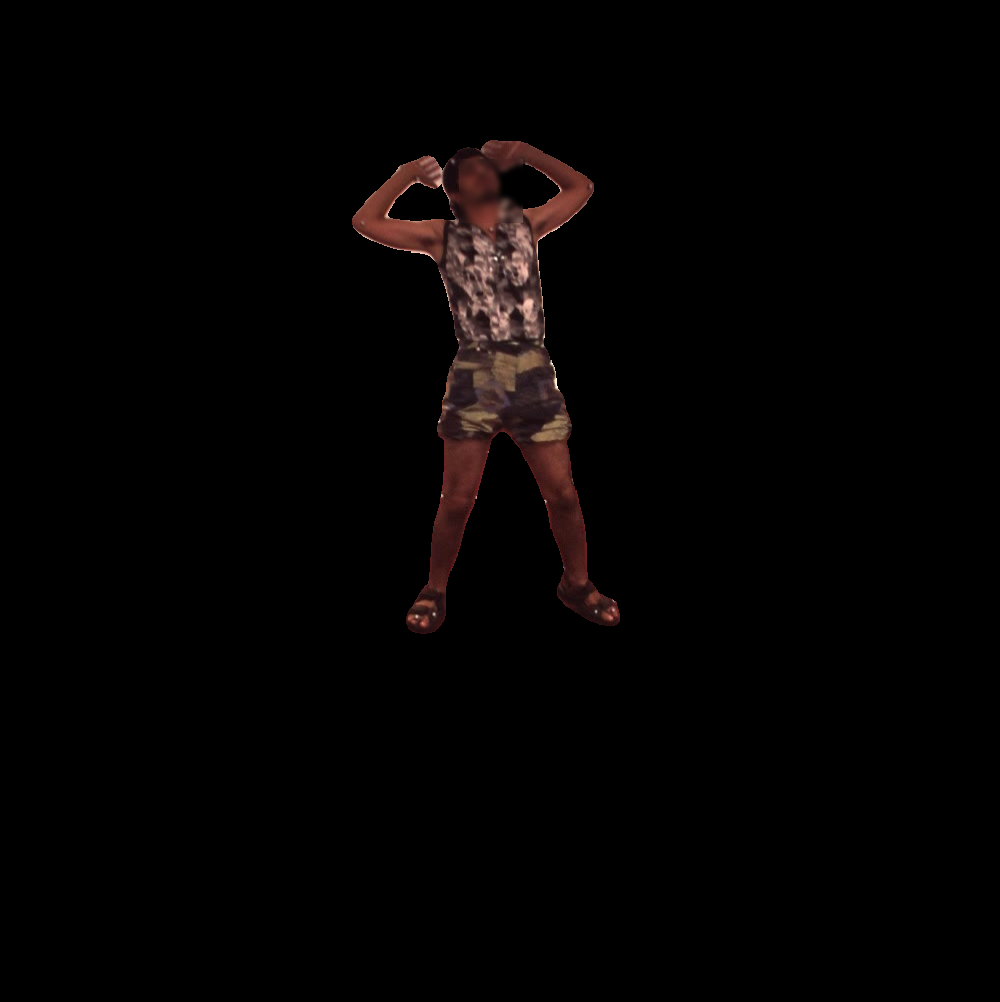}%
}\\%
{\scriptsize GT}%
}%
\hfill%
\parbox[t]{\figlimitimscale}{%
\centering%
\fbox{\includegraphics[width=\figlimitimscale,trim=330 600 385 130,clip]{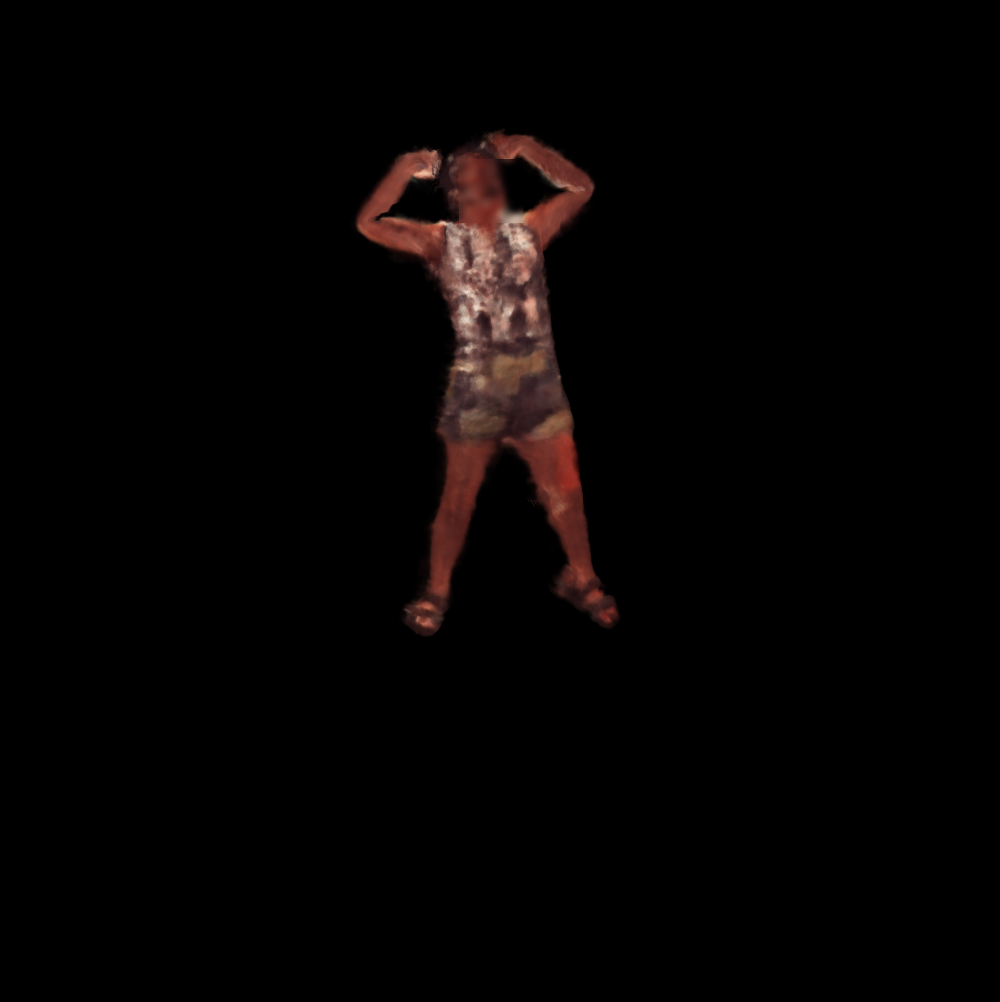}%
}\\%
{\scriptsize Anim-NeRF}%
}%
\hfill%
\parbox[t]{\figlimitimscale}{%
\centering%
\fbox{\includegraphics[width=\figlimitimscale,trim=330 600 385 130,clip]{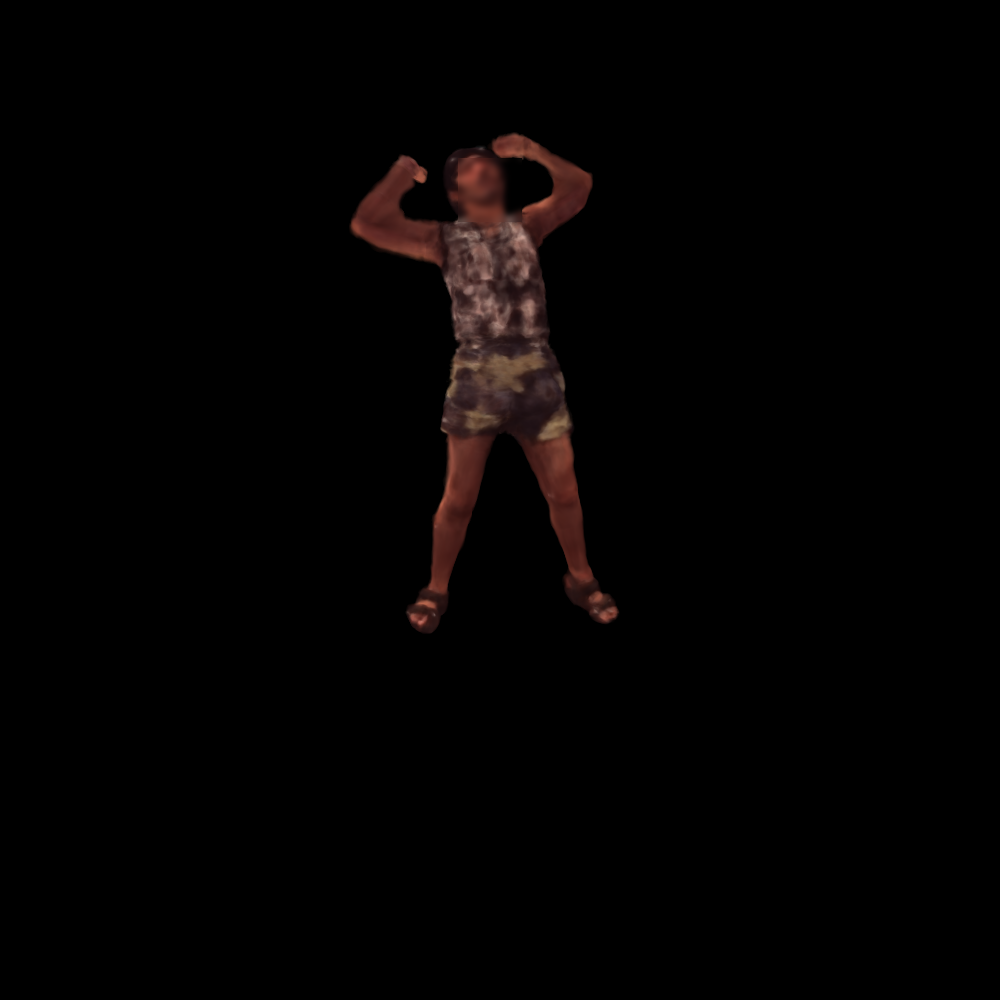}%
}\\%
{\scriptsize Ours}%
}%
\caption{\small Unseen local poses create artifacts around the joints.}
\label{fig:limitation}
\end{wrapfigure}

Similar to other neural field-based approaches, the computation time for ~\ourapproach{} remains the limiting factor for real-time applications. While~\ourapproach{} offers better generalization to unseen poses, we show in~\figref{fig:limitation} that in extreme cases it sometimes mixes the parts around joints together leading to deformation and blur. Handling such cases remains an open problem as also the surface-based method Anim-NeRF produces candy wrap artifacts around the elbow. It is also worth noting that \ourapproach{} is a person-specific model that needs to be trained individually for each person, which is desirable so long as sufficient training time and data is available.

\section{Conclusion}

We presented a surface-free approach for learning an animatable human body model from video. This is practical as it applies to monocular recordings, alleviates the restrictions of template or parametric models, and works in indoor and outdoor conditions. Our contributions on encoding pose locally with a GNN, factorized volumes, and a soft aggregation function improve upon existing models in the same class and even rival recent surface-based solutions.

\paragraph{Acknowledgements.} Shih-Yang Su and Helge Rhodin were supported by Compute Canada,  Advanced Research Computing at UBC~\cite{sockeyearc}, and NSERC DC.

\ifarxiv %
\bibliographystyle{abbrvnat}    
\else
\clearpage
\bibliographystyle{splncs04}
\fi
\bibliography{bib/egbib,bib/bibliography,bib/vision,bib/nerfhuman,bib/XNect_article}
\end{document}